\documentclass[doctor,english,final]{kaist-ucs}

\usepackage{amsmath}
\usepackage{mathtools}
\usepackage{amsfonts}
\usepackage{breqn}
\usepackage{amsmath}
\usepackage{graphicx}
\usepackage{subcaption}
\usepackage[utf8]{inputenc} % allow utf-8 input
\usepackage[T1]{fontenc}    % use 8-bit T1 fonts
\usepackage{hyperref}       % hyperlinks
\usepackage{url}            % simple URL typesetting
\usepackage{booktabs}       % professional-quality tables
\usepackage{amsfonts}       % blackboard math symbols
\usepackage{nicefrac}       % compact symbols for 1/2, etc.
\usepackage{microtype}      % microtypography
\usepackage{lipsum}
\usepackage{breakcites}
\usepackage{makecell}
\usepackage{lscape}
\usepackage{longtable}
\usepackage{algorithm} 
\usepackage{algpseudocode}

\title[korean] { 심층학습기반 도시차량궤적 분석방법론 연구}
\title[english]{ Deep Learning based Urban Vehicle Trajectory Analytics}

\author[korean] {최}{성 진}
\author[korean2] {최}{성진}    %이름을 붙여 써 주시기 바랍니다.
\author[chinese]{崔}{聖 振}
\author[english]{Choi}{Seongjin}

\advisor[major]{여 화 수}{Hwasoo Yeo}{signed}
\advisor[major2]{여화수}{Hwasoo Yeo}{signed}    %한글 성과 한글 이름을 모두 붙여 써 주시기 바랍니다.
\advisorinfo{Professor of Civil and Environmental Engineering} %제출승인서에 들어가는 교수님 정보, advisor's information 
\department{CEE}{engineering}{a}

\studentid{DCE}

\referee[1]{여 화 수}
\referee[2]{김 지 원}
\referee[3]{김 아 영}
\referee[4]{김 영 철}
\referee[5]{장 기 태}
\approvaldate{2021}{8}{15}

\refereedate{2021}{4}{20}

\gradyear{2021}

\begin{document}

    % 앞표지, 속표지, 학위논문 제출승인서, 학위논문 심사완료 검인서는
    % 클래스 옵션을 final로 지정해주면 자동으로 생성되며,
    % 반대로 옵션을 draft로 지정해주면 생성되지 않습니다.

    % 논문 서지, 초록, 핵심 낱말, 영문 초록, 영어 핵심 낱말 (Information of thesis, abstract in korean, keywords in korean, abstract in english, keywords in english)
   \thesisinfo
   %% Letters of abstract in korean must be less than 500 and words of abstract in english must be less than 300.
   %% Number of keywords must be less than 6.
   %% Don't write english letters in the abstract in korean.
    \begin{summary}      
    `궤적데이터'란, 지리적 공간에서 움직이는 물체의 위치를 시간 순서대로 정렬한 점들의 데이터로 정의할 수 있다. 위치추적 및 무선통신 기술의 발달로 여러 움직이는 물체의 궤적데이터를 수집하는 것이 가능해졌고, 이에 따라 많은 연구자들이 움직이는 물체의 이동성, 즉 모빌리티를 분석하고자 궤적데이터를 사용하고 있다. 이 논문에서는 다양한 궤적데이터 중 `도시차량궤적데이터'를 집중하고자 하며, 이를 분석하는 `도시차량궤적 분석방법론'에 대해 다루고자 한다. 도시차량궤적을 분석함으로써, 사용자 단위/네트워크 단위 교통현상들을 이해할 수 있고, 사용자들과 시스템 운영자 모두의 의사결정을 지원할 수 있기 때문이다. 도시차량궤적데이터는 시공간적 특성이 서로 구조적으로 연관되어 있다는 특징을 가지고 있다. 구조적 연관성을 분석방법론에 반영하기 위하여 많은 연구들이 진행되었고, 최근에는 뛰어난 함수근사능력과 특성표상능력을 가진 심층학습기반 분석방법론 연구가 각광을 받고 있다. 따라서 본 연구는 `심층학습기반 도시차량궤적 분석방법론'을 개발하는 것을 목적으로 한다. 이는 도시 교통 네트워크 전반의 모빌리티 패턴을 분석하여, 궤적데이터의 시공간적 연관성을 반영할 수 있기 때문이다. 구체적으로, 현 연구시점에서의 필요성과 중요도를 고려하여 다음위치예측 (Next location prediction) 문제와 가상궤적생성 (Synthetic trajectory generation) 문제를 본 연구의 주요 내용으로 선정하였다.

    % 
    % 각 문제의 난점을 해결할 수 있는 해결방안들을 제시하고, 이를 기반으로
    % 새로운 심층학습기반 모델을 개발하여 
    % 분석하고 이를 해결할 수 있는 방안을 기반으로 새로운 심층학습기반의 새로운 방법론을 제시하였다
    % 궤적데이터 전처리 단계의 중요한 문제인 지도매칭(map matching)과 궤적데이터 분석 단계 문제인 다음위치예측(next location prediction), 그리고 가상궤적생성(synthetic trajectory generation)을 이 논문의 주요 연구내용으로 설정하고, 각 문제의 난점을 해결할 수 있는 심층학습기반 모델을 개발하였다.
    % 도시차량궤적 분석방법론의 다양한 연구주제 중, 데이터 전처리 단계와 관련있는 지도매칭 (map matching)과 데이터 분석단계와 관련있는 다음위치예측, 가상궤적생성, 이렇게 세 가지 연구 주제를 이 논문의 주요 연구내용으로 선정하였다.
    % 연구 주제의 필요성, 중요도, 그리고 심층학습 적용가능성 등을 평가하여, 지도매칭(map matching), 다음위치예측(next location prediction), 가상궤적생성(synthetic trajectory generation), 세 가지를 이 논문의 주요 연구내용으로 선정하였다.

    % 첫번째 연구주제인 지도매칭에서는 데이터간의 상호 관계를 파악하여 측위오류 민감도를 줄이는 `트랜스포머'(Transformer) 기반 지도매칭 모델을 개발하였다. 
    % % 
    % 또한, 실제 차량궤적데이터에는 링크정보가 없다는 문제를 해결하기 위하여, 지도데이터로부터 가상의 레이블드 데이터(labeled data)를 생성하고, 이를 개발한 트랜스포머(Transformer) 기반 심층학습 모델에 학습시켜 실제 차량궤적을 지도에 매칭할 수 있는지를 평가하였다.

    첫번째 연구주제인 다음위치예측에서는 차량궤적의 시공간적 특성을 반영하는 심층학습기반 모델을 개발하였다. 먼저 도시 교통 네트워크를 구역으로 나누어 차량궤적을 구역시퀀스로 나타내었으며, 이를 통해 차량궤적의 공간적 특성을 추출하고, `순환신경망(Recurrent Neural Network; RNN)'을 사용하여 시공간적 패턴을 학습하는 모델을 개발하였다. 또한, 예측 정확도를 높이기 위해 추가적인 정보인 네트워크 교통 상태 정보를 다음위치예측 모델에 전달하는 `어텐션 기반 순환신경망(Attention-based Recurrent Neural Network; ARNN)' 모델을 개발하였다. 모델의 성능은 구역단위 평가지표와 궤적단위 평가지표로 나누어 평가하여 제안된 심층학습기반 모델이 비교모델보다 좋은 성능을 보인다는 결론을 도출하였다. 더 나아가, 네트워크 교통상태 정보를 통해 다음위치예측 정확도를 높일 수 있다는 것 또한 확인할 수 있었다.

    두번째 연구주제인 가상궤적생성에서는 차량궤적데이터의 시공간적 특성을 반영하여 개별차량단위/그룹단위 패턴을 모두 실제와 유사하게 생성해내는 `TrajGAIL (Generative Adversarial Imitation Learning for Urban Vehicle Trajectory)'을 개발하였다. 이를 위해 `생성적 적대 모방학습(Generative Adversarial Imitation Learning; GAIL)' 방법과 `부분적으로 관측가능한 마르코프 의사결정과정(Partially Observable Markov Decision Process; POMDP)' 방법을 활용하였다. TrajGAIL은 차량궤적의 시공간적 패턴을 이해하고 이를 통해 현실적인 차량궤적을 생성해 낼 수 있는 구조로 설계되었다.
    % 
    % 'RNN 임베딩(RNN Embeding)'을 통해 이전 위치들로부터 해당 차량의 시공간적 이동패턴 맥락을 파악하고,
    % '정책생성기(Policy Generator)'를 통해 주어진 상태에 대한 최적 행동값을 산출한다.
    % '가치평가기(Value Estimator)'는 정책생성기가 만든 행동값의 가치를 평가하여 정책생성기의 학습방향을 결정한다.
    % 마지막으로 '판별기(Discriminator)'는 실제데이터와 생성된 데이터를 비교하여 생성된 데이터가 실제와 유사하다면 높은 보상을, 다르다면 낮은 보상을 주어 정책생성기의 학습방향을 결정하는데 도움을 준다.
    % RNN embedding, policy generator, value estimator, 그리고 discriminator를 사용한다.
    % 
    차량궤적단위/데이터셋단위의 평가지표를 통해 TrajGAIL의 차량궤적 생성능력을 평가했으며, 기존에 있던 다른 생성 모델들 보다 좋은 성능을 보였다.
    
    본 연구에서는 도시 네트워크의 모빌리티 패턴을 분석하기 위해 심층학습방법을 적용한 새로운 차량궤적 분석방법론을 제안하였다. 두 주제에 대하혀 연구난점들을 파악하고 적절한 해결방안을 제시하여, 이 해결방안을 기반으로 새로운 방법론을 개발하였다. 이를 통해, 결과적으로, 도시 모빌리티에 대한 이해를 높이고, 다양한 분야에서 도시차량궤적데이터의 활용성을 증진시켜, 미래사회에 유의미한 기여를 할 것으로 판단된다.

    % 이를 통한 유의미한 사회 기여가 가능하길 기대해본다.
    % 분석기술을 발전시키고, 유의미한 사회 기여가 가능할 것으로 기대한다.
    
    % 이 연구에서는 최신 심층학습모델들을 도시차량궤적 분석방법론에 적용함으로써, 도시 네트워크의 모빌리티 패턴을 찾아내고, 도시차량궤적데이터의 시공간적 패턴을 더 잘 이해할 수 있도록, 다양한 주제를 연구하여 새로운 방법론을 개발하였다. 개발된 방법론들은 다양한 적용분야에 기여할 것이라고 기대된다.
    \end{summary}
   
    \begin{Korkeyword}
    도시차량궤적, 궤적데이터, 딥러닝, 머신러닝, 다음위치예측, 가상궤적생성
    \end{Korkeyword}

    \newpage
    \begin{abstract}
    A `trajectory' refers to a trace generated by a moving object in geographical spaces, usually represented by of a series of chronologically ordered points, where each point consists of a geo-spatial coordinate set and a timestamp.
    Rapid advancements in location sensing and wireless communication technology enabled us to collect and store a massive amount of trajectory data. As a result, many researchers use trajectory data to analyze mobility of various moving objects.
    In this dissertation, we focus on the `urban vehicle trajectory,' which refers to trajectories of vehicles in urban traffic networks, and we focus on `urban vehicle trajectory analytics.'
    The urban vehicle trajectory analytics offers unprecedented opportunities to understand vehicle movement patterns in urban traffic networks including both user-centric travel experiences and system-wide spatiotemporal patterns.
    The spatiotemporal features of urban vehicle trajectory data are structurally correlated with each other, and consequently, many previous researchers used various methods to understand this structure.
    Especially, deep-learning models are getting attentions of many researchers due to its powerful function approximation and feature representation abilities.
    As a result, the objective of this dissertation is to develop deep-learning based models for urban vehicle trajectory analytics to better understand the mobility patterns of urban traffic networks. 
    Particularly, this dissertation focuses on two research topics, which has high necessity, importance and applicability: Next Location Prediction, and Synthetic Trajectory Generation.
    % 
    
    % In map matching, we propose `Transformer-based map matching model' to reduce noise sensitivity by understanding the interrelation between trajectory data points. 
    % % 
    % Also, to resolve the issue that most of the urban vehicle trajectory data is not labeled, we propose a training technique using synthetic data generated based on map data.
    % % 
    % The Transformer-based map matching model is trained using synthetic data, and tested using real data.
    
    In next location prediction, we propose deep-learning based models that considers the spatiotemporal patterns of urban vehicle trajectories. First, we partition the urban traffic network into cells and represented urban vehicle trajectories as cell sequences to extract the spatial features from trajectory data. In addition, we used recurrent neural network (RNN) to predict the next location. Furthermore, to improve the performance of RNN model, we propose attention-based recurrent neural network (ARNN) model, which incorporates the network-wide traffic state information into next location prediction. The performance of the model is evaluated in both aggregated region level and individual trajectory level, and the proposed model has better performance than the baseline model. 
    
    In synthetic trajectory generation, we propose TrajGAIL (Generative Adversarial Imitation Learning for Urban Vehicle Trajectory), which reproduce trajectories with both patterns as individual trajectory and patterns as a group. TrajGAIL uses the Generative Adversarial Imitation Learning (GAIL) and Partially Observable Markov Decision Process (POMDP) to understand the spatiotemporal patterns and reproduce realistic trajectories. The performance of the model is evaluated in trajectory-level and dataset-level, and the results show that the proposed TrajGAIL shows an outstanding performance compared to the baseline models.
    
    In this study, we propose various novel models for urban vehicle trajectory analytics using deep learning. In three different research topics, we analyzed the current challenges in each topic, propose research approaches to resolve the challenge, and developed a novel model based on the research approaches. By using the proposed model, it is expected to increase the applicability of urban vehicle trajectories in various fields of study.
    
    \end{abstract} 
     
    \begin{Engkeyword}
    Trajectory data, Urban vehicle trajectory, Deep-learning, Machine-learning, Next location prediction, Synthetic trajectory generation
    \end{Engkeyword}

    \addtocounter{pagemarker}{1}                 % 백색별지분을 고려
    \newpage

    % 목차 (Table of Contents) 생성
    \tableofcontents

    % 표목차 (List of Tables) 생성
    \listoftables

    % 그림목차 (List of Figures) 생성
    \listoffigures

    % 위의 세 종류의 목차는 한꺼번에 다음 명령으로 생성할 수도 있습니다.
    %\makecontents

%% 이하의 본문은 LaTeX 표준 클래스 report 양식에 준하여 작성하시면 됩니다.
%% 하지만 part는 사용하지 못하도록 제거하였으므로, chapter가 문서 내의
%% 최상위 분류 단위가 됩니다.
%% You cannot use 'part'

\chapter{Introduction}

\section{Motivation and Objective}\label{sec:Intro_Motivation}
A "trajectory" refers to a trace generated by a moving object in geographical spaces, usually represented by a series of chronologically ordered points, where each point consists of a geo-spatial coordinate set and a timestamp \cite{zheng2011computing}. Throughout the last few decades, many researchers utilized different types of trajectories to enhance the understandings of movement patterns of different moving objects. For example, in meteorology, many researchers tracked meteorological events such as hurricanes and typhoons for decades and analyzed them to prevent the loss from natural disasters \cite{hubert1957hurricane,stohl1998computation}. Also, researchers in transportation engineering and urban planning are paying more attention to trajectory analytics and they analyzed patterns of pedestrians and vehicles to understand the mobility patterns of pedestrians and vehicles in cities. \cite{boltes2013collecting,rudenko2020human}.

Rapid advancements in location sensing and wireless communication technology enabled us to collect and store a massive amount of spatial trajectory data \cite{lee2011trajectory}. Over the last decade, considerable progress have been made in collecting, preprocessing, and analyzing spatial trajectory data. In transportation and urban planning, dealing with spatial trajectories is getting more and more important. This is because many moving objects including pedestrians, vehicles, and drones will be equipped with position-aware devices. Also, there will be more and more Internet of Things (IoT) devices that are communicable with the position-aware devices which collect lots of spatial trajectory data. The collected spatial trajectory data is used to explore the patterns hidden behind the data and the insights from the spatial trajectory data is very useful for planning and management of the smart cities \cite{belhadi2021deep,hu2019driving}. The spatial trajectories will be used to enhance the quality of life of people living in smart cities.

Of particular interest in transportation engineering, urban vehicle trajectory data are collected based on the location sensors installed inside vehicles or at the roadside. This high-resolution mobility data of individual users in urban transportation networks offer unprecedented opportunities to understand vehicle movement patterns in urban traffic networks. It provides rich information on both aggregated flows and disaggregated travel behaviors. The aggregated flows include the origin-destination (OD) matrix and cross-sectional link traffic volumes. The disaggregated travel behaviors include user-centric travel experiences, namely, speed profile, link-to-link route choice behavior and travel time experienced by individual vehicles, as well as system-wide spatiotemporal mobility patterns such as origin-destination pairs, routing pattern distributions, and network traffic states \cite{kim2015spatial}. Discovering and understanding the network-wide mobility patterns from the urban vehicle trajectory data can support decision-making for both individual users and system operators.

% Model based -> ML based -> DL based
Urban vehicle trajectory has both spatial and temporal features, which are structurally related to each other in the context of space and time. As a result, it is quite difficult to apply the classical data mining techniques to urban vehicle trajectory data \cite{wang2020deep}. As a result, many researchers tried to come up with a systematic way to deal with the spatiotemporal features in urban vehicle trajectory data. Early studies used models based on the statistical models and machine learning models \cite{burbey2012survey,ebrahimpour2019comparison,shi2019survey,luca2020deep,xie2020urban}. These models achieved good results in solving many problems related to urban vehicle trajectories. Nowadays, the researchers are getting more attention to the models based on \textit{Deep Learning} motivated by the outstanding successes obtained in computer vision, speech recognition, and natural language processing \cite{luca2020deep}.

% why deep learning?
Many previous studies claim that \textit{Deep Learning} has the potential to deal with complex problems in urban vehicle trajectory analytics \cite{luca2020deep, wang2020deep}. There are many advantages that models based on deep learning have \cite{goodfellow2016deep}. One of the key advantages is that the deep learning models can deal more efficiently with heterogeneous and big data source \cite{chen2016learning, yue2020deep, guo2019ifusion}. The deep learning models have automatic feature representation ability which can extract relevant features from the data automatically. This automatic feature representation ability makes it easier for deep learning models to combine raw urban vehicle trajectory data with contextual information such as weather, traffic states, traffic accidents, and census. Also, deep learning models have powerful function approximation ability, so that the models can capture complex and non-linear spatial, temporal, and sequential relationships.

% general framework of urban vehicle trajectory analytics
% Urban vehicle trajectory analytics consists of three major categories of research topics: Data collection, preprocessing, and analytics. 
% 
The urban vehicle trajectory data is collected from the GPS sensors installed in vehicles or from the road-side units (RSU) that detect vehicles passing near the RSUs. 
% Although location sensing and communication technology continues to make breakthroughs, the urban vehicle trajectory data have many noises from the collection system. Also, the urban vehicle trajectories are collected in a high sampling ratio so that the massive amount of data leads to enormous overhead in data storage, communication, and processing. 
% 
After data collection, it is necessary to preprocess the data to smoothen the effect of noises from the collection system and reduce the size of the data to properly use the urban vehicle trajectory data. Among many preprocessing techniques, the discretization of continuous urban vehicle trajectory data is gaining much attention due to its ability not only to reduce effects of local noises, but also to improve the interpretability of spatiotemporal features in urban vehicle trajectory data \cite{garcia2012survey, kim2016graph}. Such preprocessing methods by discretization include tessellation matching (using zones as a discrete set), map matching (using road links as a discrete set), and POI (Point-of-Interest) matching (using representative points as a discrete set). Particularly, map matching is considered very challenging since it cannot be modeled based on simple proximity measures unlike tessellation matching and POI matching. 

After preprocessing, the processed urban vehicle trajectory data is used in various applications. The urban vehicle trajectory analytics is gaining increasing attention from both academia and industry because of its potential to improve the performance of many applications in multiple domains. There are two widely-studied research topics: Next location prediction and synthetic trajectory generation.
Many researchers study next location prediction due to its applicability to Location-based Services (LBS). LBS uses location data of service users and provide user-specific information depending on the locations of users. Typical examples of LBS are social event recommendation, location-based advertising, and location-based incident warning system. One major advantage of next location prediction is that it provides LBSs with extended resources by giving predictive location of the users. LBSs can improve system reliability by giving more user-specific information considering their future locations \cite{karimi2003predictive}.
% 
% The next location prediction can be applied to predictively give information. 
% 
Also, synthetic trajectory generation is catching researchers' attentions to solve data sparsity problem and data privacy issues. Although the sources and availability of urban trajectory data are increasing, most of the currently available trajectory datasets cover only a portion of all vehicles in the network. From network management and operations perspectives, there is a desire to infer vehicle trajectories that represent the whole population to have a more complete view of traffic dynamics and network performance. Moreover, urban vehicle trajectory data may contain personal information of individual drivers, which poses serious privacy concerns in relation to the disclosure of private information to the public or a third party \cite{chow2011privacy}. The ability to generate synthetic trajectory data that can realistically reproduce the population mobility patterns is, therefore, highly desirable and expected to be increasingly beneficial to various applications in urban mobility.

% 
% Another example is the application on agent-based traffic simulators. Unlike traditional traffic simulators which consider traffic demand as input, an agent-based traffic simulator requires information on individual vehicle journey such as origin, destination, and travel routes \cite{martinez_agent-based_2015}. The result of trajectory generation can be used for real-time application of these agent-based traffic simulators. 

% The urban vehicle trajectory data is applied to diverse applications such as next location prediction and synthetic trajectory generation after preprocessing.
% 
% As mentioned in Section \ref{sec:Intro_Motivation}, the objective of this study is to identify research problems related to urban vehicle trajectories and develop models to solve those problems based on the state-of-the-art deep learning models. 

% objective of this dissertation
% Motivated by the current trends of researches, the overall objective of this dissertation is to identify research problems related to urban vehicle trajectories and develop novel approaches to solve those problems based on the state-of-the-art deep learning solutions, in order to better understand the spatiotemporal patterns underlying in urban vehicle trajectory data. 
Motivated by the current trends of researches, the overall objective of this dissertation is to apply state-of-the-art deep learning solutions to resolve the issues in urban vehicle trajectory analytics. Specifically, this dissertation focuses on two research topics in urban vehicle trajectory analytics which are considered important and challenging. 

% \colorbox{yellow}{add more context}
% Additionally, this dissertation focuses mainly on three research topics 
% scope of the study
% Specifically, among many research problems related to urban vehicle trajectory data, this dissertation mainly focuses on the following three research topics: map matching, next location prediction, and synthetic trajectory generation. 
\begin{itemize}
    % \item \textbf{Map matching }is a task in trajectory data preprocessing. Map matching can be defined as a problem of matching recorded geographic coordinates to a logical model of the real world. In other words, map matching is matching raw GPS points to the traffic road networks, or map data. Map matching is one of the essential part of vehicle trajectory analysis in preprocessing steps.
    % 
    \item \textbf{Next location prediction} can be defined as forecasting the next location of an individual vehicle based on the historical data. Next location prediction has gained much attentions of many researchers due to its applicability to many fields such as travel recommendation, location-based services, and location-aware advertisements.  
    \item   \textbf{Synthetic trajectory generation} can be defined as generating synthetic trajectories with realistic spatiotemporal mobility patterns based on the historical trajectory dataset. Although sources and availability of urban trajectory data are increasing, most of the currently available trajectory datasets cover only a portion of all vehicles in the networks. As a result, some kind of augmentation method is needed to have full population of the vehicles. Also, urban vehicle trajectories contain personal information of individual drivers such as the location of their home and work places.  
\end{itemize}

\section{Structure of Dissertation}
The structure of this dissertation is organized as follows. 
Chapter \ref{chapter:prelim} introduces general framework of urban vehicle trajectory analytics, and discuss each step in general framework by reviewing the previous researches related to each step. 
In Chapter \ref{chapter:framework}, two main research topics (next location prediction, synthetic trajectory generation) are introduced, which are considered important and challenging based on the review in Chapter \ref{chapter:prelim}. Chapter \ref{chapter:framework} covers current challenges in each research topics and address research approaches to resolve the issues found in the challenges. 

There are three main chapter; two for next location prediction, and one for synthetic trajectory generation.
%
% with definition and explanation on why these research topics are important and why they are challenging to solve. 
% In Chapter \ref{chapter:mapmatching}, a novel map matching algorithm based on \textit{Transformer} is proposed to understand correlation of each data point, and utilize this understanding for map matching task. Also, in this chapter, a novel training technique for map matching task is proposed which use synthetic dataset generated by the map data (i.e. shp files). 
Chapter \ref{chapter:nlp_rnn} and Chapter \ref{chapter:nlp_arnn} studies deep learning application in next location prediction. In Chapter \ref{chapter:nlp_rnn}, urban vehicle trajectories are summarized by clustering-based Voronoi tessellation and represented as sequence of cells (spatial tessellations), and a novel model based on recurrent neural networks (RNN) is proposed to predict the next location (next cell) of individual vehicles. In addition, Chapter \ref{chapter:nlp_arnn} introduces attention-based recurrent neural networks (ARNN) which incorporates network-wide traffic states into next location prediction.
Chapter \ref{chapter:trajgen} presents TrajGAIL for synthetic trajectory generation. TrajGAIL uses partially observable Markov decision process (POMDP) and Generative Adversarial Imitation Learning (GAIL) to generate urban vehicle trajectories with realistic mobility patterns. 

\begin{figure}[!ht]
  \centering
  \includegraphics[width=\textwidth]{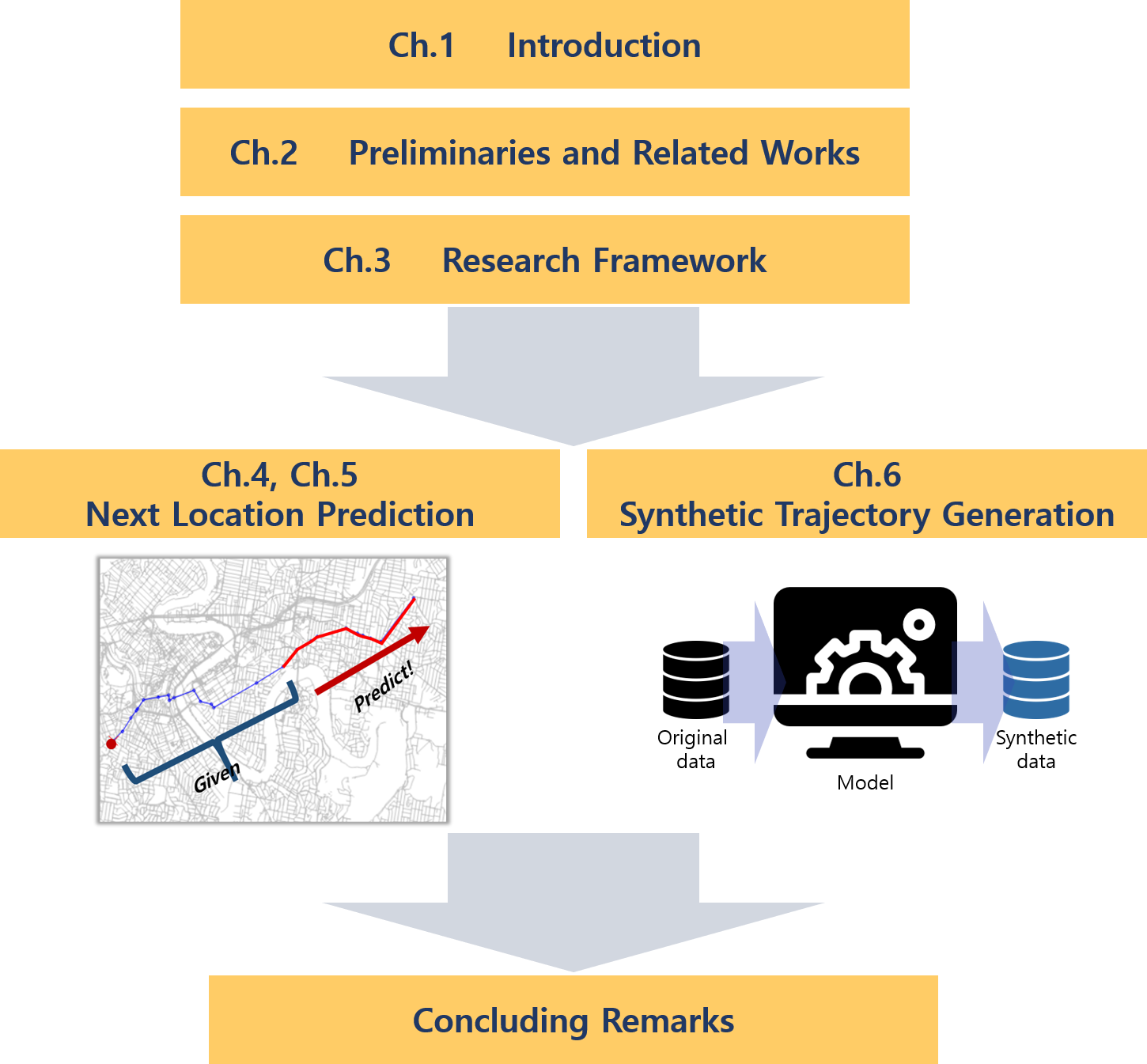}
  \caption{A graphical representation of the structure of this dissertation}
  \label{fig:introduction_dissertation_structure}
\end{figure}

Chapter \ref{chapter:conclusion} provides conclusion of dissertation and summary of each main research topics with contributions and limitations of the current study, and future research directions in urban vehicle trajectory analytics. A graphical representation of the structure of this dissertation is presented in Figure \ref{fig:introduction_dissertation_structure}

\newpage
% \begin{center}
    
% \end{center}
\newpage

\chapter{Preliminaries and Related Works}\label{chapter:prelim}

\section{Urban Vehicle Trajectory Analytics}
\begin{figure}[!ht]
  \centering
  \includegraphics[width=\textwidth]{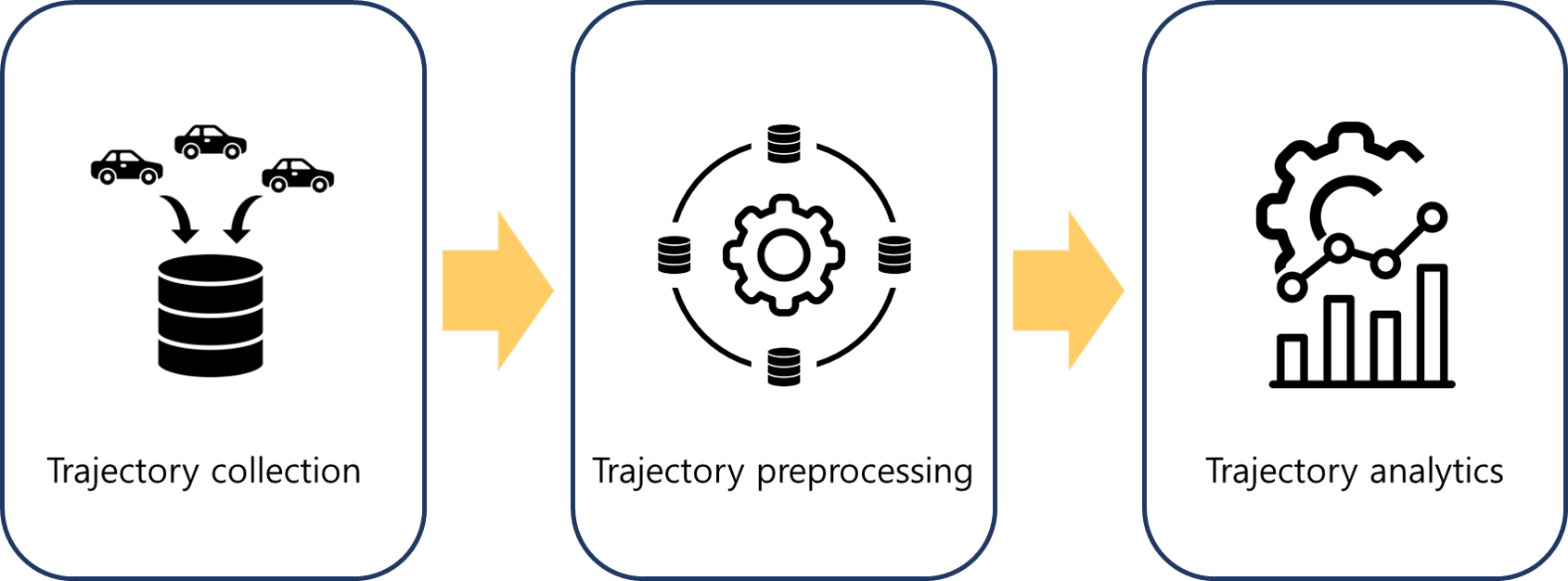}
  \caption{General framework of urban vehicle trajectory data analysis}
  \label{fig:gen_frame}
\end{figure}

% general framework of urban vehicle trajectory data analysis
In this section, the preliminaries on the urban vehicle trajectory data mining is explained. Over the last few decades, many researchers had dealt with urban vehicle trajectory data, and many literatures have a common general structure when dealing with urban vehicle trajectory data. Figure \ref{fig:gen_frame} shows the general framework of urban vehicle trajectory data analytics. The general framework of urban vehicle data analytics have three steps: Trajectory collection, trajectory preprocessing, and trajectory analytics.

\subsection{Urban Vehicle Trajectory Data Collection} \label{sec:prelim_datacollection}

The most common way to achieve urban vehicle trajectories is to collect the coordinates of the subject vehicle by GPS sensors installed in the vehicle. Global Navigation Satellite Systems (GNSS) are an essential source of human mobility data \cite{luca2020deep}. The US Global Positioning System, or GPS, is one of the most well-known and widely-used GNSS. GPS receivers are ubiquitous in many tools of everyday life, such as mobile phones \cite{alessandretti2018evidence}, vehicles \cite{gallotti2016stochastic, pellungrini2017data}, and vessels \cite{praczyk2019ship}. Especially on vehicles, the GPS device automatically turns on when the vehicle starts. Also, nowadays, the mobile phone connected to the vehicle can work as a GPS device when the phone is connected to the navigation applications. A typical GPS trace is a set of tuples $(u, t, lng, lat)$, where $u$ is a user, $t$ is the timestamp of the measurement and $lng$, $lat$ are the longitude and latitude of the current position. The precision of GPS receivers varies from a few centimeters to meters, depending on the quality of the GPS receiver and the errors generated by the system \cite{carlson2010mapping}. GPS is complex, and the errors arise from a wide variety of sources with different dependencies and characteristics. The raw GPS data is embedded with an inseparable error from the location positioning system. As a result, it is required to preprocess raw GPS data to mitigate these errors and extract meaningful semantics. 

Another way of achieving urban vehicle trajectory data is by using roadside units (RSUs) installed in the cities. The \textit{RSU-based} urban vehicle trajectories are collected from the infrastructures installed alongside the roads. The vehicles are equipped with communication devices, usually Bluetooth receivers and Dedicated Short Range Communication (DSRC). When the vehicles pass near the roadside infrastructure, the roadside unit collects the timestamp, as a passage time of a vehicle, and the unique identifier assigned to the passing vehicle. The urban vehicle trajectories are obtained by tracing this unique identifier of the communication device. Each trajectory represents a sequence of the locations of roadside infrastructures that a vehicle passed along its journey.
% where the error is coming from
The quality of the RSU-based urban vehicle trajectories is highly dependent on the roadside units that detect the vehicles \cite{michau2017bluetooth}. It is ideal for installing the roadside units at every intersection to collect a complete set of vehicle trajectories. However, due to the cost problem, many cities selectively install roadside units at the main intersections. For example, Brisbane has Bluetooth scanners scattered along with the road networks. The coverage of this scanner is dense in the Central Business District (CBD) area, but scarce in the suburb area. In the areas where the scanner is not installed, it is not possible to detect the vehicles and analyze the movement patterns. Also, there are some problems in the areas where the roadside units are densely installed. Sometimes, these roadside units have overlapping detection areas, which makes errors in the sequence. For example, in some cases, a vehicle can be detected in the downstream detector first and then detected by the upstream detector. Furthermore, the scanners installed in the roadside units can have missed detection. According to \cite{michau2017bluetooth}, about 20\% of the detections are missing in the case of Bluetooth scanner in Brisbane city. 

% need to preprocess
% 
% There also are many sensor errors from passively recorded urban vehicle trajectories. 
% According to \cite{michau2017bluetooth}, there are two main cause of errors. First cause is the overlapping detection areas 
\subsection{Urban Vehicle Trajectory Preprocessing}
% why need preprocessing
Both \textit{GPS-based} and \textit{RSU-based} urban vehicle trajectories contain meaningful spatiotemporal patterns in urban transportation networks, which can be used in a variety of applications. However, there are a number of problems to be resolved. 
If the urban vehicle trajectories are collected in a high sampling ratio, a massive amount of data would lead to enormous overhead in data storage, communications, and processing.
Also, as we discussed in Section \ref{sec:prelim_datacollection}, urban vehicle trajectory data usually contains inevitable noise from the sensors and collection systems. Sometimes, the noise of the raw data reduces the effectiveness of systems that use such trajectories. As a result, many previous researchers use many different preprocessing techniques to filter out the noisy points and reduce the size of the data.

One of the most common preprocessing techniques for urban vehicle trajectories is denoising by trajectory filtering. Since the vehicle trajectories are not perfectly accurate due to sensor noise and other factors, it is required to use various filtering techniques to the trajectory to smooth the noise and potentially decrease the error in the measurements. The simplest form of filtering the noise from trajectories is using \textit{mean and median} filters. For a measured point, $z_i$, the estimate of the actual location $\hat{z}_i$ is the mean (or median) of $z_i$ and its $n-1$ predecessors in time. Although mean and median filters are both simple and powerful techniques to deal with the noise, one of the most significant disadvantage is that they have lags. If the actual location changes suddenly, the estimate from the mean and median filter cannot react as suddenly as the actual value. The estimate will only respond gradually. Kalman filters and particle filters use a measurement model and dynamic models to improve the accuracy of the estimates. Although it is not a simple task to formulate both measurement models and dynamic models, Kalman filters and particle filters overcame the lag problem in mean and median filters.

Another way of preprocessing raw vehicle trajectories is to convert the continuous values of the data points in raw vehicle trajectories to a finite set of discrete values. Discretization of continuous features is one of the common techniques in data mining. Through discretization, it is possible to smoothen the local noises, reduce the size of the raw data \cite{pyle1999data}. Furthermore, a proper discretization would increase the interpretability of the features from the data \cite{garcia2012survey}. The discretization process transforms quantitative data into qualitative data. In other words, it transforms continuous or numerical attributes into discrete or nominal attributes with a finite number of intervals, obtaining a non-overlapping partition of a continuous domain. 
 
There are several types of discretization one can use when dealing with the urban vehicle trajectory data. The first approach is to partition the transportation networks into zones, sometimes called cells, and use the sequence of zones (or sequence of cells) instead of the raw vehicle trajectories. Matching continuous coordinates to the predefined tessellations (zones) is called "\textit{Tessellation Matching}." The second approach is to use the links in the road networks as the discrete value and use the sequence of links instead of the raw vehicle trajectories. Matching continuous coordinates to the road link is called "\textit{Map Matching}." Map matching has widely been studied by many different researchers due to its importance in urban vehicle trajectory analytics. The last approach is to define \textit{Point-of-Interest} (POI) first and use the POIs as the representative locations for continuous coordinates. The process of matching coordinates to POIs is called "\textit{POI matching}," and usually, researchers select the nearest POI for each coordinate.
There is certainly a trade-off when using a sequence of aggregated discrete zones or cells because it loses microscopic features such as the speed profile of the subject vehicle within a link. However, when dealing with a massive amount of vehicle trajectory data in large-scale urban traffic networks, it is desirable to use zones and links because it is easier to analyze the spatiotemporal patterns of the urban traffic networks.

% TODO :: cell-level analytics vs link-level analytics
In fact, deciding a right discretization method relies on the scale of the analytics that a researcher would like to conduct. Using tessellation matching for preprocessing urban vehicle trajectory data can be beneficial for large-scale network analytics, while using map matching can be beneficial for analyzing movement patterns in smaller networks. POI matching can be used for both scales, because the same process applies when deciding which POIs to use (distantly located POIs for large-scale networks, and closely located POIs for small networks). Some examples of analytics using tesellation matching for large-scale networks include \cite{krumm2006predestination,krumm2007predestination,calabrese2010human,endo2017predicting,choi2018network,choi2019attention}, and examples of analytics using map matching for small networks include \cite{horvitz2012some, ziebart2008maximum, ziebart2008navigate, choi2019real, chen2021trajvae}

\subsection{Urban Vehicle Trajectory Analytics}

Applying data analytics on urban vehicle trajectory data makes it possible to discover complex patterns in urban vehicle mobility and obtain deeper insights into travel behaviors and traffic dynamics. Also, it allows road operators and transit agencies to identify opportunities to improve their systems. 
The high-resolution mobility data of individual users in urban road networks offer unprecedented opportunities to understand vehicle movement patterns in urban traffic networks. It provides rich information on both aggregated flows and disaggregated travel behaviors. The aggregated flows include the origin-destination (OD) matrix and cross-sectional link traffic volumes. The disaggregated travel behaviors include user-centric travel experiences, namely, speed profile, link-to-link route choice behavior and travel time experienced by individual vehicles, as well as system-wide spatiotemporal mobility patterns, such as origin42 destination pairs, routing pattern distributions, and network traffic states \cite{kim2015spatial}. 
In recent years, the urban vehicle trajectory analytics is gaining increasing attention from both academia and industry because of its capability to analyze the mobility pattern of vehicles in cities on different scales. The urban vehicle trajectory analytics gives detailed information on vehicle mobility, which consists of both patterns from the individual vehicles and patterns as aggregated traffic flow. In contrast to conventional traffic data analytics, which focuses on data collected in fixed locations, urban vehicle trajectory analytics include both analytics on an individual vehicle and aggregated traffic flow.

There are several examples of applications of urban vehicle trajectory analytics. One example is Location-based Service (LBS). LBS uses location data of service users and provide user-specific information depending on the locations of service users. Typical examples of LBS are social event recommendation, location-based advertising, and location-based incident warning system. The location prediction can be applied to predictively give information; for example, if a user’s next location is expected to be disastrous or congested, the service informs the user to change route. Furthermore, when it is not possible to continue to give service because the position of the user is lost due to sensor malfunctioning, predicting the locations of the user can temporally replace the role of positioning system and continue the service \cite{monreale2009wherenext, morzy2007mining}. Another example is the application on agent-based traffic simulators. Unlike traditional traffic simulators which consider traffic demand as input, an agent-based traffic simulator requires information on individual vehicle journey such as origin, destination, and travel routes \cite{martinez_agent-based_2015}. The result of urban vehicle trajectory analytics can be used for real-time application of these agent-based traffic simulators. The urban vehicle trajectory analytics can also be applied to inter-regional traffic demand forecasting. As the market of ride-sharing is continuously growing and Shared Autonomous Vehicles (SAV) are expected to be on our roads in the near future, there is a strong need to be able to predict inter-regional traffic demand so as to dispatch the proper number of SAV to areas of high demand. Location prediction model can be used to identify the demand hotspots by learning the mobility pattern of the users.

There are several research topics in urban vehicle trajectory analytics. 
% Among many research topics, this dissertation focuses on two research topics in urban vehicle trajectory analytics: 
\textit{Next Location Prediction} and \textit{Synthetic Trajectory Generation} are two most widely-studied research topics.
% 
% These research topics can be categorized into two categories. 
% The researches in first category studies methods to estimate and predict the aggregated traffic variables from the urban vehicle trajectory data. The examples include \textit{crowd flow prediction} and \textit{origin-destination estimation}. 
% % add examples
% 
% On the other hand, the researches in the second category studies methods to deal with the individual trajectories. The examples include \textit{next location prediction} and \textit{synthetic trajectory generation}. 
% 
\textit{Next location prediction} aims to develop model to predict the future location of a vehicle based on the historical trajectory data. Predicting future locations based on previously visited locations has been widely studied in terms of predicting location where the user will visit next \cite{noulas2012mining,gambs_next_2012,mathew2012predicting}, the location where the user ends the trip \cite{krumm2006predestination,krumm2007predestination,horvitz2012some,xue2015solving,ziebart2008navigate,marmasse2002user}, and the location where the user will visit in the next time interval \cite{hawelka2015collective,alhasoun2017city,lu2013approaching,calabrese2010human,zhao2017mobility}. 

The first and the second types understand individual trips as a sequence of locations, similar to the one explained previously. The latter one predicts the location that the user will visit in the next time interval, which is usually set as an hour. This may be widely applicable since this adapts temporal characteristics of mobility; however, such a task requires frequent updates of the user’s actual location. Also, one of the major problems is that most of the trips end in less than 30 minutes to 1 hour in the urban area as the majority of the trips are for commuting or visiting a commercial area. Therefore, it is hard to distinguish if the users are still traveling or staying. These methods are modeled to solve more macroscopic trips than city-scaled problems.  For example, in \cite{zhao2017mobility}, the authors presented N-gram model to predict the trip time, entry and exit station. They used the Oyster entry and exit records data collected from the London Underground, Overground and National Rail. 

There are several previous studies that used machine-learning models to predict the future location or the destination of a trip. One of them, \cite{gambs_next_2012}, used Mobility Markov Chain to predict the next location of an individual. The research was based on the observations of individual’s mobility so that the model must be specified by each individual. Also, in \cite{mathew2012predicting}, Hidden Markov Model is used to predict pedestrian movement by using GeoLife dataset. Hidden Markov Model computes latent state at each sequence, which maximizes the likelihood of input sequence. Usually, the number of latent states or the number of clusters is given. The Hidden Markov Model calibrates the transition matrix among the latent states and emission probability to decode latent states to observable sequences. Some previous studies also tried to use Artificial Neural Networks (ANN) in trajectory prediction. Recent work by \cite{de2015artificial} includes a study on the prediction of taxi destination using Multilayer Perceptron (MLP). They represented the destination as a linearly weighted combination of predefined destination clusters. The result showed that the overall distance error is considered negligible; however, it is pointed out that it is challenging to predict unpopular destinations. 

With the recent development of Deep Neural Network, including RNN models, and computation powers, there have been some researches in the transportation field to predict the microscopic vehicle location for autonomous vehicles \cite{kim2017probabilistic} and also predict mobility sequences \cite{endo2017predicting,liu2016predicting}. The research in \cite{kim2017probabilistic} used RNN with Long Short Term Memory (LSTM) to predict the vehicle movement in front of a subject vehicle. \cite{endo2017predicting} used RNN to predict destination. In \cite{endo2017predicting}, trajectory sequence is represented as a sequence of locations in a discretized grid space, which is an arbitrary network partitioning. \cite{choi2019real} used a feed-forward neural network to predict the next intersection in a grid-structured road network. A set of intersections in Brisbane, Australia, are treated as POI's to capture the link-to-link route choice behavior. \cite{jin2019augmented} used an augmented-intention recurrent neural network model to predict locations of vehicle trajectories of individual users. \cite{jin2019augmented} incorporated additional information on individual users' historical records of frequently visited locations into a next location prediction model. The past visited locations in historical records are represented as an edge-weighted graph, and a graph convolution network is used to incorporate this information into trajectory prediction. In \cite{choi2018network}, an urban road network is partitioned into zones based on the clustering of trajectory data points. The prediction model based on recurrent neural network (RNN) is proposed to predict the zone that the subject vehicle would visit. \cite{choi2019attention} extended the idea of predicting the next zone and used network traffic state information to improve the RNN model's performance.

\textit{Synthetic trajectory generation} aims to develop a model to generate synthetic (fake) trajectories with realistic spatiotemporal mobility patterns. Synthetic data generation has gained increasing importance as the data generation process plays a significant role in various research fields in an era of data-driven world \cite{popic2019data}. It is mainly used to serve two purposes. The first purpose is to deal with the lack of real data. In many research fields, data collection is costly, and, therefore, it is often difficult to collect enough data to train and validate models properly. In this case, it is useful to generate synthetic data similar to the real observations to increase training and test samples. The second purpose is to address the issue with the privacy and confidentiality of real data. Many types of data contain personal information, such as gender, name, and credit card usage. Synthetic data can be combined with or replace such privacy-sensitive data with a reasonable level of similarity, thereby protecting privacy while serving the intended analysis.

Urban vehicle trajectory analytics has both challenges: data sparsity and data privacy issues. Although the sources and availability of urban trajectory data are increasing, most of the currently available trajectory datasets cover only a portion of all vehicles in the network. From network management and operations perspectives, there is a desire to infer vehicle trajectories representing the whole population to have a complete view of traffic dynamics and network performance. Moreover, urban vehicle trajectory data may contain personal information of individual drivers, which poses serious privacy concerns in relation to the disclosure of private information to the public or a third party \cite{chow2011privacy}. Therefore, the ability to generate synthetic trajectory data that can realistically reproduce the population mobility patterns is highly desirable and expected to be increasingly beneficial to various applications in urban mobility.

While synthetic trajectory generation is a relatively new topic in transportation research communities, several existing research areas have addressed similar problems. 
One example is \textit{trajectory reconstruction}. When two points in a road network are given as an initial point (treated as sub-origin) and a target point (treated as sub-destination), the models reconstruct the most plausible route between the two points. The trajectory reconstruction can be considered as generating trajectories between sub-origins and sub-destinations. Previous studies such as \cite{chen2011discovering} and \cite{hu2018graph} investigated discovering the most popular routes between two locations. \cite{chen2011discovering} first constructs a directed graph to simplify the distribution of trajectory points and used the Markov chain to calculate the transfer probability to each node in the directed graph. The transfer probability is used as an indicator to reflect how popular the node is as a destination. The route popularity is calculated from the transfer probability of each node. \cite{hu2018graph} also used a graph-based approach to constructing popular routes. The check-in records which contain the route's attributes are analyzed to divide the whole space into zones. Then, the historical probability is used to find the most plausible zone sequences. Also, \cite{feng2015vehicle} and \cite{rao2018origin} estimated origin-destination patterns by using trajectory reconstruction. Both studies used particle filtering to reconstruct the vehicle trajectory between two points in automatic vehicle identification data. The reconstructed vehicle trajectory is then used to estimate the real OD matrix of the road network.

In fact, the existing models developed for the next location prediction problem can be applied for synthetic trajectory data generation. By sequentially applying the next location predictions, a synthetic vehicle trajectory can be generated. However, most of the existing models for next location prediction adopt a discriminative modeling approach, where the next locations are treated as labels, and the model is trained to predict only one or two next locations. The discriminative models have limitations in generating full trajectories, especially when sample trajectory data are sparse. It is only the decision boundaries between the labels that the models are trained to predict, not the underlying distributions of data that allow proper generalization for sampling realistic trajectories. As a result, it is necessary to develop a model based on the generative modeling approach to perform synthetic trajectory data generation successfully.

% generative adversarial framework : GAN
Recently, there have been remarkable breakthroughs in generative models based on deep learning. In particular, \cite{goodfellow2014generative} introduced a new generative model called Generative Adversarial Networks (GAN), which addressed inherent difficulties of deep generative models associated with intractable probabilistic computations in training. GANs use an adversarial \textit{discriminator} to distinguish whether a sample is from real data or from synthetic data generated by the \textit{generator}. The competition between the generator and the discriminator is formulated as a minimax game. As a result, when the model is converged, the optimal generator would produce synthetic sample data similar to the original data. The generative adversarial learning framework is used in many research fields such as image generation \cite{radford2015unsupervised}, audio generation \cite{oord2016wavenet}, and molecular graph generation \cite{de2018molgan}. 

GANs have also been applied in transportation engineering. \cite{zhang2019novel} proposed trip travel time estimation framework called \textit{T-InfoGAN} based on generative adversarial networks. They used a dynamic clustering algorithm with Wasserstein distance to make clusters of link pairs with similar travel time distribution. They applied Information Maximizing GAN (InfoGAN) to travel time estimation. \cite{xu2020ge} proposed Graph-Embedding GAN (GE-GAN) for road traffic state estimation. Graph embedding is applied to select the most relevant links for estimating a target link, and GAN is used to generate the road traffic state data of the target link. In \cite{li2020coupled}, GAN is used as a synthetic data generator for GPS data and travel mode label data. To solve the sample size problem and the label imbalance problem of a real dataset, the authors used GAN to generate fake GPS data samples of each travel mode label to obtain a large balanced training dataset.
The generative adversarial learning framework is also used for synthetic trajectory generation. \cite{liu2018trajgans} proposed a framework called trajGANs. Although this study does not include specific model implementations, it discusses the potential of generative adversarial learning in synthetic trajectory generation. Inspired by \cite{liu2018trajgans}, \cite{rao2020lstm} proposed LSTM-TrajGAN with specific model implementations. The generator of LSTM-TrajGAN is similar to RNN models adopted in the next location prediction studies. 

Previous studies which does not use deep neural networks can be categorized as "probabilistic" models or "pattern-matching" models. The probabilistic models, or Markov-based models, use Markov assumptions to model the probability distribution of the next location (or a trajectory). Common examples are \cite{gambs2010show,gambs_next_2012,calabrese2010human}. The pattern-matching models use tree structures to find the similar patterns from the historical dataset. The examples are \cite{monreale2009wherenext,wang2013mining,xia2018decision}. Although these models have some degree of interpretability and can achieve good performances with a small amount of data, one of the major disadvantages of this approach is that they require a considerable effort in feature engineering and have limited memory, making it hard for them to capture long-range temporal dependencies \cite{sabarish2015survey}. On the other hand, recent approaches using deep learning models can overcome the disadvantages of the non-deep-learning models by using great function approximation and pattern recognition ability of deep neural networks.

% add examples

% Analyzing the mobility patterns from the individual vehicles is called \textit{"Agent-based modelling"}

\newpage
\begin{landscape}

\begin{table}[!ht]
\caption{Examples of researches on urban vehicle trajectory analytics}
\label{tab:review_urban_vehicle_trajectory_analytics}
\begin{center}
\resizebox{1.3\textheight}{!}{
\begin{tabular}{l l l l l}
\hline\hline
Reference & Data & Preprocessing & Model & Analytics  \\ \hline
% \cite{ziebart2008maximum} & 2008 & GPS-based & Road link & Next location prediction \\
% \cite{monreale2009wherenext} & 2009 & GPS-based & - & Next location prediction\\
% \cite{chen2011discovering} & 2011 & GPS-based & Road link & Trajectory reconstruction \\
% \cite{fan2015citymomentum} & 2015 & GPS-based & POI & Crowd flow prediction \\
% \cite{feng2015vehicle} & 2015 & GPS-baaed & POI & Trajectory reconstruction \\
% \cite{zhang2017deep} & 2017 & GPS-based & Squre grid tessellation & Crowd flow prediction\\
% \cite{choi2018network} & 2018 & RSU-based & Clustering-based Voronoi tessellation & Next location prediction      \\
% \cite{hu2018graph} &2018 & GPS-based & Square grid tessellation & Trajectory reconstruction\\
% \cite{rao2018origin} & 2018 & RSU-based & POI  & OD estimation \\
% \cite{choi2019attention} & 2019 & RSU-based & Clustering-based Voronoi tessellation & Next location prediction      \\
% \cite{choi2019real} & 2019 & RSU-based & POI (intersections) & Next location prediction\\
% \cite{rao2020lstm} & 2020 & GPS-based & POI & Trajectory generation \\
% \cite{ren2020hybrid} & 2020 & GPS-based & Square grid tessellation & Crowd flow prediction \\
% 
\hline
\cite{noulas2012mining} & Foursquare (GPS-based) & - & Mobility feature-based model & Next location prediction \\ \hline
\cite{gambs_next_2012} & Phonetic, GeoLife (GPS-based) & POI matching & Mobility Markov Chain & Next location prediction \\ \hline
\cite{mathew2012predicting} & GeoLife (GPS-based) & - & Hidden Markov Model & Next location prediction \\ \hline
\cite{krumm2006predestination} & Microsoft Multiperson Location Survey & Square grid tessellation & Predestination & Destination prediction \\ \hline
\cite{horvitz2012some} & Seattle GPS data & Map matching & Opportunistic routing & Destination prediction\\ \hline
\cite{xue2015solving} & T-drive & - & SubSyn & Destination prediction\\ \hline
\cite{ziebart2008maximum} & Yellow Cab Taxi data & Map matching & MaxEnt & Next location prediction \\ \hline
% \cite{ziebart2008navigate} & & & & \\
\cite{marmasse2002user} & GPS-based & - & \makecell[l]{Bayes Classifier\\Histogram Modeling\\Hidden Markov Model}  & Next location prediction  \\ \hline
\cite{alhasoun2017city} & CDR (RSU-based) & POI matching & Dynamic Bayesian Networks & Next location prediction\\ \hline
\cite{lu2013approaching} & GPS-based & - & Entropy-based model & Next location prediction \\ \hline
\cite{calabrese2010human} & AirSage (GPS-based) & Square grid tessellation & \makecell[l]{Individual and collective\\ behavior modeling} & Next location prediction \\ \hline
\cite{zhao2017mobility} & GPS-based & - & Bayesian n-gram & Next location prediction \\  \hline
\cite{kim2017probabilistic} & GPS-based & Grid tessellation & Recurrent Neural Networks & Next location prediction \\ \hline
\cite{endo2017predicting} & \makecell[l]{Taxi service trajectory (GPS-based) \\ Geolife (GPS-based)} & Square grid tessellation & Recurrent Neural networks & Destination prediction \\ \hline
\cite{liu2016predicting} & \makecell[l]{Gowalla (GPS-based) \\ GTD (GPS-based) } & POI matching & Factorizing Personalized Markov Chain & Next location prediction \\ \hline
\cite{jin2019augmented} & \makecell[l]{Wifi sensor (RSU-based) \\ Foursquare (GPS-based)} & - & Augmented Intent Neural Network & Next location prediction \\ \hline
\cite{choi2018network} & \makecell[l]{Brisbane Bluetooth Data \\(RSU-based)} & Clustering-based Voronoi tessellation& Recurrent Neural Networks& Next location prediction \\ \hline
\cite{choi2019attention} & \makecell[l]{Brisbane Bluetooth Data \\(RSU-based)} & Clustering-based Voronoi tessellation & Attention-based Recurrent Neural Networks & Next location prediction  \\ \hline
\cite{choi2019real} & \makecell[l]{Brisbane Bluetooth Data \\(RSU-based)} & POI matching & Multi-layer Perceptron & Next location prediction  \\ \hline
\cite{zhang2019novel} & Didi Chuxing (GPS-based) & Map matching & T-infoGAN & Travel time estimation \\ \hline
% \cite{xu2020ge} & & & GE-GAN & Traffic state estimation \\ \hline
% \cite{li2020coupled} & GeoLife (GPS-based) & & & Mode Detection \\ \hline
% \cite{liu2018trajgans} & & & & \\ \hline
\cite{rao2020lstm} & New York data (GPS-based) & POI matching & LSTM-TrajGAN & Synthetic trajectory generation \\ \hline
\cite{chen2021trajvae} & GAOTONG (GPS-based) & Map matching & TrajVAE & Synthetic trajectory generation \\ 
\hline\hline
\end{tabular}
}
\end{center}
\end{table}

\end{landscape}
\newpage

% Detection

% Prediction

% Generation & Reconstruction

%% vehicle trajectory data & collections

%% noise in raw vehicle trajectory data & need of preprocessing
%% preprocessing methods : denoising, 1g, tessellation matching

% \subsection{Model-based Approaches}

% \subsection{Data-driven Approaches}
% \subsubsection{Machine Learning Approaches}
% \subsubsection{Deep Learning Approaches}

% \section{Deep Learning Models}
% \subsection{Sequence Models}
% \subsubsection{Recurrent Neural Networks}
% \subsubsection{Encoder-Decoder Recurrent Neural Networks}
% \subsubsection{Attention-based Recurrent Neural Networks}
% \subsubsection{Transformer}

% \subsection{Generative Adversarial Networks}

% \subsection{Imitation Learning}
% \subsubsection{Behavior Cloning}
% \subsubsection{Inverse Reinforcement Learning}
% \subsubsection{Generative Adversarial Imitation Learning}

% Despite its potential to improve the throughput, spatial domain diversity was not fully considered in the studies of original CCRNs. Utilizing the spatial domain for the communications, the concept of MIMO has been adopted in many cases to increase the wireless capacity

% : Transformer-based Map Matching Algorithm and Training Technique based on Synthetic Data

% \newpage
% \section{Deep Learning Models}

% \subsection{Recurrent Neural Networks}
% \subsection{Transformer}
% \subsection{Imitation Learning}

\chapter{Research Framework}\label{chapter:framework}

This chapter presents the research framework of this dissertation. There are numerous research problems in urban vehicle trajectory analytics as explored in Chapter \ref{chapter:prelim}. Among many research problems, this dissertation mainly focuses on the following two research topics: next location prediction, and synthetic trajectory generation. In this chapter, the definition of each research problem is presented, as well as challenges related to the research problem. After identifying the challenges, research approaches that this dissertation used to resolve the challenges in each research problem.

\section{Research Problems and Research Approaches}

\subsection{Next Location Prediction}
% Next location prediction problem is defined as a task of forecasting the next location of an individual vehicle based on the historical data. 
Next location prediction can be defined as forecasting the next location of an individual vehicle based on the historical data. Next location prediction has gained much attention from many researchers due to its applicability to many fields such as travel recommendation, location-based services, and location-aware advertisements. 

There are mainly two challenges in \textit{next location prediction} as follws:

\begin{itemize}
    \item \textbf{Design of a dense representation of temporal and spatial characteristics of urban vehicle trajectory} \\
    The mobility pattern in human mobility, including urban vehicle trajectories, is characterized by a high degree of regularity, which is mainly encoded in the temporal order of the visitation patterns \cite{song2010limits}. It is required to design a dense representation of temporal and spatial patterns embedded in the urban vehicle trajectory to predict the next location pattern. A proper representation of these spatiotemproal patterns makes it easier to understand the structurally related features in urban vehicle trajectory data.
    
    \item \textbf{Need for combining heterogeneous data sources to model multiple factors influencing next location prediction}\\
    Although the temporal order of the visitation patterns is mainly used for the next location prediction in many previous researches \cite{gambs_next_2012, ziebart2008navigate, choi2018network}, human drivers consider other factors in deciding where to go next and which route to choose. These external factors include traffic states, trip purposes, weather conditions, and social contacts \cite{luca2020deep}. As a result, it is required to combine heterogeneous data sources with the next location prediction.
    
\end{itemize}

To address these issues in next location prediction, this dissertation proposes three research approaches as follows:

\begin{itemize}
    \item \textbf{Spatial feature extraction via clustering-based Voronoi tessellation}
    Given massive amounts of vehicle trajectories, there will be an infinite number of possible data points used to describe all those trajectories as longitude and latitude coordinates are continuous in space. Also, RSU-based urban vehicle trajectories are sensitive to noise from the collection system, so that it is required to pre-process the raw data to use it for the next location prediction. As such, we partition an urban traffic network into smaller regions or \textit{cells} and express each urban vehicle trajectory in terms of a \textit{sequence of cells} that it has passed. In partitioning the network into cells, we use the method based on \cite{kim2016graph}. In this method, data points in all the trajectories are combined and clustered in space based on a desired radius, denoted by $R$, so that for each spatial cluster the distance between the centroid of the point cluster and its farthest member point is approximately $R$. The centroid of each point cluster is estimated by finding the mean of the data points within the cluster. Once the centroids of all point clusters are obtained, a Voronoi tessellation method is used to construct cell boundaries (Voronoi polygons) using the centroid points as seeds. Through clustering the data points in the urban vehicle trajectory dataset, partitioning the network and representing urban vehicle trajectories as cell sequences can be understood as a way of spatial feature extraction from urban vehicle trajectory data.
    
    \item \textbf{Recurrent Neural Networks to model spatiotemporal relationship}
    This study employs a deep learning method using Recurrent Neural Network (RNN) among various methods for sequence prediction. RNN \cite{hochreiter1997long,cho2014learning,chung2014empirical} is a deep neural network system designed to use sequential information. Unlike other traditional deep neural network models, which assume independence among all inputs (and outputs), RNN can capture temporal dependencies in sequential data. Thus, it is suitable for performing tasks that require memories of previous inputs. As a result, RNN can be used to model spatiotemporal relationships by using cell sequences that have aggregated spatial information of urban vehicle trajectory.
    
    \item \textbf{Attention mechanism to incorporate heterogeneous data source}
    Nowadays, drivers can easily access the network traffic state data via navigation apps in smartphones. The network traffic state is one of the most important factors when planning their journey and deciding the route. As a result, it is desirable to incorporate network-wide traffic state information into the next location prediction. Network-wide traffic state data is a heterogeneous data source compared to urban vehicle trajectories represented as cell sequences. As a result, it is required to design a systematic way to link the heterogeneous input to the next location prediction structurally. The attention mechanism \cite{bahdanau2014neural} can be used to resolve this issue. The attention mechanism allows the next location prediction model to concentrate on a certain part of the network traffic state input and use the information for the next location prediction.
\end{itemize}

Chapter \ref{chapter:nlp_rnn} and Chapter \ref{chapter:nlp_arnn} present the specific details on three research approaches. Chapter \ref{chapter:nlp_rnn} presents Recurrent Neural Network (RNN) model for urban vehicle trajectory prediction with spatial feature extraction via clustering-based Voronoi tessellation, and analyze the performance of the RNN model in both sequence level and aggregated region level. Chapter \ref{chapter:nlp_arnn} presents attention-based Recurrent Neural Network (ARNN) model which incorporates network-wide traffic state information into RNN model developed in Chapter \ref{chapter:nlp_arnn}.

% Next location prediction aims to develop model to predict the future location of a vehiclebased on the historical trajectory data. 

\subsection{Synthetic Trajectory Generation}
Synthetic trajectory generation can be defined as generating synthetic trajectories with realistic spatiotemporal mobility patterns based on the historical trajectory dataset. Although the sources and availability of urban trajectory data are increasing, most of the currently available trajectory datasets cover only a portion of all vehicles in the networks. As a result, some kind of augmentation method is needed to have full population of the vehicles. Also, urban vehicle trajectories contain personal information of individual drivers, such as the location of their homes and work places.  

There are mainly two challenges in \textit{synthetic trajectory generation} as follows:

\begin{itemize}
    \item \textbf{Capturing the temporal and spatial patterns of vehicle trajectory}\\
    Generating urban vehicle trajectories requires to understand the underlying distribution of the urban vehicle trajectories by capturing the temporal and spatial patterns in the dataset. As discussed in next location prediction, modeling urban vehicle trajectory requires a dense representation of spatiotemporal patterns in the input dataset. Therefore, it is important to find a suitable learning structure to reflect spatiotemporal characteristics in the synthetic trajectory generation.
    
    \item \textbf{Capturing both patterns as an individual and patterns as a group}\\
    The objective of synthetic trajectory generation is to generation urban vehicle trajectories that are similar to the real vehicle travel paths observed in a road traffic network. The "similarity" between the real vehicle trajectories and the synthetic vehicle trajectories can be defined from two different perspectives. First, the \textit{trajectory-level} similarity measures the similarity of an individual trajectory to a set of reference trajectories. For instance, the probability of accurately predicting the next locations---single or multiple consecutive locations as well as the alignment of the locations---are examples of trajectory-level similarity measures. Second, the \textit{dataset-level} similarity measures the statistical or distributional similarity over a trajectory dataset. This type of measure aims to capture how closely the generated trajectory dataset matches the statistical characteristics such as origin-destinations (OD) and route distributions in the real vehicle trajectory dataset. 
    
\end{itemize}

To address these issues in synthetic trajectory generation, this dissertation proposes three research approaches as follows:

\begin{itemize}
    \item \textbf{Generative Adversarial Imitation Learning to learn various patterns from the given dataset}\\
    We apply \textit{imitation learning} to develop a generative model for urban vehicle trajectory data. Imitation learning is a sub-domain of reinforcement learning for learning sequential decision-making behaviors or "policies." Unlike reinforcement learning that uses "rewards" as signals for positive and negative behavior, imitation learning directly learns from sample data, so-called "expert demonstrations," by imitating and generalizing the expert' decision-making strategy observed in the demonstrations. 
    Nowadays, the development of many generative models made it possible to capture the complex distribution of a dataset. Especially, deep generative models such as generative adversarial networks (GAN) \cite{goodfellow2014generative} show outstanding performance in reproducing images \cite{radford2015unsupervised}.
    Generative Adversarial Imitation Learning (GAIL) \cite{ho2016generative} is a variant of GAN applied in an imitation learning problem. 
    Let us consider an urban vehicle trajectory as a sequence of decisions for choosing road links along a travel path. GAIL can be applied to develop a generator that can reproduce synthetic data by imitating the decision-making process (i.e., driver' route choice behavior) demonstrated in the observed trajectory dataset.
    
    \item \textbf{Assuming partially observable MDP to understand spatiotemporal patterns from previous locations}\\
    GAIL, proposed by \cite{ho2016generative}, uses a combination of IRL's idea that learns the experts' underlying reward function and the idea of the generative adversarial framework. GAIL effectively addresses a major drawback of IRL, which is high computational cost. However, the standard GAIL has limitations when applied to the vehicle trajectory generation problem because it is based on the IRL concept that only considers a vehicle's current position as states in modeling its next locations \cite{ziebart2008maximum,ziebart2008navigate,zhang2019unveiling}, which is not realistic as a vehicle's location choice depends on not only the current position but also the previous positions. To overcome these limitations, this study proposes a new approach that combines a partially-observable Markov decision process (POMDP) within the GAIL framework. POMDP can map the sequence of location observations into a latent state, thereby allowing more generalization of the state definition and incorporating the information of previously visited locations in modeling the vehicle's next locations. 

    \item \textbf{Performance evaluation to assess both trajectory-level similarity and distributional similarity of datasets.}\\
    This study proposes a multi-level performance evaluation that includes both trajectory-level and dataset-level performance metrics to assess the model's performance. In the trajectory-level evaluation, we measure how similar each generated vehicle trajectory is to a real trajectory. Two widely used evaluation metrics in sequence modeling are used to evaluate this trajectory-level similarity: \textit{BLEU} score \cite{papineni_bleu:_2002}  and \textit{METEOR} score \cite{banerjee_meteor:_2005}. The statistical similarity between a generated trajectory dataset and a real trajectory dataset is assessed in the dataset-level evaluation. Many aspects of a dataset can be considered for statistical similarities, such as the distributions of trajectory length, origin, destination, origin-destination pair, and route. Among these variables, \textit{route distribution} is the most difficult to match since producing a similar route distribution requires matching all other variables, including the lengths, origins, and destinations of vehicle trajectories in a real dataset. As such, we use a measure of route distribution similarity to evaluate dataset-level model performance.
    
\end{itemize}

Chapter \ref{chapter:trajgen} presents the specific details on both research approaches. Chapter \ref{chapter:trajgen} presents TrajGAIL; a generative adversarial imitation learning for generating urban vehicle trajectories. In TrajGAIL, the generation procedure of urban vehicle trajectories is formulated as an imitation learning problem based on Partially Observable Markov Decision Process (POMDP), which can effectively deal with sequential data, and this imitation learning problem is solved using GAIL, which enables trajectory generation that can scale to large road network environments.

% \section{Problem Definition}

% \chapter{Map Matching: Transformer based Map Matching Algorithm} \label{chapter:mapmatching}
% % \chapter{Map Matching: Deep Learning based Map Matching Algorithm and Training Technique based on Generated Dataset from Map Data }\label{chapter:mapmatching}

% \section{Introduction}
% \section{Methodology}
% \section{Performance Evaluation}
% \section{Conclusion}

\chapter{Next Location Prediction: Network-Wide Vehicle Trajectory Prediction in Urban Traffic Networks using Deep Learning}\label{chapter:nlp_rnn}\footnote{This chapter is based on following article: Choi, Seongjin, Hwasoo Yeo, and Jiwon Kim. "Network-wide vehicle trajectory prediction in urban traffic networks using deep learning." Transportation Research Record 2672.45 (2018): 173-184. \cite{choi2018network}}
\section{Introduction}

Large-scale mobility data that record detailed movements or trajectories of people and vehicles have become increasingly available in recent years. A trajectory in this study refers to a sequence of locations and the passage times describing the path that a vehicle follows along its journey. Applying data analytics on large-scale trajectory datasets, it is possible to discover complex patterns in human mobility and obtain deeper insights into travel behaviors and traffic dynamics, allowing road operators and transit agencies to identify opportunities to improve their systems. It is also possible to build predictive models for predicting movement patterns of travelers. In this study, we address the problem of predicting individual trajectories in an urban network using a data-driven approach based on deep learning.

Network-wide trajectory prediction aims to predict movements of individual vehicles across the network by predicting where each vehicle may be going next at a given time based on where it is now and how it got there. Viewing each trajectory as a sequence of locations, where locations can be defined at various spatial resolutions such as link-level (e.g., links and intersections) and region-level (e.g., areas and geographic subdivisions), the trajectory prediction can be considered as a sequence prediction problem, in which we wish to predict the next location in a sequence given the previous locations visited.

Among various methods for sequence prediction, this study employs a deep learning method using Recurrent Neural Network (RNN). RNN \cite{hochreiter1997long,cho2014learning,chung2014empirical} is a deep neural network system designed to make use of sequential information. Unlike other traditional deep neural network models, which assume independence among all inputs (and outputs), RNN can capture temporal dependencies in sequential data and thus is suitable for performing tasks that require memories of previous events. As such, RNN showed great performance in learning patterns in sequential data, particularly in the areas of language modeling such as auto-texting, text recommendation, speech recognition, and auto-translation system.

Trajectory prediction has similarities with the problem of predicting words and sentences in language modeling. In sentence prediction, a large number of sentence sets are collected and words are extracted. A language model is trained to learn word sequence patterns from sentence data and predicts a certain word that would come next given a sequence of words. In the context of movement data, we can view a set of locations in a network as a set of words in a dictionary and a trajectory (location sequence) as a sentence (word sequence), thereby linking the problem of predicting the next location in a trajectory to that of predicting the next word in a sentence. Motivated by this idea, this study applies an RNN method that has been successfully applied in sequence prediction to solve our problem of trajectory prediction by adapting it to urban movement data.

\section{Related Researches}
Predicting future trip based on previously visited locations has been widely studied in terms of predicting location where the user will visit next \cite{noulas2012mining,gambs_next_2012,mathew2012predicting}, the location where the user ends the trip \cite{krumm2006predestination,krumm2007predestination,horvitz2012some,xue2015solving,ziebart2008navigate,marmasse2002user}, and the location where the user will visit in the next time interval \cite{hawelka2015collective,alhasoun2017city,lu2013approaching,calabrese2010human,zhao2017mobility}. The first and the second ones understand individual trips as a sequence of locations, similar to the one explained previously. The latter one predicts the location that user will visit in the next time interval, which is usually set as an hour. This may be widely applicable since this adapts temporal characteristic of mobility, however, such task requires frequent updates of user’s actual location. Also, one of the major problems is that most of the trips end in less than 30 minutes to 1 hour in the urban area as the majority of the trips are for commuting or visiting a commercial area. Therefore, it is hard to distinguish if the users are still traveling or staying. In fact, these methods are modeled to solve more macroscopic trips than city-scaled problems.  For example, in \cite{zhao2017mobility}, the authors presented N-gram model to predict the trip time, entry and exit station. They used the Oyster entry and exit records data collected from the London Underground, Overground and National Rail.

There are several previous studies that used machine-learning models to predict the future location or the destination of a trip. One of them, \cite{gambs_next_2012}, used Mobility Markov Chain to predict the next location of an individual. The research was based on the observations of individual’s mobility so that the model must be specified by each individual. In this research, however, we aim to build trajectory prediction model for more general purposes in both microscopic and macroscopic perspectives, therefore more generalized model with aggregated data is used. Also, in \cite{mathew2012predicting}, Hidden Markov Model is used to predict pedestrian movement by using GeoLife dataset. Hidden Markov Model computes latent state at each sequence, which maximizes the likelihood of the existence of input sequence. Usually, the number of latent states or the number of clusters is given and Hidden Markov Model calibrates the transition matrix among the latent states and emission probability to decode latent states to observable sequences. 

Some previous studies also tried to use Artificial Neural Networks in trajectory prediction. Recent work by \cite{de2015artificial} includes a study on the prediction of taxi destination by using Multilayer Perceptron (MLP). They represented the destination as a linearly weighted combination of predefined destination clusters. The result showed that the overall distance error is considerably negligible, however, it is pointed out that it is difficult to predict unpopular destinations.

With the recent development of Deep Neural Network, including RNN models, and computation powers, there have been some researches in transportation field to predict the microscopic vehicle location for autonomous vehicles \cite{kim2017probabilistic} and also predict mobility sequences \cite{endo2017predicting,liu2016predicting}. The research in \cite{kim2017probabilistic} used RNN with Long Short Term Memory (LSTM) to predict the vehicle movement in front of a subject vehicle. Endo et al. in \cite{endo2017predicting} used RNN to predict destination. In this research, trajectory sequence is represented as a sequence of locations in a discretized grid space which is an arbitrary network partitioning. Therefore, in this study, we used network partitioning method based on the vehicle trajectory data.

\section{Methodology}
\subsection{Representing Urban Vehicle Trajectories as Cell Sequence Data}

\begin{figure}[!ht]
  \centering
  \includegraphics[width=\textwidth]{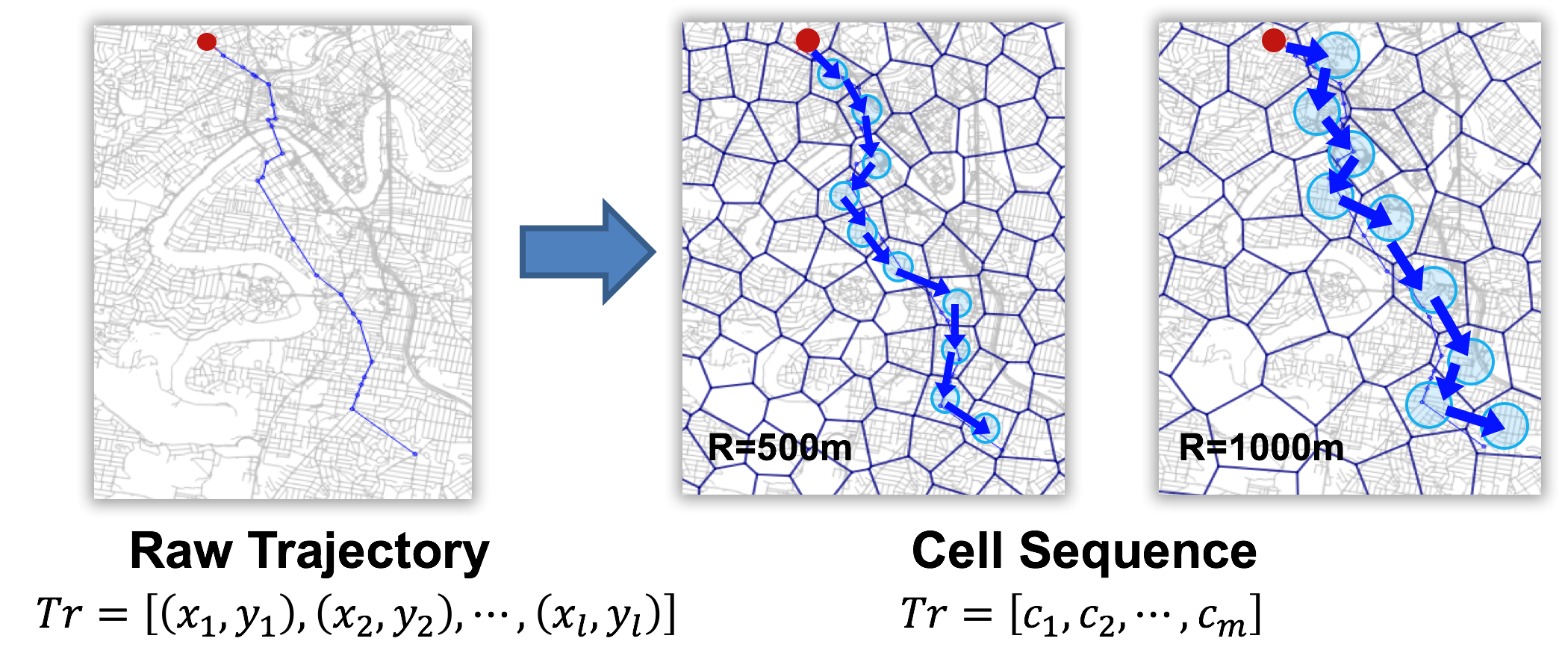}
  \caption{Representing urban vehicle trajectory as cell sequence}
  \label{fig:trbrnn_fig1}
\end{figure}

Let $Tr=[(x_1,y_1 ),(x_2,y_2 ),⋯,(x_l,y_l )]$ represent an urban vehicle trajectory consisting of $l$ data points, where data point $(x_i,y_i)$ represents the longitude and latitude coordinates of the vehicle’s $i^th$ position. Typical trajectory datasets include the timestamp information for each data point, but we will only consider the spatial path of each trajectory in this study as our goal is to predict the next location given the previous path regardless of the time-of-day and travel time along the journey. Incorporating such temporal information in movement prediction will be explored in future research. Given massive amounts of vehicle trajectories, there will be an infinite number of possible data points that are used to describe all those trajectories as longitude and latitude coordinates are continuous in space. To apply the concept of sentence prediction in language modeling to trajectory prediction, however, it is necessary to define a finite set of locations with which all the trajectories can be expressed in a similar way to defining word vocabulary in sentence prediction. As such, we partition an urban network into smaller regions or cells and express each trajectory in terms of a sequence of cells that it has passed. In partitioning the network into cells, we use the method based on \cite{kim2016graph}. In this method, data points in all the trajectories are combined and clustered in space based on a desired radius, denoted by $R$, so that for each spatial cluster the distance between the centroid of the point cluster and its farthest member point is approximately $R$. The centroid of each point cluster is estimated by finding the mean of the data points within the cluster. Once the centroids of all point clusters are obtained, a Voronoi tessellation method is used to construct cell boundaries (Voronoi polygons) using the centroid points as seeds.

Given $N$ cells in the network, trajectory $Tr$ can be mapped onto the underlying cells and expressed as in terms of cell sequence $[c_1,c_2,⋯,c_m]$, where $c_j$ is the index of the $j^th$ visited cell $(1\leq c_j \leq N)$ within trajectory $Tr$. Figure \ref{fig:trbrnn_fig1} illustrates the process of converting a raw trajectory to cell sequences. The length of cell sequence, $m$, can be smaller than the original trajectory length, l, if two or more consecutive trajectory data points belong to the same cell (i.e., $m\leq$). Also, a fewer number of cells would be needed to represent the same trajectory as the cell radius, $R$, increases. In addition to the cell sequence, two special tokens $StartCode$ and $EndCode$ are added to the front and the back of the cell sequence to indicate the start and the end of the trip, respectively, where value 0 is used for $StartCode$ and $N+1$ is used for $EndCode$. As a result, the original trajectory Tr is converted to the following cell sequence form:

% Eq (1)
\begin{equation}
\label{eq:rnn_seqdef}
\begin{split}
Tr = \Big[(x_1,y_1),(x_2,y_2),\cdots,(x_i,y_i) \Big] \rightarrow \Big[ StartCode, c_1, c_2, \cdots EndCode \Big]
\end{split}
\end{equation}

\newpage

\subsection{Predicting Urban Vehicle Trajectory using Recurrent Neural Networks}

\begin{figure}[!b]
  \centering
  \includegraphics[width=0.7\textwidth]{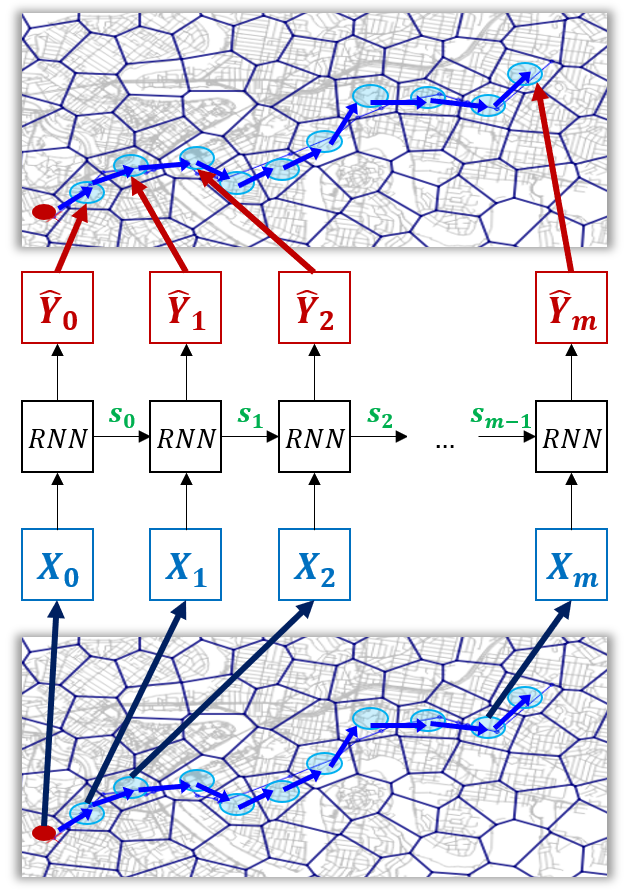}
  \caption{Structure of RNN model}
  \label{fig:trbrnn_fig2}
\end{figure}

Using the information from the previous locations is one of the key ideas to design the model structure. The conventional methods which do not consider or partially consider the previous locations fail to successfully predict the next location. As a result, it is required to design a proper structure to incorporate previous locations in next location prediction
Figure \ref{fig:trbrnn_fig2} shows the overall structure of a GRU-based RNN model. The model consists of a sequence of connected GRU layers, where each layer $i$ represents a function unit called GRU that receives input ($X_i$) and a hidden state or “memory” from the previous GRU layer ($s_{i-1}$) and produces output ($\hat{Y}_i$) which is the prediction of the correct label ($Y_i$). 
At the beginning and the end of a cell sequence, $StartCode$ and $Endcode$ is added to indicate the start and the end of the trip respectively. As a result, the number of GRU layers correspond to the length of the cell sequence of interest plus one, where each layer performs the operation of predicting one cell based on its previous cells. For instance, if we consider a trajectory that visits $m$ cells, i.e., $Tr =  \Big[ StartCode, c_1, c_2, \cdots EndCode \Big]$, the RNN model is set to have $m+1$ GRU layers, where input sequence ($X$) and the associated label sequence ($Y$) for training the model are coded in the following format:

% Eq (2)
\begin{equation}
\label{eq:rnn_trajdef}
\begin{split}
& X = \Big[ X_0, X_1, \cdots, X_m \Big] = \Big[ StartCode, c_1, c_2, \cdots, c_m \Big]\\
& Y = \Big[ Y_0, Y_1, \cdots, Y_m \Big] = \Big[ c_1, c_2, \cdots, c_m, EndCode \Big]
\end{split}
\end{equation}

Inside each GRU, there are two functions called reset gate (r) and update gate (z), where the reset gate decides how to combine new input with memory from previous computations and the update gate decides how much memory to keep from previous computations \cite{chung2014empirical,noulas2012mining}. The mathematical expressions for the operation taking place within the GRU of the $i^th$ layer are presented as follows:

% Eq (3)
\begin{equation}
\label{eq:rnn_gru}
\begin{split}
& z=σ(X_i U^z+s_{i-1} W^z ) \\
& r=σ(X_i U^r+s_{i-1} W^r ) \\
& h=\tanh{(X_i U^h+(s_{i-1} \circ r) W^h} \\
& s_i=(1-z)\circ h+z\circ s_(i-1) \\
& \hat{Y}_i = softmax(V \circ s_i) \\ 
\end{split}
\end{equation}

In Eq. (\ref{eq:rnn_gru}), $\circ$ represents element-wise multiplication between two vectors and σ represents sigmoid function to limit function output between 0 and 1. $X_i$ represents a one-hot vector of size N+1 (N cells + Start_code) indicating the cell visited at the ith position of the cell sentence (or indicating the start of the trip in case of Start_code) and $\hat{Y}_i$ represents a real-valued vector of size N+1 (N cells + End_code), where each value in $\hat{Y}_i$ represents the probability of visiting each cell (or the probability of terminating the trip in case of End_code) at the (i+1)th position of the sequence. $s_(i-1)$ denotes the memory from previous computations or the hidden state computed in GRU. U and W are matrices of parameters used inside GRU units, and V is a vector of parameters used to translate internal hidden state to vector of probability of each cell being visited in the next step $\hat{Y}_i$, where the superscripts on U and W indicate the functions these parameters are used in Eq. (\ref{eq:rnn_gru}) basically shows a series of computations to obtain Y from X, entailing computing the probability of each cell being visited in the next step $\hat{Y}_i$ based on the hidden state at the current step ($s_i$), where this hidden state ($s_i$) is in turn computed based on the current cell ($X_i$) and the hidden state transferred from the previous step or the memory of the previous visited cells ($s_(i-1)$). 

Next, we define the loss function that calculates the loss or error in a model prediction, which we will aim to minimize during training. The cross-entropy loss function is used in this study as shown in Eq. (4). By treating the full trajectory (hence its cell sequence Y) as one training case, the cross-entropy loss (L) is calculated by summing the errors at each step (cell) in the sequence over the entire trajectory as follows: 

% Eq (4)
\begin{equation}
\label{eq:rnn_bceloss}
\begin{split}
& L(Y, \hat{Y}) = \sum_{i \in (1,\cdots,m+1)} - \frac{1}{N} \sum_{n \in N} (Y_{i,n} \log \hat{Y}_{i,n} + (1-Y_{i,n} \log (1-\hat{Y}_{i,n}))) \\
\end{split}
\end{equation}

\noindent
where $m$ is the cell sequence length; $N$ is the number of cells; $Y_(i,n)$ is the binary label set to 1 if cell n is visited at the (i+1)th step in the sequence or 0 otherwise; and $\hat{Y}_{i,n}$ denotes the probability of cell n being visited at the (i+1)th step in the sequence. 

To train the RNN model, a large number of cell sequence data are prepared in the format shown in Eq. (\ref{eq:rnn_trajdef}) and fed into the RNN. Since the lengths of input sequences are different, the number of GRU layers in the RNN model needs to be dynamically changed. The RNN model scans one input sequence at a time and adjusts the number of GRU layers so that the number of layers becomes the length of the given input sequence ($m$) plus one. As a result, every time when a new input sequence with length m is fed into the model, the model constructs RNN with $m+1$ GRU layers.

The model then calculates cross-entropy loss ($L$) between the correct label ($Y$) and predicted label probability based on the current parameter ($\hat{Y}$). The model uses the stochastic gradient descent (SGD) method (see, e.g., \cite{bengio2013advances} and \cite{pascanu2013difficulty}) to update the model parameters in the direction of decreasing the loss. This process is repeated until the parameters converge. Once the model is trained and the parameters are fixed, the model can be used to predict the next cell for any given sequence of previous cells.

\section{Model Performance Evaluation}
\subsection{Data}
To evaluate the performance of the RNN-based trajectory prediction model, a case study was designed using vehicle trajectories collected from the Bluetooth sensors in Brisbane, Australia, provided by Queensland Department of Transport and Main Roads (TMR) and Brisbane City Council (BCC). The Bluetooth sensors installed in state-controlled roads and city intersections detect Bluetooth devices (e.g., in-vehicle navigation systems and mobile devices) passing the sensors and record their passage times. By tracking the identifier of each Bluetooth device, the trajectories of individual vehicles (or Bluetooth devices) can be constructed, where each trajectory represents a time-ordered sequence of Bluetooth sensor locations that a given vehicle passed. Vehicle trajectories containing a resting period (time without moving) over 1 hour are considered to have multiple trips and they are separated into multiple trajectories. For this case study, we used the data from one day on 1 March 2016 for training and the data from one day on 8 March 2016 for testing. Each dataset contains approximately 350,000 trajectories per day.

Using the method described in Figure \ref{fig:trbrnn_fig1}, the Brisbane network is partitioned into cells using three different cell sizes 300m, 500m, and 1000m in terms of the desired cell radius ($R$). The network is represented as 5712 cells, 2204 cells, and 319 cells under $R=300m$, $R=500m$, and $R=1000m$, respectively. For each cell network, trajectories are mapped onto the underlying cells and represented as the associated cell sequences.

\begin{figure}[!ht]
  \centering
  \includegraphics[width=\textwidth]{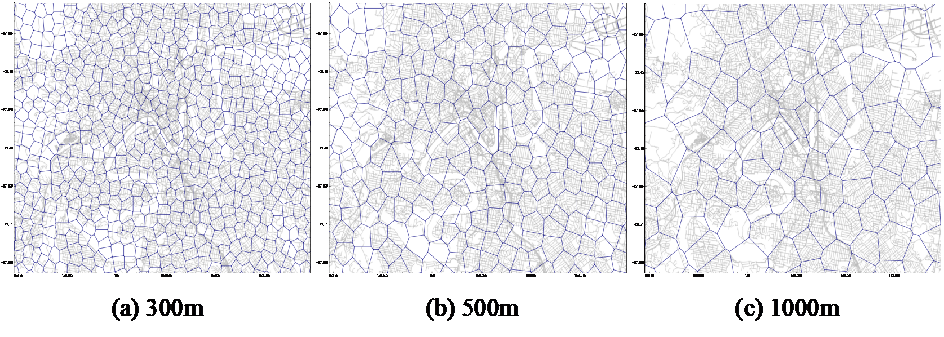}
  \caption{Cell boundaries with different desired radius ($R$)}
  \label{fig:trbrnn_fig3}
\end{figure}

\subsection{Basecase Model}
To better understand the performance of the proposed RNN-based model, a simple statistical model was designed as a base-case model for a comparison. In predicting the next cell in a cell sequence, the base case model, which we will call Transition matrix method (TRN), relies on the transition matrix that describes the probability of going to a particular cell from the current cell. The transition probabilities are estimated based on the historical data, where the transition probability from cell $i$ to cell $j$, denoted by $p_{i \rightarrow j}$, is determined by computing the fraction of the outgoing flows from cell $i$ to cell $j$ given the total outgoing flows from cell i as follows:

% Eq (5)
\begin{equation}
\label{eq:rnn_trn}
\begin{split}
& p_{i \rightarrow j} = \frac{f_{i \rightarrow j}}{\sum_{n=1}^N f_{i \rightarrow n}} \\
\end{split}
\end{equation}

\noindent
where $f_{i\rightarrow j}$ is the inter-cell flow from cell i to cell j (i.e., vehicle flows passing cell $i$ and cell $j$ consecutively) and N is the number of cells in the network. In predicting where a vehicle will go next given a sequence of cells the vehicle has passed so far, TRN determines the next cell only based on the last cell of the input sequence and does not use any information about the vehicle’s travel history. The key difference between RNN and TRN is thus that the former uses the memory of the previous cells visited and incorporates sequential characteristics in predicting the next cell, while the latter is memoryless and the next cell only depends on its immediate predecessor.

\section{Results}
\subsection{Cross-entropy Loss and Validation}

\begin{figure}[!ht]
  \centering
  \includegraphics[width=\textwidth]{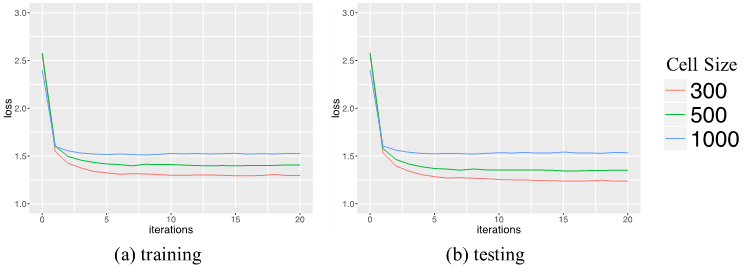}
  \caption{The result of cross-entropy loss (a) with training dataset (b) with testing dataset}
  \label{fig:trbrnn_fig4}
\end{figure}

\begin{figure}[!ht]
  \centering
  \includegraphics[width=\textwidth]{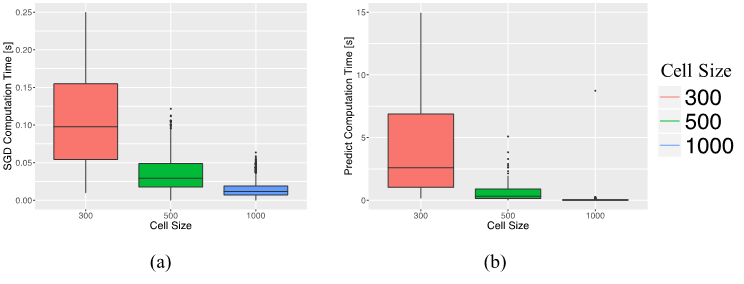}
  \caption{Computation time of (a) Stochastic Gradient Descent (b) computation to predict next location}
  \label{fig:trbrnn_fig5}
\end{figure}

Figure \ref{fig:trbrnn_fig4} shows the cross-entropy loss calculated during model training, plotted for different cell sizes. At each iteration, the parameters of the RNN model are updated based on the training dataset and the average loss ($\Bar{L}$) is calculated for both training and testing datasets under the given parameters as shown in Figure \ref{fig:trbrnn_fig4} (a) and (b), respectively. The RNN parameters converge after 5 iterations in all cases and the loss decreases for both training and testing datasets, indicating that the model is trained properly and not overfitting to the training data. Comparing plots across different cell sizes, the case with cell radius of 300m produces the smallest loss. This may suggest that, the smaller the cell size, the better the RNN model recognizes sequential patterns and, hence, the easier it is for the model to predict the next cell. However, as shown in Figure \ref{fig:trbrnn_fig5}, the computation time exponentially increases as the cell size decreases. As such, there is a trade-off between model accuracy and computation cost and this should be incorporated when determining the cell size. The computing capacity used for this chapter is as follows: Intel Core i7-7700 CPU @ 3.60 GHz with RAM of 64GB and NVIDIA GeForce GTX 1080 Ti.

\subsection{Cell Visit Count and Inter-cell Flow}
One of the important applications of trajectory prediction is to anticipate how many vehicle will use a particular region in the network by predicting individual vehicle’s moving paths. For instance, when a major disruption (e.g., construction, special event) is anticipated for a certain region, road operators will want to identify who will travel to or pass through that region during the day to identify the potential impact of the disruption and provide the relevant travelers with more personalized and targeted information. To assess model performance in this regard, this section aims to evaluate the sequence prediction accuracy in term of cell-level (area-level) aggregated measures, namely cell visit count and inter-cell flow. Cell visit count measures the number of unique travelers who visit a particular cell during a given day and inter-cell flow (or cross-boundary flow) measures the total daily volume of vehicle traffic from one cell to another between neighboring cells. 

We first generate 700,000 synthetic trajectories (in terms of cell sequences) using RNN and TRN models for three different cell sizes (R=300m, 500m and 1000m). For both RNN and TRN models, we only give origin cells, which are randomly sampled from the historical data (training set), and let the models determine the remaining sequences. Based on the 700,000 cell sequences generated by each model for each cell size, we compute the cell visit counts and inter-cell flows for all cells. As a ground truth, we also obtain the cell visit count and inter-cell flow measures from real data using the training set. To compare the measures across these three cases (RNN, TRN, and real data), the measures are normalized with the number of trajectories. To exclude abnormal trajectories, we impose the maximum trajectory length of 50 km, which is roughly equivalent to the longest straight-line distance covering the Brisbane network. To reflect this condition, the maximum number of cells in a sequence is set to 80 for R=300m, 50 for R=500m, and 25 for R=1000m in generating cell sequences. The historical data also contains unrealistically long cell sequences. These are presumed to be from taxi vehicles and ride-sharing vehicles that usually keep moving for a long period. Therefore, the limit is also applied to the historical data to discard long sequences.

A real-world trajectory often visits a certain cell more than once during its journey. It is important for a trajectory prediction model to mimic this behavior as realistically as possible because the number of cells in a sequence (sequence length) and the number of unique cells have different implications: the former gives information on traffic volume in a given cell while the latter gives information on the number of actual travelers who visit the cell. To measure the extent to which each cell sequence contains duplicate cells, we define cell re-visit ratio (D) as follows:

% Eq (6)
\begin{equation}
\label{eq:rnn_revisit}
\begin{split}
& D = \frac{m-m^u}{m} \times 100 \% \\
\end{split}
\end{equation}

\noindent
where $m$ is the length of a cell sequence in terms of the number of cells and $m^u$ is the number of unique cells excluding duplicate cells. The $D$ measure is a relative measure and the $D$ measure of RNN and TRN are compared with the $D$ measure obtained from the real data. On average, for R=300m, 500m and 1000m, real data show $D$=18.6\%, 20.2\%, and 17.3\% with $m$=26.83, 17.05, and 9.29 and $m^u$=21.85, 13.60, and 7.68, respectively. For trajectories generated by RNN, on average, D=25.7\%, 24.7\%, and 19.9\% for $R$=300m, 500m and 1000m, respectively (with $m$=18.38, 14.26, and 8.33 and $m^u$=13.66, 10.74, and 6.67). Trajectories generated by TRN produce the average values of D=62.8\%, 56.7\%, and 45.6\% for R=300m, 500m and 1000m, respectively (with $m$=21.61, 15.80, and 8.89 and $m^u$=8.03, 6.84, and 4.84). In terms of average sequence length ($m$), model-generated trajectories are on average shorter than real-world trajectories, showing that both models tend to end trips earlier than real-world trips. Between RNN and TRN, the average lengths from TRN are closer to the real-world case than those from RNN. However, TRN-generated trajectories contain a large number of duplicate cells as can be seen from very high cell re-visit ratios ($D$), resulting in unrealistic movement tendency that a trajectory repeatedly goes back and forth between few cells instead of progressing toward its destination. On the other hand, RNN produces the cell re-visit ratios that are similar to the real data, suggesting that the ability to incorporate “memory” of the previously visited cells in RNN can prevent such an unrealistic “memoryless” behavior from happening.

\begin{figure}[!ht]
  \centering
  \includegraphics[width=\textwidth]{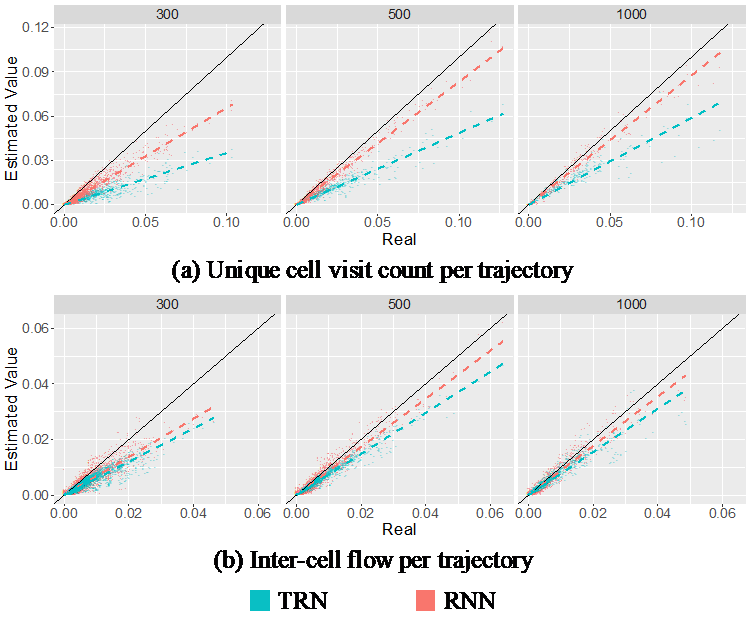}
  \caption{The result in aggregated region level. (a) Unique cell visit count per trajectory (b) Inter-cell flow per trajectory}
  \label{fig:trbrnn_fig6}
\end{figure}

The Figure \ref{fig:trbrnn_fig6} shows the estimated cell visit count and inter-cell flow for TRN (blue dots) and RNN (red dots) with respect to the corresponding real-world measures from the historical data, where all measures are normalized with the number of trajectories. Each data point on the plots in Figure \ref{fig:trbrnn_fig6}(a) represents the cell visit count for each cell and each data point in Figure \ref{fig:trbrnn_fig6}(b) represents the inter-cell flow for each neighboring cell-pair. Linear regression lines (dashed lines) are also shown, where the closer the regression line is to the 45-degree line (y=x), the closer the model prediction is to the reality. In all cases, the regression lines from RNN are closer to the 45-degree line than those from TRN, suggesting that RNN predicts cell-level aggregated measures more accurately than TRN.

\begin{figure}[!ht]
  \centering
  \includegraphics[width=\textwidth]{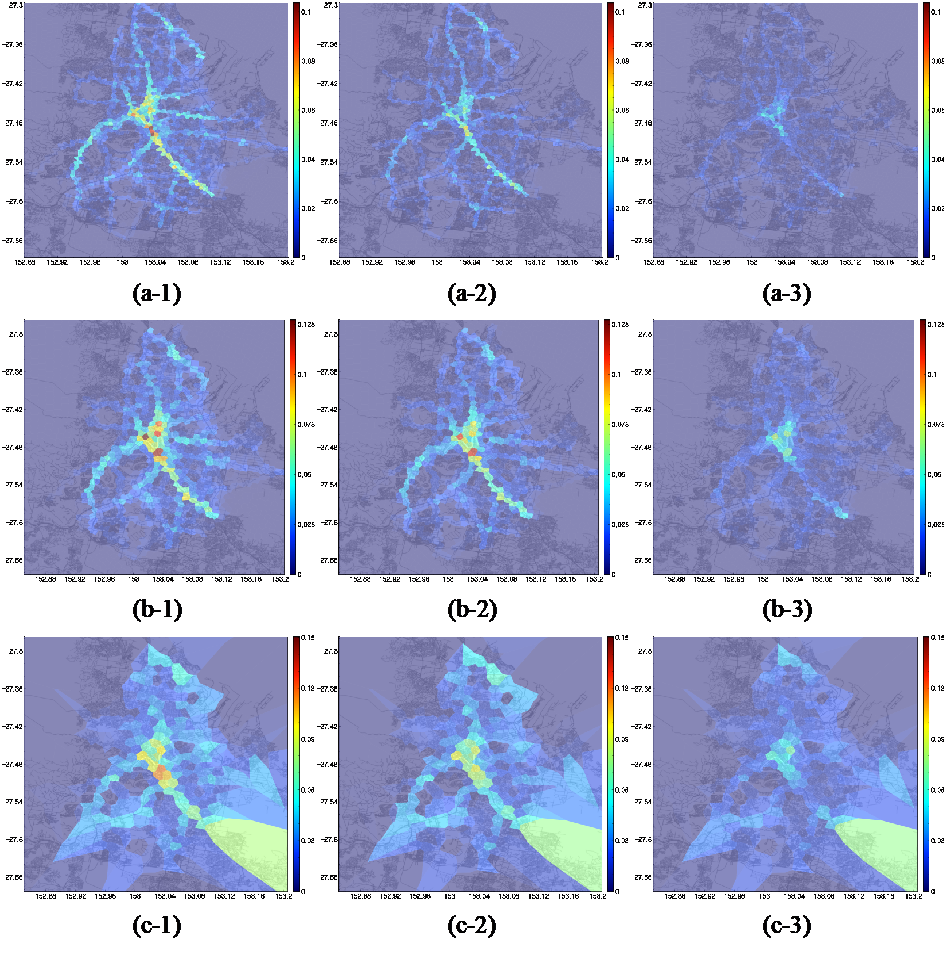}
  \caption{Spatial distribution of unique cell visit count per trajectory}
  \label{fig:trbrnn_fig7}
\end{figure}

For cell visit count, we create color maps to visualize the spatial distribution of cell measures across the Brisbane network. Figure \ref{fig:trbrnn_fig7} presents nine color maps created using the data points in Figure \ref{fig:trbrnn_fig6}(a), where the results can be compared across different trajectory datasets (by column) and different cell sizes (by row). From the maps, it is clear that RNN (center) reflects the flow magnitudes and spatial patterns in the real data (left) much closely than TRN (right) does. 

\subsection{Sequence Prediction}

\begin{figure}[!ht]
  \centering
  \includegraphics[width=\textwidth]{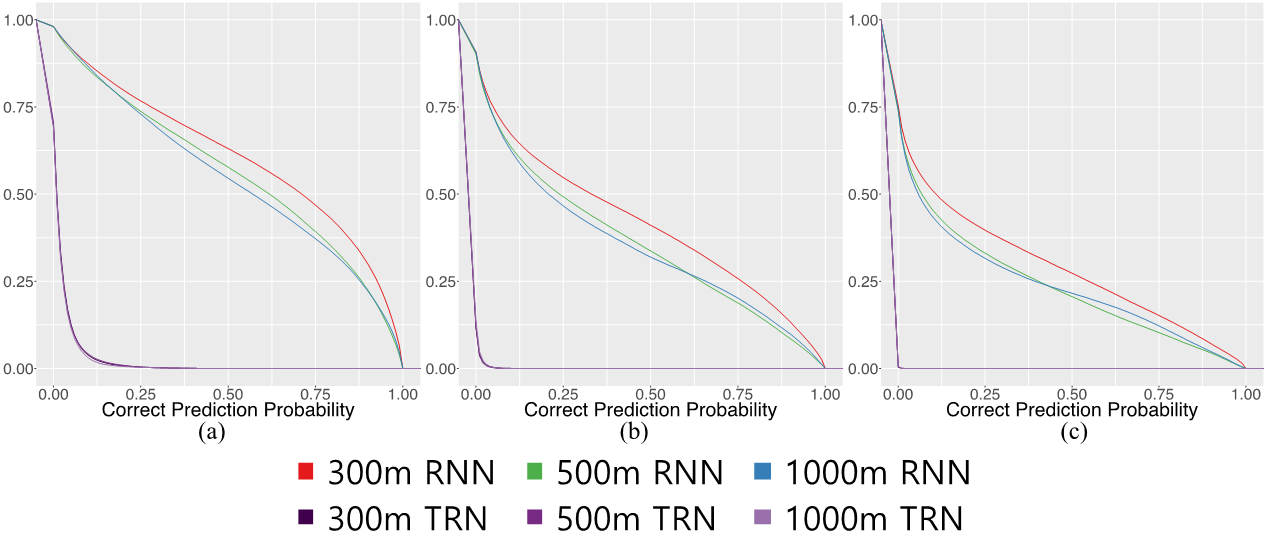}
  \caption{The result in individual sequence level (aggregated)}
  \label{fig:trbrnn_fig9}
\end{figure}

\begin{figure}[!ht]
  \centering
  \includegraphics[width=\textwidth]{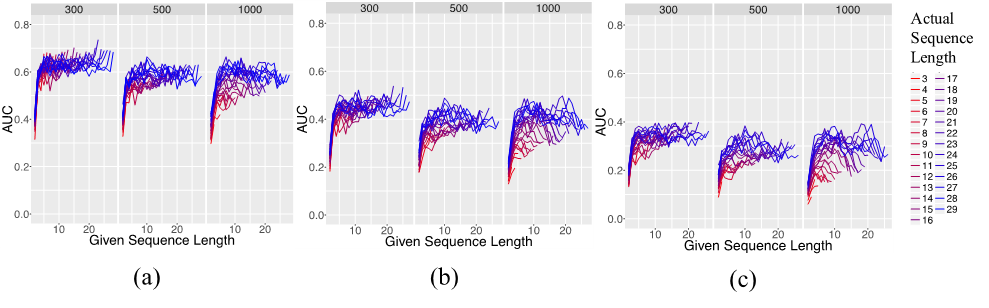}
  \caption{The result in individual sequence level (sensitivity analysis)}
  \label{fig:trbrnn_fig10}
\end{figure}

In this section, we focus on a more direct performance measure of cell sequence prediction, namely correct prediction probability (CPP_k), which represents the probability of correctly predicting the next k cell(s) given previously visited cells. The procedure for obtaining CPP_k measures is as follows: 

% \begin{table}[!ht]
%   \centering
%   \includegraphics[width=\textwidth]{img/trbrnnfig8.png}
% %   \caption{The model framework of TrajGAIL}
%   \label{fig:trbrnn_fig7}
% \end{table}

For a prediction scenario, we consider three scenarios of predicting next one cell (k=1), next two cells (k=2), and next three cells (k=3), respectively. With three k values, two models (RNN and TRN) and three cell sequence databases (one for each cell size R), a total of 18 cases are generated, each of which produces the distribution of $CPP_k$ over all combinations of \textit{actual sequence length} (m) and \textit{given sequence length} (i).

Figure \ref{fig:trbrnn_fig9} (a)-(c) present the $CPP_k$ distributions for $k=1,2,⋯,3$, respectively, in terms of the complementary cumulative distribution function ($CCDF$). The $CCDF$ of $CPP_k$, denoted by F ̅_(CPP_k ) (x), is defined as the probability that CPP_k is greater than x, i.e., $\Bar{F}_{CPP_k} (x) = P(CPP_k > x) = 1-F_(CPP_k)$
% F ̅_(CPP_k ) (x)=P(CPP_k>x)=1-F_(CPP_k ) (x),
, where $F_(CPP_k ) (x)$ is the cumulative distribution function (CDF). For instance, in Figure \ref{fig:trbrnn_fig9} (a), the value of $\Bar{F}_{CPP_k} (x=0.75)$ for 1000m RNN (blue curve) is approximately 0.375, meaning that 37.5\% of the CPP_1 values produced via Figure 6 are greater than 0.75, i.e., RNN predicted the next cell correctly 35\% of the time with the probability higher than 0.75. Given this interpretation, we can see that, the closer the $\Bar{F}_{CPP_k} (x)$ curve is toward the top right corner (1,1) and the farther the curve is away from the bottom left corner (0,0), the better the prediction performance is. The area under curve (AUC) of $\Bar{F}_{CPP_k} (x)$, thus, provides a good summary metric quantifying the prediction performance described by ) with a single number. The value of AUC of $\Bar{F}_{CPP_k} (x)$ varies from 0 to 1, where 1 occurs when the curve passes (1,1) and 0 occurs when the curve passes (0,0). 

As shown in Figure \ref{fig:trbrnn_fig9}, in all cases, RNN performs much better than TRN as the $\Bar{F}_{CPP_k} (x)$ curves for RNN are always above those for TRN, where the curves of RNN predicting one or two consecutive cell sequences are convex toward (1,1) and those of TRN are concave toward (0,0). In case of predicting three consecutive cell sequences, although both curves for RNN and TRN are concave towards (0,0), RNN is still able to predict cell sequence where correct prediction probability of TRN at this point reaches zero at all points. For the case of predicting next one cell in Figure \ref{fig:trbrnn_fig9} (a), the average AUC of $\Bar{F}_{CPP_k} (x)$ across the three cell sizes is 0.0198 for TRN and 0.5096 for RNN. When it comes to predicting multiple consecutive cells (k>1), the performance difference between RNN and TRN becomes substantial as shown in Figure \ref{fig:trbrnn_fig9} (b)-(c). The $\Bar{F}_{CPP_k} (x)$ curves for TRN exhibit a sharp drop at very low values of x, resulting in the AUC values being nearly zero. On the other hand, the $\Bar{F}_{CPP_k} (x)$ curves for RNN still show relatively high probabilities of having high CPP_k. For instance, for k=3, there are still approximately 10-15\% of cases where RNN predicts the next three cells correctly with the probability higher than 0.75 (see $\Bar{F}_{CPP_k} (x=0.75)$) in Figure \ref{fig:trbrnn_fig9} (c)). Overall, the average AUC of $\Bar{F}_{CPP_k} (x)$ is 0.2849 for k=2 and 0.1585 for k=3.

Next, we take a closer look at the prediction results from the RNN models, focusing on the impact of original sequence length (m) and given sequence length (i) on the prediction performance expressed by AUC. Within each combination of k and cell size R, we split the associated CPP_k observations into different m and i groups. We then construct a separate CCPF curve for each group to compute AUC. Figure \ref{fig:trbrnn_fig10} shows six groups of AUC plots, where each plot contains a set of AUC curves and each AUC curve represents a set of AUC values obtained for predicting cell sequences of certain length m. The x-axis represents given sequence length i in predicting the next k cells. Some observations from Figure \ref{fig:trbrnn_fig10} are summarized as follow:

\begin{itemize}
	\item 	From each AUC curve (given a fixed m), AUC increases sharply as i increases up to a certain point (e.g., i=5 or 6), but remains stable afterward. This means that until a certain point, giving a longer initial sequence helps predicting the next cells (e.g., predicting the 3rd cell given two previous cells is easier than predicting the 2nd cell given one previous cell). Beyond that point, however, the length of initial sequence does not significantly affect prediction performance.

	\item 	From each AUC plot (across different m), AUC increases as m increases. This indicates that, even when the same number of cells are given, the prediction performance depends on the length of the actual cell sequence (e.g., predicting the 3rd cell in a sequence of 10 cells is easier than predicting the 3rd cell within a sequence of 5 cells).

	\item 	Across AUC plots (across different k), AUC decreases as k increases, which is expected since predicting further steps toward the future is more difficult than predicting immediate next step.

    \item	Across AUC plots (across different R), AUC does not significantly change with different cell sizes. There is, however, a tendency of decreasing AUC and increasing variance in AUC values when spatial resolution is low (e.g., R=1000m).

\end{itemize}

\section{Conclusion}

The overall goal of this research is to leverage massive amounts of urban movement data, which become increasingly available nowadays, to better understand city mobility dynamics and enhance the design and operations of transportation systems. Of particular interest is the ability to predict individual vehicles’ movements—at least in terms of a sequence of aggregated spatial locations—and hence anticipate the flow of vehicles at a given location and time more accurately. This study showed a promising direction toward achieving this ability by applying deep learning with Recurrent Neural Networks (RNN) on vehicle trajectory data. As a way to represent complex vehicle trajectories as simpler location sequences, this study proposes a method to partition the network into cells so that entire vehicle movements can be expressed in terms of combinations of a finite set of cells. Mapping trajectories onto cells not only reduces computational complexity but also allows working with multi-source and multi-resolution trajectories. We test different cell sizes and provide discussions on the impacts of cell size on trajectory prediction performance. Using large amounts of Bluetooth vehicle trajectory data collected in Brisbane, Australia, this study trains a RNN model to predict cell sequences. We test the model performance by computing the probability of correctly predicting the next k consecutive cells. Compared to a base-case model that relies on a simple transition matrix, the proposed RNN model shows substantially better prediction results. We also test network-level aggregate measures such as total cell visit count and inter-cell flow and observe that the RNN model can replicate real-world traffic patterns.
In summary, the contribution of this study is the development of a novel network-wide trajectory prediction framework that entails (i) transforming raw trajectories into location sequence data using the proposed cell construction method, (ii) applying the RNN model to learn and predict trajectory sequence patterns by recognizing the similarity between trajectory sequence prediction and language modeling where RNN showed a great success, and (iii) proposing different performance measures, at both individual sequence level (e.g., correct prediction probability of predicting next cells in a sequence) and aggregated region level (i.e., cell visit count, inter-cell flow, network-wide usage pattern), to evaluate and demonstrate the application of the proposed model from different angles.

\chapter{Next Location Prediction: Attention-based Recurrent Neural Network for Urban Vehicle Trajectory Prediction}\label{chapter:nlp_arnn}\footnote{This chapter is based on following article: Choi, Seongjin, Jiwon Kim, and Hwasoo Yeo. "Attention-based recurrent neural network for urban vehicle trajectory prediction." Procedia Computer Science 151 (2019): 327-334. \cite{choi2019attention}}

\section{Introduction}
Recently, with abundance of various location sensors and location-aware devices, a large amount of location data are collected in urban spaces. These collected data are studied in the form of so-called moving object trajectory which is a trace of moving object in geographical spaces represented by a sequence of chronologically ordered locations \cite{zheng_trajectory_2015}. Of particular interest are urban vehicle trajectory data that represent vehicle movements in urban traffic networks. Such urban vehicle trajectory data offer unprecedented opportunities to understand vehicle movement patterns in urban traffic networks by providing rich information on both aggregate flows (e.g., origin-destination matrix and cross-sectional traffic volume) and disaggregate travel behaviours including user-centric travel experiences (e.g., speed profile and travel time experienced by individual vehicles) as well as system-wide spatiotemporal mobility patterns (e.g., origin-destination pairs, routing information, and network traffic state) \cite{kim2015spatial}. Previous studies have used urban vehicle trajectory data to perform travel pattern analysis \cite{kim2015spatial, yildirimoglu2018identification} and develop real-world applications such as trajectory-based bus arrival prediction \cite{zimmerman2011field} and trajectory-based route recommendation system \cite{yuan2011t}.

Among many applications of trajectory data mining \cite{mazimpaka2016trajectory}, this study focuses on trajectory-based location prediction problem. This problem concerns analyzing large amounts of trajectories of people and vehicles moving around a city to make predictions on their next locations \cite{noulas2012mining, gambs_next_2012, mathew2012predicting}, destinations \cite{krumm2006predestination,krumm2007predestination, horvitz2012some,xue2015solving, ziebart2008navigate}, or the occurrences of traffic related events such as traffic jams and incidents \cite{wang2016prediction}. In this study, we address the problem of predicting the sequence of next locations that the subject vehicle would visit, based on the information on the previous locations from the origin of the current trip and historical database representing the urban mobility patterns.

Trajectory-based location prediction is gaining increasing attention from both academia and industry because of its potential to improve the performance of many applications in multiple domains. One example is Location-based Service (LBS). LBS uses location data of service users and provide user-specific information depending on the locations of service users. Typical examples of LBS are social event recommendation, location-based advertising, and location-based incident warning system. The location prediction can be applied to predictively give information; for example, if a user’s next location is expected to be disastrous or congested, the service informs the user to change route. Furthermore, when it is not possible to continue to give service because the position of the user is lost due to sensor malfunctioning, predicting the locations of the user can temporally replace the role of positioning system and continue the service \cite{monreale2009wherenext, morzy2007mining}. Another example is the application on agent-based traffic simulators. Unlike traditional traffic simulators which consider traffic demand as input, an agent-based traffic simulator requires information on individual vehicle journey such as origin, destination, and travel routes \cite{martinez_agent-based_2015}. The result of vehicle location prediction can be used for real-time application of these agent-based traffic simulators. Vehicle location prediction can also be applied to inter-regional traffic demand forecasting. As the market of ride-sharing is continuously growing and Shared Autonomous Vehicles (SAV) are expected to be on our roads in the near future, there is a strong need to be able to predict inter-regional traffic demand so as to dispatch the proper number of SAV to areas of high demand. Location prediction model can be used to identify the demand hotspots by learning the mobility pattern of the users.

In our previous work \cite{choi2018network}, we proposed a Recurrent Neural Network (RNN) model to predict next locations in vehicle trajectories by adopting ideas from text generation model in natural language processing, where RNN has shown great success, and adapting them for use in our problem of location sequence prediction. The RNN model \cite{choi2018network} considered the previously visited locations as the only input to predict the next location. Despite its simple structure, the model produced promising results. For instance, for more than 50\% of all the tested trajectory samples, our RNN model showed a high prediction accuracy in that the probability of correctly predicting the next location was greater than 0.7, whereas the referenced non-RNN model (used for performance comparison) showed the similar accuracy level only for less than 5\% of the tested samples \cite{choi2018network}. To further improve the model performance, this study considers additional inputs that are likely to help predictions and proposes methodology that allows the incorporation of heterogeneous input soucres into the existing RNN framework. A specific input that we consider in this study is the surrounding traffic conditions of a vehicle at the time when it starts its journey. Nowadays, drivers can easily observe the current traffic state in the urban traffic networks and plan their journey (choose their routes) by using various traffic information and routing services \cite{adler2001investigating, cabannes2017impact}. As a result, the location sequences (chosen routes) of individual vehicles are expected to be influenced by the network traffic conditions at the beginning of their journeys. Inspired by this idea, this study proposes an Attention-based RNN model, which embeds an attention interface to enable the RNN model to consider the current traffic state as an additional input to location prediction. A detailed explanation is in the Methodology section.

\section{Methodology}
\subsection{Representing Urban Vehicle Trajectories as Cell Sequences}
Urban vehicle trajectory refers to a sequence of locations and times describing the path that a vehicle follows along its journey in urban traffic networks. Various sensors collect the location $(x,y)$ of vehicles and passage time($t$) to form the vehicle trajectory data. The data points in vehicle trajectories are continuous in space; that is, points are continuous-scaled coordinates of longitude and latitude. To learn movement patterns from a large amount of trajectory data, however it is necessary to define a finite set of representative locations that are common to all the trajectories that have the similar path. As such, the first step in building the trajectory prediction model is to discretize the vehicle trajectory data and convert each trajectory to a sequence of discretized locations. Based on the previous studies \cite{choi2018network, kim2016graph, kim2017trajectory}, we partition the urban traffic network into smaller regions, or cells, so that continuous-scaled raw vehicle trajectory data are represented as discretized cell sequence data.

Let $Tr=[(x_1,y_1 ),(x_2,y_2 ),\cdots,(x_l,y_l )]$ represent a raw vehicle trajectory consisting l number of data points, where data point $(x_i,y_i)$ represents the longitude and latitude coordinate of the vehicle’s $i^{th}$ position. Using a large number of vehicle trajectory data allows the data points in all trajectories to combine and cluster in space based on the desired radius, denoted by $R$. Accordingly, the distance between the centroid of the point cluster and its farthest member point is approximately R for each spatial cluster. The centroid is the mean location of the data points within the cluster, and by using Voronoi tessellation method, the cell boundaries of the clusters (Voronoi polygons) are determined. 

Given N cells in the network, a vehicle trajectory can be expressed as a cell sequence $[c_1,c_2,\cdots,c_m ]$, where $c_j$ is the index of the $j^th$ visited cell $(1 \leq c_j \leq N)$ within trajectory $Tr$. Since each of the visited cells covers multiple trajectory data points, the length of the cell sequence ($m$) is always less than or equal to the length of original vehicle trajectory ($l$) (i.e. $m \leq l$). In addition to the cell sequence covering the original trajectory, two virtual spatial tokens $\#start$ and $\#end$ are added to the front and the back of the cell sequence. These virtual tokens are treated as virtual cells that do not exist in the actual network but only indicate the start and the end of the trip.

The cell sequence is then separated into input vector $X$ and output label vector $Y$ for training, validating, and testing. Given the cell sequence containing $m+2$ cells including the start and the end tokens, input vector $X$ consists of first m+1 cells and output label vector $Y$ consists of $m+1$ elements starting from the second element ($c_1$). 

\begin{equation}
	\begin{array}{lcl}
		&X&=[X_0,X_1,X_2,...,X_m ]\equiv [\#start,c_1,c_2,...,c_m ] \\[6pt]
		&Y&=[Y_0,Y_1,Y_2,...,Y_m ]\equiv [c_1,c_2,...,c_m,\#end]
	\end{array}
\end{equation}

\begin{figure}[!ht]
%		\centering
	  \includegraphics[width=\textwidth]{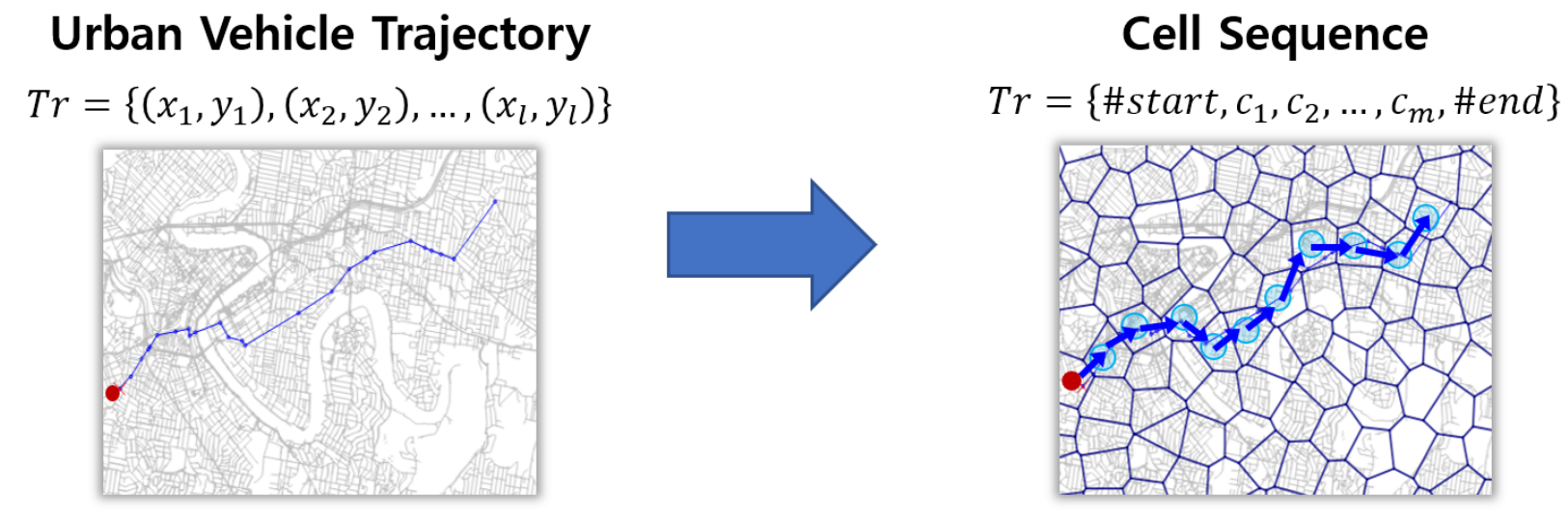}
	\caption{Representing urban vehicle trajctory as cell sequence}
	\label{figure1}
\end{figure}

\subsection{Cell Sequence Prediction using Recurrent Neural Network}
\begin{figure}[!ht]
	\centering
	  \includegraphics[width=0.75\textwidth]{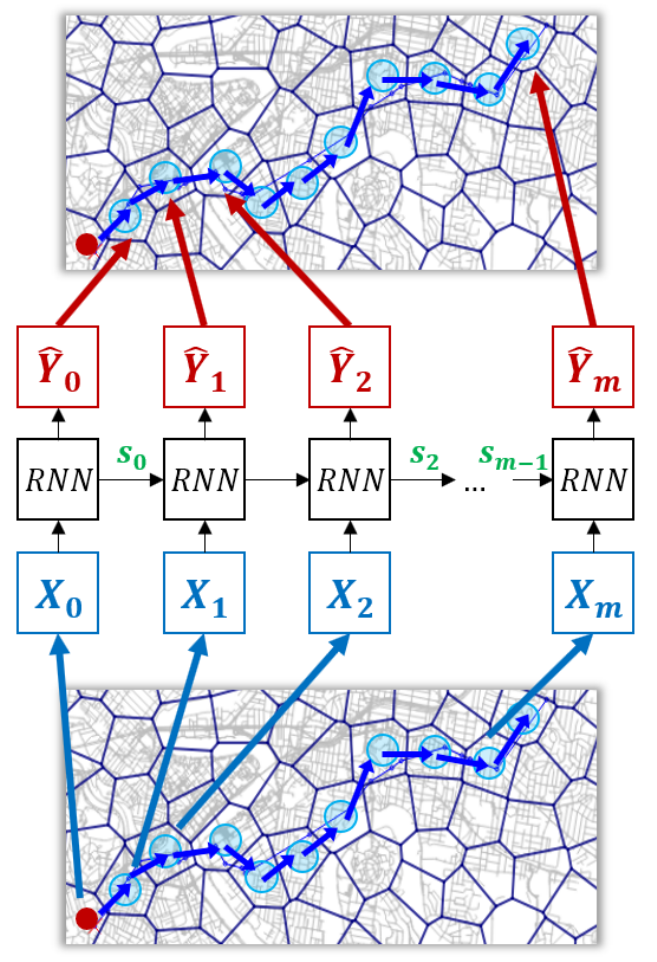}
	\caption{Structure of the basic Recurrent Neural Network model (RNN) for cell sequence prediction}
	\label{figure2}
\end{figure}

In our previous work \cite{choi2018network}, a Recurrent Neural Network (RNN) model for the cell sequence prediction was developed and evaluated. This previous RNN model was designed to predict the future cell sequences based purely on the previously visited cell sequence. In the training step, the model calculates the probability of each cell being visited in the next step ($\hat{Y}_i$) based on input vector $X$. The model structure is shown in the Fig. \ref{figure2}. Then, the model calculates the cross-entropy loss ($L(Y,\hat{Y} )$) between the correct label ($Y$) and the predicted label probability based on the current parameter ($\hat{Y}$). A basic Long Short Term Memory (LSTM) unit \cite{hochreiter1997long} is used as the hidden unit in the RNN model, i.e., RNN units in Fig. \ref{figure2}. For parameter estimation, the Adam optimizer was used to update the model parameters \cite{kingma2014adam}. A detailed explanation of the earlier model can be found in Chapter \ref{chapter:nlp_rnn}
% our previous paper \cite{choi2018network}. 

\subsection{Incorporating Network Traffic State Data into Cell Sequence Prediction}
The drivers can easily obtain the current traffic state in the urban traffic networks, and plan their journey by using various traffic information and routing services \cite{adler2001investigating, cabannes2017impact}. For example, between routes A and B, a driver is likely to choose route A when route B is congested and vice versa. As a result, the location sequences (chosen routes) of individual vehicles are expected to be influenced by the network traffic conditions at the beginning of their journeys. It is thus desriable to incorporate network-wide traffic state information and route choice behavior depending on the prevailing traffic state into the RNN-based cell sequence prediction model to increase the model’s prediction accuracy.

Adding additional information to RNN models, which is network traffic state in our case, is not a straightforward task. RNN models are specialied to process sequential data considering temporal dependency across time or sequence steps. When input data are all in the form of sequence, adding another sequence input can be done through a straightforward extention as RNN model can have multiple input layers and multiple hidden features to incorporate multiple sequence inputs and combine them to calculate the output. However, when the additonal input is non-sequential data, it cannot be directly represented as an input layer of the RNN model but rather should be processed outside the RNN model. The traffic state information we wish to add as an additional input to our RNN model is network-wide traffic density level at the beginning of the sequence, i.e., traffic information available at the origin ($X_0$) of a given trajectory), which is non-sequential data, making the problem more challenging. It may be possible to generate network traffic state data in a sequential form by feeding network traffic state at the time that a subject vehicle visits each cell in cell sequence. However, it requires a model that preidicts the location and visiting time simultaneously and that is beyond the scope of our current study, which focuses on location prediction only.

\begin{figure}[!ht]
%		\centering
	  \includegraphics[width=\textwidth]{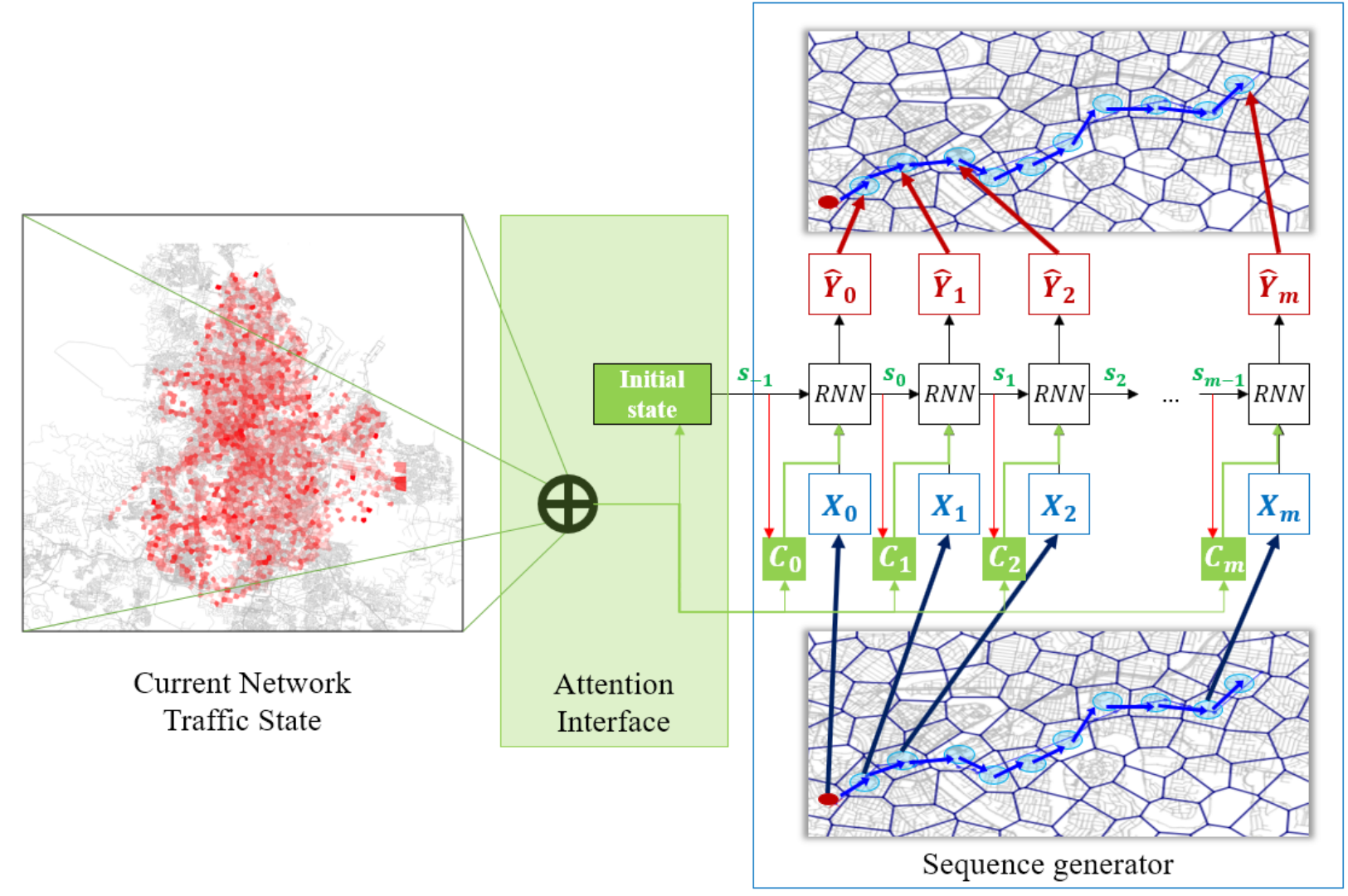}
	\caption{Structure of the proposed Attention-based Recurrent Neural Network model (ARNN) for cell sequence prediction}
	\label{figure3}
\end{figure}

One way to address this challenge is to introduce attention mechanism. The attention mechanism can be understood as an interface between external information processed outside the RNN model and sequential inputs processed inside the RNN model, as illustrated in Fig. \ref{figure3}. The attention mechanism in neural networks was first introduced to imitate the “attention mechanism” in human brain. When humans are asked to translate a sentence from one language to other language, humans try to think of words that matche the alignment and meaning of word while also considering the global context of the sentence. Similarly, when humans are asked to write a sentence based on an image, humans not only concentrate on the important part of the image but also think of the global context of the image to write a sentence. Using the attention mechanism in neural networks has shown significant improvements in model performance in applications such as machine language translation \cite{vaswani2017attention} and video captioning \cite{xu2015show}.

The attention mechanism allows the cell sequence prediction model or cell sequence generator to concentrate on certain part of the network traffic state input and use the information for cell sequence generation. There are mainly two tasks given to the attention mechanism: first is to set initial state for the RNN and second is to provide the network-wied traffic state information at each cell generation step. Usually, the initial state vector of RNN cell is set as zero vector since the simplest form of RNN does not consider additional information from other models or inputs. However, in the case of ARNN, there is an additional information of network traffic state. To consider this input in cell sequence generation, this information should be embedded into the model. Also, the attention mechanism allows the RNN to consider the traffic state in predicting the next location, or cell, at each step. The model is trained to calculate which information, or which region, to consider among the network traffic state data by calculating the context vector and attention weights.

Fig. \ref{figure3} shows the structure of the Attention-based Recurrent Neural Network (ARNN) model for the cell sequence prediction. There are two types of input data in this model: the first is the network traffic state data and the second is the cell sequence representation of vehicle trajectory data. The model first processes the current network traffic state and calculates the initial state ($s_{-1}$) for the RNN unit. Then, the attention interface calculates the context vector ($C_i$) based on the previous state vector. The context vector ($C_i$) is used as input to the $i^{th}$ RNN unit as well as the corresponding input vector element ($X_i$) to update current state vector ($s_i$). The attention weight $\alpha_(i,j)$ is calculated based on the context vector and previous state vector ($ \alpha_(i,j)=f(C_i,s_(i-1)  )  $ ). The attention weight represents the probability to attend to $j^{th}$ cell at $i^{th}$ sequence. Therefore, the sum of $\alpha_(i,j)$ at each sequence is 1 ( $ \sum_{\forall j} \alpha_{(i,j)} = 1$ ).

The input cell sequence ($X$) is processed based on the word-embedding method to represent the hidden features of the cells. In the training step, input vector X is directly used as an input of each RNN unit in order to calculate the output vector ($\hat{Y}_i$). However, in the testing step, only the front n cell sequence elements are directly used. Afterward, since the output vector represents the probability of each cell being visited, we use a random sampling based on the multinomial distribution with probability $\hat{Y}_i$ to extract the next cell, also it is used as the next input vector element.

A basic Long Short Term Memory (LSTM) cell \cite{hochreiter1997long} is used as RNN cell. And the model also uses the Adam optimizer to update the model parameters \cite{kingma2014adam}.

\section{Model Performance Evaluation}
\subsection{Data}
\subsubsection{Urban Vehicle Trajectory Data}
The vehicle trajectory data used in this research are collected from the Bluetooth sensors in Brisbane, Australia, provided by Queensland Department of Transport and Main Roads (TMR) and Brisbane City Council (BCC). The Bluetooth sensors are installed in state-controlled roads and intersections located inside the Brisbane City, and they detect Bluetooth devices (e.g., in-vehicle navigation systems and mobile devices) passing the sensors and record their passage time. By connecting the data points containing the same identifier of the Bluetooth device (MAC ID), the vehicle trajectories of individual vehicles can be constructed. Each vehicle trajectory represents a time-ordered sequence of Bluetooth sensor locations that a subject vehicle passes. If the corresponding vehicle does not move for more than an hour, it is considered that the vehicle trip has terminated. For this case study, we used the vehicle trajectory data collected in March 2016. There are approximately 276,000 trajectories in one day, and a total of 8,556,767 vehicle trajectories are collected in March 2016. We randomly sampled 200,000 vehicle trajectories for the training dataset, 10,000 vehicle trajectories for the validation dataset (used in hyper-parameter searching), and 200,000 vehicle trajectories for the testing dataset.

Brisbane urban traffic network is divided into “cells” to use the vehicle trajectory clustering and cell partitioning method proposed in the previous research \cite{kim2016graph, kim2017trajectory}. The desired radius of the cells is set to be 300m. Accordingly, a total of 5,712 cells are generated. Among them, 2,746 cells are considered to be active since the rest of the cells are not visited by any vehicles in the historical data of vehicle trajectories. The vehicle trajectory data are processed and transformed into cell sequence data.

\subsubsection{Network Traffic State Data}
There are several ways to represent the network traffic state such as density and average speed. In this study, vehicle accumulation, which is understood as the density of cells, is used to represent the network traffic state. The vehicle accumulation for a given cell is estimated by counting the number of vehicles that are present within the cell at a given instant point in time. We processed the vehicle trajectory data and calculated the vehicle accumulation of each cell at each minute. The vehicle accumulation data are normalized by dividing the vehicle accumulation by the historical maximum number of vehicle accumulation in each cell. 

The vehicle accumulation data are used as the network traffic state input to the ARNN model. When the ARNN model is trainend through each cell sequence, the model receives the vehicle accumulation data on a whole network from 10 minutes before the start time of the sequence. As a result, the shape of the input vehicle accumulation data is [N,10], where N is the number of cells in the study network.

\begin{figure}[!ht]
%		\centering
	  \includegraphics[width=\textwidth]{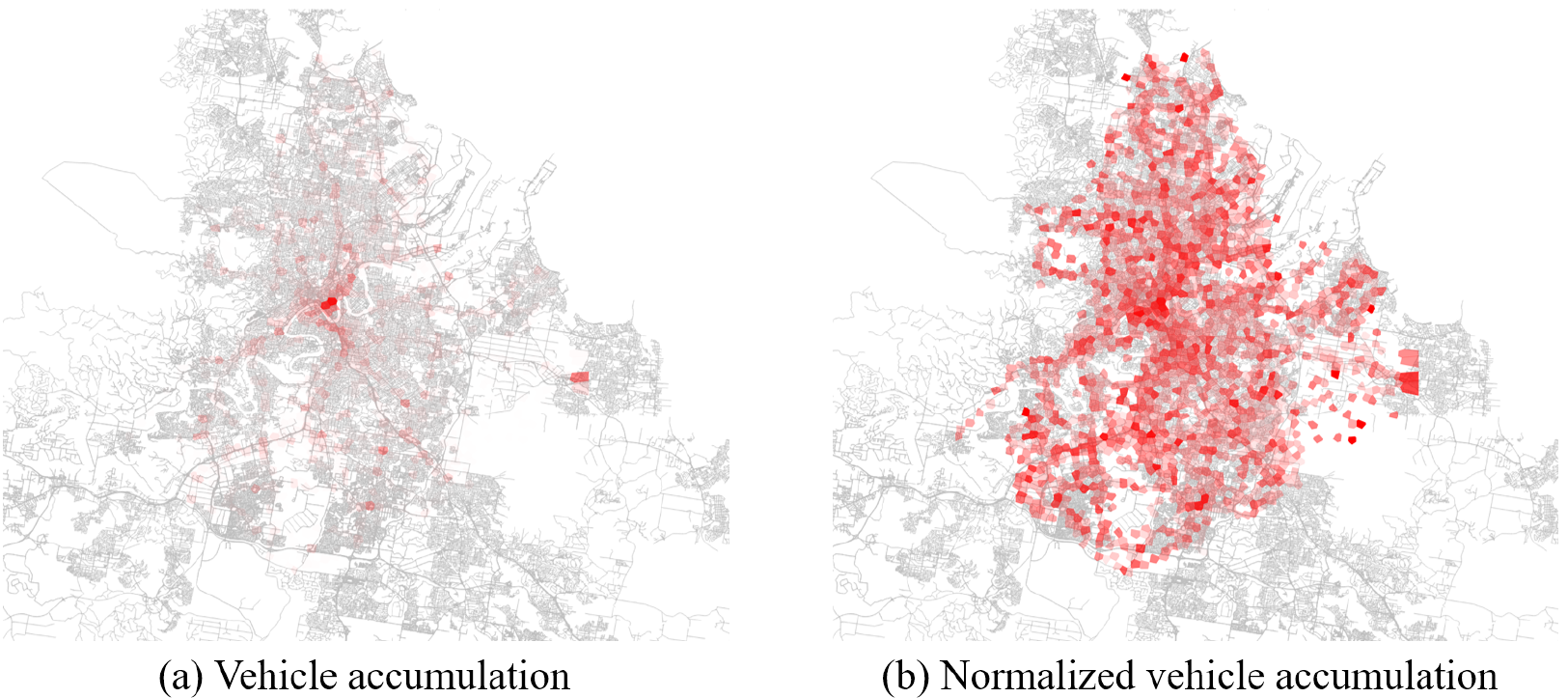}
	\caption{Spatial distribution of (a) vehicle accumulation, and (b) normalized vehicle accumulation at 12pm on March 1, 2016.}
	\label{figure4}
\end{figure}

\subsection{Hyperparameter and Model Training}
For each model, we applied hyperparameter searching algorithm to ensure that each model is trained to acheive its maximum performance. The hyperparameter searching algorithm used in this study is from a Python package called “Scikit-optimize” The hyperparameter searching algorithm is based on Bayesian optimization using Gaussian Process (GP). This algorithm approximates the function by assuming that the function values follow a multivariate Gaussian. The covariance of the function values is given by a GP kernel between the parameters. Then a smart choice to choose the next parameter to evaluate can be made by the acquisition function over the Gaussian prior which is much quicker to evaluate. 

There are three hyperparameters to search: learning rate, embedding layer dimension, and hidden layer dimension. The learning rate determines the updating step-size at each training step. If it is too large, the model is unlikely to converge. On the other hand, if it is too small, the speed of convergence is too slow, and the model is likely to fall into local minima. Therefore, finidng an appropriate learning rate is crucial in learning neural networks. The embedding layer dimension is used to convert the cell sequence input which is treated as one-hot vector to a vector in the latent space. In other words, the embedding layer extracts feature of each cell input and represents it as a numeric vector. The hidden layer dimension determines the dimension of LSTM cells and cell decoding layer. LSTM cell is used to calculate the state vector ($s_i$) and cell decoding layer is used to calculate the cell-visiting probability ($\hat{Y}_i$) from the state vector ($s_i$). 

The models are trained for 10 epochs for each hyperparameter set and the prediction accuracy is measured by applying the trained models to the validation dataset. The result of hyperparameters are shown in Table \ref{table1}.

\begin{table}[h]
	\caption{Hyperparameter result}
	\begin{tabular*}{\hsize}{@{\extracolsep{\fill}}llll@{}}
		\toprule
		Model &  Learning rate &  Dimension of Embedding Layer & Dimension of Hidden Layer\\
		\midrule
		RNN   &  6.216234e-05 &  413 &854\\
		ARNN &  5.842804e-04 &  659 &574\\
		\bottomrule
	\end{tabular*}
	\label{table1}
\end{table}

\section{Result}
\subsection{Score based Evaluation of Generated Cell Sequences}
In this study, we use two widely used evaluation metrics in sequence modeling to evaluate the accuracies of generated cell sequences: BLEU score \cite{papineni2002bleu}  and METEOR score \cite{banerjee2005meteor}. 
%	In this study, two widely used evaluation metrics in sequence modeling, BLEU score and METEOR score, are used to evaluate the performance of the models (RNN and ARNN). 
In the previous study \cite{choi2018network}, we used the complementary cumulative distribution function of the probability to measure how accurately the model predicts the next 1, 2, or 3 consecutive cells. While this measure is intuitive and easy to interpret, there is a drawback in this method in that it considers element-wide prediction accuracy and does not take into account the whole sequence. The element-wide performance measure can be sensitive to small local mis-predictions and tend to underestimate the performance of the model. For example, when the original cell sequence is $[\#start,c_1,c_2,c_3,\#end]$ and a model is asked to predict the next cells based on the given cell sequence $[\#start,c_1 ]$, the prediction of $[c_2,c_4,c_3,\#end]$ will be considered as incorrect and performing poorly by our previous method because the model miss-predicted one cell $c_4$, even though the overall sequence is very similar to the original sequence. 
As such, this study employs  $BLEU$ score and $METEOR$ score that consider the whole sequence and thus more robust and accurate as performance measure for sequence modeling. 

%	When reference sequence is given, BLEU uses three methods to evaluate the similarity between the reference sequence and the generated sequence. BLEU scans through the sequence and check if the generated sequence contains identical chunks which are found in the reference sequence. 

\subsubsection{BLEU score}
When reference sequence is given, $BLEU$ uses three methods to evaluate the similarity between the reference sequence and the generated sequence. This metric is one of the most widely used metrics in natural language processing and sequence-to-sequence modeling. $BLEU$ scans through the sequence and check if the generated sequence contains identical chunks which are found in the reference sequence. Here, $BLEU$ uses a modified form of precision to compare a reference sequence and a candidate sequence by clipping. For the generated sequence, the number of each chunk is clipped to a maximum count ($m_{max}$) to avoid generating same chunks to get higher score.

\begin{equation}
	\begin{array}{lcl}
		P_n=\frac{ \sum_{i \in C}   \min{(m_i,m_{i,max})}   }{w_t}
	\end{array}
\end{equation}

where $C$ is a set of cells (or chunks) in the generated sequence, $m_i$ is the number of the cell (or chunk) $i$ in the generated sequence, $m_{i, max}$ is the number of the cell (or chunk) $i$ in reference sequence, and $w_t$ is the total number of cells in candidate sequence. When $n$ is 1, the chunks represent the cells in the sequences. Otherwise, we consider $n$ consecutive cells as chunk and calculate the precision for each $n$-cell-unit.

The $BLEU-n$ score represents the geometric mean of $P_i$'s with different $i$'s multiplied by a brevity penalty to prevent very short candidates from receiving too high score. 

%	\text
\begin{equation}
	\begin{array}{lcl}
		BLEU_n = min(1, \frac{L_{gen}}{L_{ref}}  ) \cdot (\prod_{i=1}^{n} P_i)^{\frac{1}{n}}
%			P_n=\frac{ \sum_{i \in C}   \min{(m_i,m_{i,max})}   }{w_t}
	\end{array}
\end{equation}

where $L_{gen}$ represents the length of generated sequence, $L_{ref}$ represents the length of reference sequence.

%	where $m$ refers to the number of a single cell in candidate cell sequence which are also found in the reference cell sequence, $m_max$ is maximum number of each cell in reference sequence which are in candidate sequence, and $w_t$ refers to the number of cells in candidate sequence.

%	Basically, the $BLEU$ score uses a single cell as a unit for score calculation. If the sequences are divided by multiple cells ($n$ cells), we calculate the $BLEU$ score for each $n$-cell-unit. This measures whether the model can predict the next cells consecutively. The $BLEU-n$ score is the average of $BLEU$ scores, in which the sequence is divided at all orders from $1$ to $n$.

\subsubsection{METEOR score}
$METEOR$ \cite{banerjee2005meteor} first creates an alignment between candidate cell sequence and reference cell sequence. The alignment is a set of mappings between the most similar cells. Every cell in the candidate sequence should be mapped to zero or one cell in the reference sequence. $METEOR$ chooses an alignment with the most mappings and the fewest crosses (fewer intersection between mappings). 

To calculate $METEOR$ score, we first define precision $P$ and recall $R$.

\begin{equation}
	\begin{array}{lcl}
		P=\frac{\min{(m,m_{max})}}{w_t}
	\end{array}
\end{equation}

\begin{equation}
	\begin{array}{lcl}
		R=\frac{\min{(m,m_{max})}}{w_r}
	\end{array}
\end{equation}

where m refers to the number of single cells in candidate cell sequence which are also found in the reference cell sequence, $m_{max}$ refers to the sum of maximum number of each cell in reference cell sequence which are in candidate cell sequence, $w_t$ refers to the number of cells in candidate cell sequence, and $w_r$ refers to the number of cells in reference cell sequence. 

Then, we calculate the weighted harmonic mean between precision and recall, where the ratio of the weights is 1:9.

\begin{equation}
	\begin{array}{lcl}
		F_{mean}  = \frac{10}{\frac{1}{P} + \frac{9}{R}} = \frac{10PR}{R+9P}
		%F_mean=10/(1/P+9/R)=10PR/(R+9P)	
	\end{array}
\end{equation}

To account for congruity with respect to a longer cell segment that appears both in reference and candidate cell sequences, we generate mappings based on the longer cell segment and use it to compute the penalty p. The more mappings there are, which are not adjacent in the reference and the candidate cell sequence, the higher the penalty will be. The penalty is calculated as follows:

\begin{equation}
	\begin{array}{lcl}
		p=0.5 (\frac{c}{u_m})^3
		%
		%p=0.5(c/u_m )^3	
	\end{array}
\end{equation}

where c is a set of single cells that are not adjacent in the candidate and refernce sequence, and $u_m$ is the number of single cells that have been mapped. This penalty reduces $F_{mean}$ up to 50\% and calculate the $METEOR$ score (M).

\begin{equation}
	\begin{array}{lcl}
		M=F_{mean}(1-p) 
	\end{array}
\end{equation}

\subsection{Score Result}\label{arnn:scoreresult}

For each sequence in the test dataset, the scores are calculated by the following procedure.

Let $Tr$ be the subject cell sequence with length m, which is expressed as:

\begin{equation}
	\begin{array}{lcl}
		Tr=[\#start,c_1,c_2,...,c_m,\#end]  
	\end{array}
\end{equation}
The subject cell sequence is divided into 2 parts: The sequence given ($(Tr)_{1:g}$) and the sequence to be predicted ($(Tr)_(g+1:m)$), where g is the number of cells given to the models (ARNN and RNN).
\begin{equation}
	\begin{array}{lcl}
		&Tr_{1:g}=[\#start,c_1,c_2,...,c_g ]  
		\\
		&Tr_{g+1:m}=[c_{(g+1)},...,c_m,\#end]	
	\end{array}
\end{equation}
Each model predicts 100 candidate cell sequences based on $Tr_{1:g}$ producing a set of 100 $(Tr)_{g+1:m}$ sequences for each $(Tr)_{1:g}$. The generated candidate cell sequences are the cell sequences that have \#end token at the end, representing that the trip has terminated. These candidate cell sequences may not have the same length with the original cell sequence. The length can be longer or shorter depending on when the model predicts \#end token. The $Tr_{g+1:m}$ is used as reference cell sequence to calculate the score presented above. For each score ($BLEU_1$, $BLEU_2$, $BLEU_3$, $BLEU_4$, and $METEOR$), 100 score values are calculated based on the generated 100 candidate cell sequences. The average value of each score is used to represent the model performance of the corresponding cell sequence ($Tr$). 

\begin{algorithm}
	\caption{Pseudo-code for score evaluation}
\hspace*{\algorithmicindent} \textbf{Input:} $Tr_{1:g}$ (given sequence), \\
\hspace*{\algorithmicindent} \hspace{0.45in} $Tr_{g+1:m}$ (target sequence)\\
\hspace*{\algorithmicindent} \hspace{0.45in} $F$ (prediction model)\\
\hspace*{\algorithmicindent} \hspace{0.45in} $f_{score}$ (score calculation function)\\
\hspace*{\algorithmicindent} \hspace{0.45in} $N$ (number of predictions)\\
\hspace*{\algorithmicindent} \textbf{Output:} $Score$ \\
	\begin{algorithmic}[1]
	\State $scoreList \gets List()$
	\For{$i = 1,2,\cdots,N$}
	    \State $\hat{Tr}_{g+1:m} = F(Tr_{1:g})$ \Comment{Predict future location sequence}
	    \State $score_i = f_{score} (Tr_{g+1:m} , \hat{Tr}_{g+1:m})$ \Comment{Calculate score}
	    \State $scoreList.append(score_i)$ \Comment{Append score to $scoreList$}
	\EndFor
	\State $Score \gets Average(scoreList)$ \Comment{Take average of $scoreList$ as score}
	\State\Return $Score$
	\end{algorithmic} 
	\label{algorithm:arnn}
\end{algorithm}

\begin{figure}[!ht]
%		\centering
	  \includegraphics[width=\textwidth]{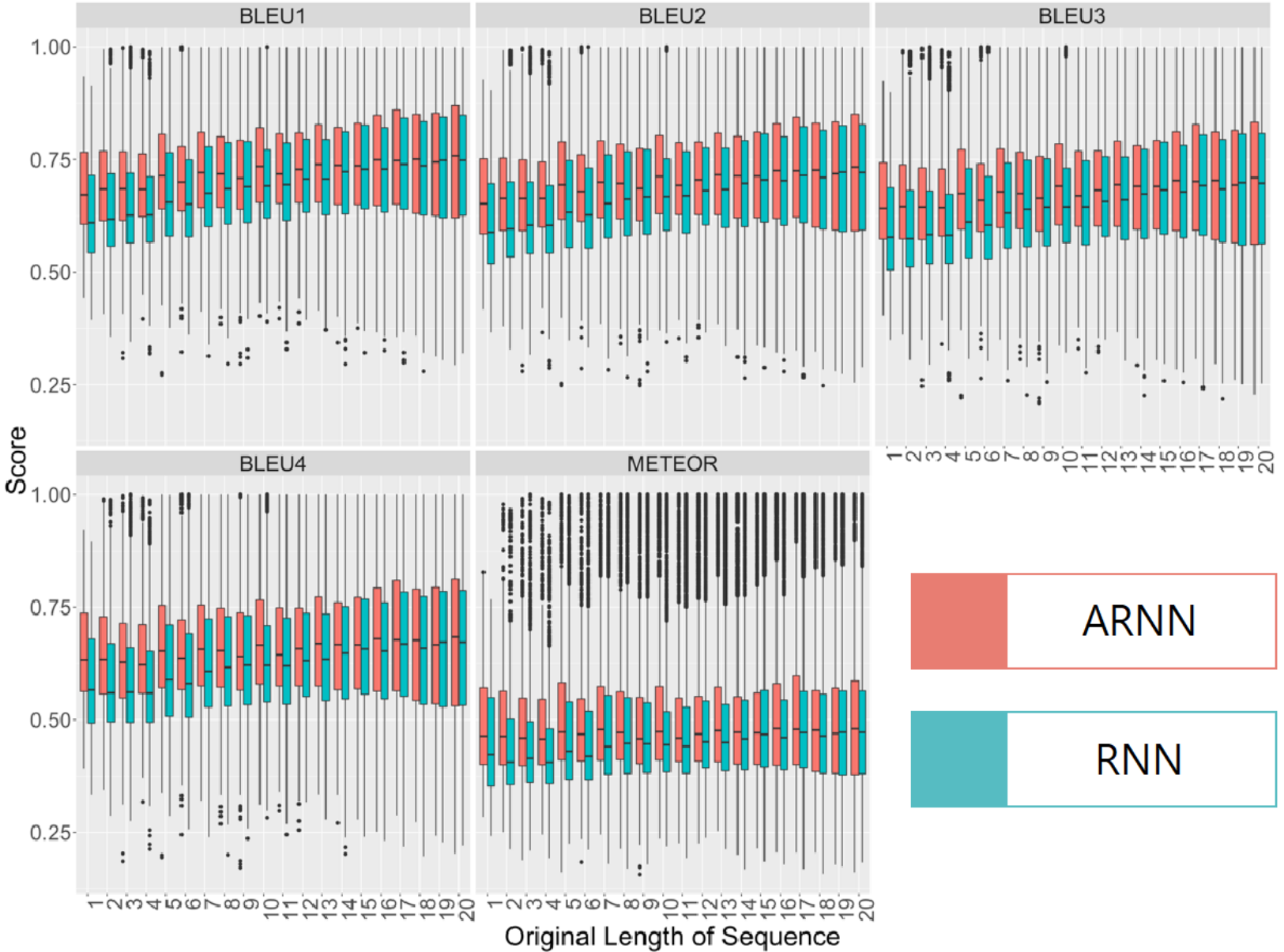}
	\caption{Boxplot of models (ARNN, RNN) for each original length of sequence (m)}
	\label{figure5}
\end{figure}

10,000 cell sequences in test dataset is used to calculate the five scores.	Fig. \ref{figure5} shows the score result of each model. The x-axis represents the original length of the sequence, and the y-axis represents the value for each score metrics. The result of ARNN model (red color) shows better performance compared to the result of RNN model (blue color). 

The result shows that ARNN model can predict short cell sequences more accurately up to 12\% and long cell sequences more accurately up to 5\%. ARNN model outperforms the RNN model in terms of both $BLEU$ score and $METEOR$ score. It is worth noting that the ARNN model had performance improvement in terms of $METEOR$ score as a high $METEOR$ score not only requires good prediction of visited cells but also accurate description of cell alignment (the visiting order of cells). The result thus confirm that the ARNN model using the attention mechanism achieves improvements in predicting the composition of cells in the sequence accurately as well as the alignment of the cells in the sequence.

\begin{figure}[!ht]
%		\centering
	  \includegraphics[width=\textwidth]{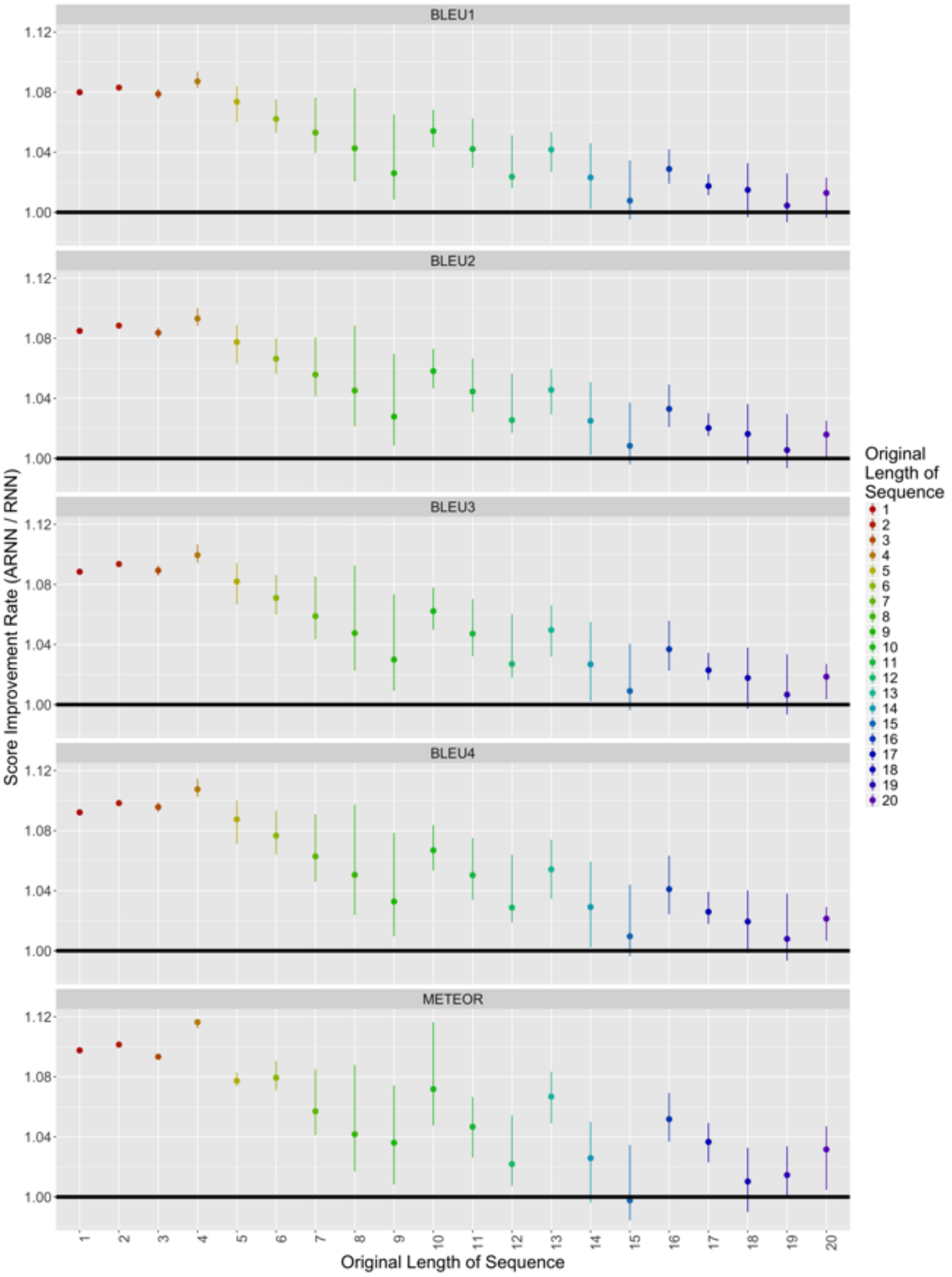}
	\caption{Score improvement rate for each original length of sequence (m). The points represent the average value, and the lines represent the range of the score improvement rate (from the minimum value to the maximum value)}
	\label{figure6}
\end{figure}

The performance gap between two models tends to decrease as the original length of the sequence increases. Fig. \ref{figure6} shows the result in terms of score improvement rate. The score improvement rate is defined as ratio of the performance score of ARNN model to the performance score of RNN model ($=(score_{ARNN})/(score_{RNN} )$). For each number of given cell sequence (g) and original length of cell sequence (m), this performance improvement rate is measured. And the Fig. \ref{figure6} shows the summarized result. Points in Fig. \ref{figure6} represents the average performance improvement rate for each original length of cell sequence (m), and the line represents the range of this value (from the minimum value to the maximum value).

In Fig. \ref{figure6}, one interesting observation is that the performance improvement rate decreases and converges to 1 (the black lines) as the original length of the cell sequence increases. This can be because the input feature given to the ARNN model is the network traffic state at the beginning of each trajectory journey. This observation has an important implication for the influence of pre-trip information in route choice behaviors. The fact that the ARNN improves the prediction at the early stages of a journey implies that the pre-trip information indeed influences travellers’ route choice decisions and differentiates route choice patterns between different pre-trip traffic conditions. The fact that the effect of pre-trip information fades away at the later stages of the journey may be the indication of drivers’ reliance on en-route trip information instead of pre-trip information and thus indicates a need for incorporating such en-route information into the model to further improve the model performance.

%	\clearpage
%	\clearpage
\clearpage

\section{Conclusion and Future Studies}

This research studies urban vehicle trajectory prediction, one of the applications of trajectory data mining. Based on the previous work \cite{choi2018network}, in this study, we proposed a novel approach to incorporate network traffic state data into urban vehicle trajectory prediction model. Attention mechanism is used as an interface to connect the network traffic state input data to the vehicle trajectory predictor proposed in the previous work. ARNN model, which is Attention-based RNN model for cell sequence prediction, is compared with RNN model, which is RNN model for cell sequence prediction, in terms of conventional scoring methods in sequence prediction. The result shows that ARNN model outperformed RNN model. The result shows that it is effective to use attention mechanism to structurally connect the network traffic state input to RNN model to predict the vehicle’s future locations. Especially, it is promising that the ARNN model showed significant performance improvement in terms of METEOR which considers not only the cells to be visited but also the alignment of the cells in sequence. The performance improvement rates tend to decrease and converge to 1 as the original number of cell sequence increase. For the further improvement of the ARNN model, this problem should be studied to maintain the performance improvement rate at steady level.

There are some limitations in this study, so further works should cover such topics. First of all, in this study, the network traffic state data were normalized by using the historical maximum value of each cell. This makes easy to represent the network traffic state, but this may lead to some problems that normalized data of the cells with very low traffic may be too sensitive to small number of vehicles and count it as heavy congestion. This makes the model overreact to these cells and makes the cell sequence prediction confused. For the further improvement of this study, different types of normalization methods should be tested. Second, although the performance measures ($BLEU$ and $METEOR$) are widely used in the fields studying sequence prediction such as natural language processing, the application of these metrics is new in the transportation domain. The interpretation and implication of these metrics in the context of traffic modeling should be further investigated and proposed.

\chapter{Synthetic Trajectory Generation: TrajGAIL - Generating Urban Vehicle Trajectories using Generative Adversarial Imitation Learning  }\label{chapter:trajgen}\footnote{This chapter is based on following article: Choi, Seongjin, Jiwon Kim, and Hwasoo Yeo. "TrajGAIL: Generating urban vehicle trajectories using generative adversarial imitation learning." Transportation Research Part C: Emerging Technologies 128 (2021): 103091. \cite{choi2021trajgail}}
\section{Introduction}
% Trajectory Data
% \colorbox{yellow}{this}
Rapid advancements in location sensing and wireless communication technology enabled us to collect and store a massive amount of spatial trajectory data, which contains geographical locations of moving objects with their corresponding passage times \cite{lee2011trajectory}. Over the last decade, considerable progress is made in collecting, pre-processing, and analyzing trajectory data. Also, the trajectory data analysis is applied in various research areas, including behavioral ecology \cite{de2019trajectory}, transportation engineering \cite{wu2018location}, and urban planning \cite{laube2014computational}.
    
% Vehicle Trajectory Data
In transportation engineering, urban vehicle trajectory data are collected based on the location sensors installed inside vehicles or at the roadside and analyzed with various methods. The high-resolution mobility data of individual users in urban road networks offer unprecedented opportunities to understand vehicle movement patterns in urban traffic networks. It provides rich information on both aggregated flows and disaggregated travel behaviors. The aggregated flows include the origin-destination (OD) matrix and cross-sectional link traffic volumes. The disaggregated travel behaviors include user-centric travel experiences, namely, speed profile, link-to-link route choice behavior and travel time experienced by individual vehicles, as well as system-wide spatio-temporal mobility patterns, such as origin-destination pairs, routing pattern distributions, and network traffic states \cite{kim2015spatial}.
    
% examples of vehicle trajectory data analysis in transportation 
% Among many research problems in urban vehicle trajectory data analysis, the \textit{location prediction problem} draws the attention of many researchers because of its applicability to Location-Based Services (LBSs) \cite{monreale2009wherenext, martinez_agent-based_2015}. In this problem, researchers analyze a large number of trajectories of vehicles moving in an urban traffic network and make predictions on the future locations of the subject vehicle. This problem can be classified into two categories based on the target prediction variables. The first category is the next location prediction problem, where locations in trajectories are defined in various ways. In \cite{gambs2010show} and \cite{choi2019real}, for example, the authors proposed algorithms to predict the next location represented as Point-of-Interest (POI). In \cite{choi2018network}, an urban road network is partitioned into zones based on the clustering of trajectory data points, and the Recurrent Neural Network-based prediction model is proposed to predict the zone that the subject vehicle would visit. The second category is the destination prediction problem. \cite{ebel2020destination} proposed a multi-step approach to predicting the most probable routes to the destination. 
% Also, in \cite{xue2015solving}, an algorithm to predict the destination with sparse data is proposed. 

% Vehicle Trajectory analysis problem 
%--> new problem (trajectory data generation) and why it is important
% discriminative vs generative
Most of the studies in the vehicle trajectory data analysis use machine learning methods. The recurrent neural network, for example, is used by many previous researchers due to its ability to learn sequential information in trajectory data. In machine learning, there are mainly two approaches to modeling: the \textit{discriminative} and the \textit{generative} modeling. A discriminative model learns a direct map from input $X$ to output (label) $Y$ or posterior probability $P(Y|X)$, which is the conditional probability of each label $Y$ given the input variable $X$. It only learns the decision boundaries between labels and does not care about the underlying distribution of data. In contrast, a generative model captures the underlying probability distribution, i.e., joint probability $P(X, Y)$, from which $P(Y|X)$ can be computed. One advantage of a generative model is that we can generate new (synthetic) data similar to existing data by sampling from $P(X, Y)$. 

Synthetic data generation based on generative models has gained increasing importance as the data generation process plays a significant role in various research fields in an era of data-driven world \cite{popic2019data}. It is mainly used to serve two purposes. The first purpose is to deal with the lack of real data. In many research fields, data collection is costly, and, therefore, it is often difficult to collect enough data to properly train and validate models. In this case, it is useful to generate synthetic data that are similar to the real observations to increase training and test samples. The second purpose is to address the issue with the privacy and confidentiality of real data. Many types of data contain personal information, such as gender, name, and credit card usage. Synthetic data can be combined with or replace such privacy-sensitive data with a reasonable level of similarity, thereby protecting privacy while serving the intended analysis.

Urban vehicle trajectory analysis has both challenges: data sparsity and data privacy issues. Although the sources and availability of urban trajectory data are increasing, most of the currently available trajectory datasets cover only a portion of all vehicles in the network. From network management and operations perspectives, there is a desire to infer vehicle trajectories that represent the whole population to have a more complete view of traffic dynamics and network performance. Moreover, urban vehicle trajectory data may contain personal information of individual drivers, which poses serious privacy concerns in relation to the disclosure of private information to the public or a third party \cite{chow2011privacy}. The ability to generate synthetic trajectory data that can realistically reproduce the population mobility patterns is, therefore, highly desirable and expected to be increasingly beneficial to various applications in urban mobility.

% generative model in next location prediction    
% It is possible to use the models developed in the \textit{next location prediction problem} for data generation. 

While synthetic trajectory data generation is a relatively new topic in transportation research communities, there are several existing research areas that have addressed similar problems. 
One example is \textit{trajectory reconstruction}. When two points in a road network are given as an initial point (treated as sub-origin) and a target point (treated as sub-destination), the models reconstruct the most plausible route between the two points. The trajectory reconstruction can be considered as generating trajectories between sub-origins and sub-destinations. Previous studies such as \cite{chen2011discovering} and \cite{hu2018graph} investigated on discovering the most popular routes between two locations. \cite{chen2011discovering} first constructs a directed graph to simplify the distribution of trajectory points and used the Markov chain to calculate the transfer probability to each node in the directed graph. The transfer probability is used as an indicator to reflect how popular the node is as a destination. The route popularity is calculated from the transfer probability of each node. \cite{hu2018graph} also used a graph-based approach to constructing popular routes. The check-in records which contain the route's attributes are analyzed to divide the whole space into zones. Then, the historical probability is used to find the most plausible zone sequences. 
Also, \cite{feng2015vehicle} and \cite{rao2018origin} estimated origin-destination patterns by using trajectory reconstruction. Both studies used particle filtering to reconstruct the vehicle trajectory between two points in automatic vehicle identification data. The reconstructed vehicle trajectory is then used to estimate the real OD matrix of the road network. 
Another problem that is relevant to trajectory generation is the \textit{next location prediction} problem, where the "next location" of a subject vehicle is predicted based on the previously visited locations of the subject vehicle. 
% Many studies adopt a concept of predicting the predefined location to simplify the prediction problem. 
\cite{monreale2009wherenext}, for example, presented a decision tree to predict the next location based on the previously visited locations. Decision-tree based models, however, occasionally overfit the training dataset and lack the generalization ability to produce diverse trajectory patterns. \cite{gambs_next_2012} used Mobility Markov chain (MMC) to predict the next location among the clustered points or Point-of-Interests (POIs). 
The POIs considered in \cite{gambs_next_2012} are home, work, and other activity locations to model human activity trajectories throughout the day, rather than vehicle movement trajectories reflecting link-to-link vehicle driving behavior considered in this study. 
\cite{choi2019real} used a feed-forward neural network to predict the next intersection in a grid-structured road network. A set of intersections in Brisbane, Australia are treated as POI's to capture the link-to-link route choice behavior. \cite{jin2019augmented} used an augmented-intention recurrent neural network model to predict locations of vehicle trajectories of individual users. \cite{jin2019augmented} incorporated additional information on individual users' historical records of frequently visited locations into a next location prediction model. The past visited locations in historical records are represented as edge-weighted graph, and graph convolution network is used to incorporate this information into trajectory prediction. In \cite{choi2018network}, an urban road network is partitioned into zones based on the clustering of trajectory data points, and the prediction model based on recurrent neural network (RNN) is proposed to predict the zone that the subject vehicle would visit. \cite{choi2019attention} extended the idea of predicting the next zone and used network traffic state information to improve the RNN model's performance.

% \newpage

% trajectory reconstruction (imputation)

% trajectory prediction

% While synthetic trajectory data generation is a relatively new topic in transportation research communities, there is some existing work that can be applied to address this topic. 

% The \textit{next location prediction problem} is one research problem in vehicle trajectory data analysis, in which the "next location" of a subject vehicle is predicted based on the previously visited locations of the subject vehicle \cite{monreale2009wherenext,gambs2010show,martinez_agent-based_2015,choi2018network,choi2019real}. 

In fact, the existing models developed for the next location prediction problem can be applied for synthetic trajectory data generation. By sequentially applying the next location predictions, a synthetic vehicle trajectory can be generated. However, most of the existing models for next location prediction adopt a discriminative modeling approach, where the next locations are treated as labels and the model is trained to predict one or two next locations. The discriminative models have limitations in generating full trajectories, especially when sample trajectory data are sparse. it is only the decision boundaries between the labels that the models are trained to predict, not the underlying distributions of data that allow proper generalization for sampling realistic trajectories. As a result, it is necessary to develop a model based on the generative modeling approach to successfully perform synthetic trajectory data generation.

% \newpage

% 
% However, most of the models in the next location prediction problem are based on the discriminative modeling approach. Some studies use the generative modeling approach, but the training procedures of these studies are designed similarly to a discriminative modeling approach. 
% . In these studies, 
% The next locations are treated as labels, and the model is trained to predict one or two next locations. Consequently, the models with the discriminative modeling approach may capture some useful patterns in vehicle trajectories, but the accuracy declines when the models are used in a generative manner. \cite{ouyang2018non}. 
% 
% As a result, it is necessary to develop a model based on generative modeling approach with properly designed training procedure.

% trajectory generation problem
% In this context, it is necessary to develop a generative model for urban vehicle trajectories. Furthermore, the training procedure should be designed appropriately. A well-trained generative model for urban vehicle trajectories should be able to create realistic synthetic data. The individual generated trajectories should be similar to the real vehicle trajectories generated by human drivers. The dataset of generated trajectories should have statistical similarity with the given real dataset. Moreover, It is required to consider the next locations, as suggested by the previous researches on the  \textit{next location prediction problem}, as well as the overall similarity to learn the underlying distribution of the given dataset.

In this paper, we apply \textit{imitation learning} to develop a generative model for urban vehicle trajectory data. Imitation learning is a sub-domain of reinforcement learning for learning sequential decision-making behaviors or "policies". Unlike reinforcement learning that uses "rewards" as signals for positive and negative behavior, imitation learning directly learns from sample data, so-called "expert demonstrations," by imitating and generalizing the expert' decision-making strategy observed in the demonstrations. If we consider an urban vehicle trajectory as a sequence of decisions for choosing road links along a travel path, imitation learning can be applied to develop a generator that can reproduce synthetic data by imitating the decision-making process (i.e., driver' route choice behavior) demonstrated in the observed trajectory dataset. One approach to imitation learning is called \textit{Inverse Reinforcement Learning} (IRL), which aims to recover a reward function that explains the behavior of an expert from a set of demonstrations. Using the recovered expert reward function as feedback signals, the model can generate samples similar to the expert' decisions through reinforcement learning. \cite{ziebart2008maximum} and \cite{ziebart2008navigate} used maximum entropy IRL (MaxEnt) to generate synthetic trajectories similar to a given taxi dataset. One of the advantages of using IRL is that the model generates trajectories using both current states and expected returns of future states to determine an action---as opposed to considering only the knowledge up to the current state (e.g., previous visited locations)---, thereby enabling a better generalization of travel behavior along the whole trajectory.

% generative adversarial framework : GAN
Recently, there have been remarkable breakthroughs in generative models based on deep learning. In particular, \cite{goodfellow2014generative} introduced a new generative model called Generative Adversarial Networks (GAN), which addressed inherent difficulties of deep generative models associated with intractable probabilistic computations in training. GANs use an adversarial \textit{discriminator} to distinguish whether a sample is from real data or from synthetic data generated by the \textit{generator}. The competition between the generator and the discriminator is formulated as a minimax game. As a result, when the model is converged, the optimal generator would produce synthetic sample data similar to the original data. The generative adversarial learning framework is used in many research fields such as image generation \cite{radford2015unsupervised2}, audio generation \cite{oord2016wavenet}, and molecular graph generation \cite{de2018molgan}. 
% add GAN-based studies in transportation engineering

% \cite{zhang2019novel}
% \cite{xu2020ge}
% \cite{li2020coupled}
% \cite{kuefler2017imitating}
% % 
GANs have been also applied in transportation engineering. \cite{zhang2019novel} proposed trip travel time estimation framework called \textit{T-InfoGAN} based on generative adversarial networks. They used a dynamic clustering algorithm with Wasserstein distance to make clusters of link pairs with similar travel time distribution, and they applied Information Maximizing GAN (InfoGAN) to travel time estimation. \cite{xu2020ge} proposed Graph-Embedding GAN (GE-GAN) for road traffic state estimation. Graph embedding is applied to select the most relevant links for estimating a target link and GAN is used to generate the road traffic state data of the target link. In \cite{li2020coupled}, GAN is used as a synthetic data generator for GPS data and travel mode label data. To solve the sample size problem and the label imbalance problem of a real dataset, the authors used GAN to generate fake GPS data samples of each travel mode label to obtain a large balanced training dataset.
The generative adversarial learning framework is also used for synthetic trajectory generation. \cite{liu2018trajgans} proposed a framework called trajGANs. Although this paper does not include specific model implementations, it discusses the potential of generative adversarial learning in synthetic trajectory generation. Inspired by \cite{liu2018trajgans}, \cite{rao2020lstm} proposed LSTM-TrajGAN with specific model implementations. The generator of LSTM-TrajGAN is similar to RNN models adopted in the next location prediction studies. 
% It only uses the previously visited locations to determine the next location in generating process without considering the remaining trip.

% However, directly applying GAN for synthetic trajectory generation 

% % Research Objective
This study proposes \textit{TrajGAIL}, a generative adversarial imitation learning (GAIL) model for urban vehicle trajectory data. GAIL, proposed by \cite{ho2016generative}, uses a combination of IRL's idea that learns the experts' underlying reward function and the idea of the generative adversarial framework. GAIL effectively addresses a major drawback of IRL, which is high computational cost. However, the standard GAIL has limitations when applied to the vehicle trajectory generation problem because it is based on the IRL concept that only considers a vehicle's current position as states in modeling its next locations \cite{ziebart2008maximum,ziebart2008navigate,zhang2019unveiling}, which is not realistic as a vehicle's location choice depends on not only the current position but also the previous positions. To overcome these limitations, this study proposes a new approach that combines a partially-observable Markov decision process (POMDP) within the GAIL framework. POMDP can map the sequence of location observations into a latent state, thereby allowing more generalization of the state definition and incorporating the information of previously visited locations in modeling the vehicle's next locations. In summary, the generation procedure of urban vehicle trajectories in TrajGAIL is formulated as an imitation learning problem based on POMDP, which can effectively deal with sequential data, and this imitation learning problem is solved using GAIL, which enables trajectory generation that can scale to large road network environments.

% We formulate the urban vehicle trajectory generation as the imitation learning problem and use the framework of GAIL to train the trajectory data generator. TrajGAIL is capable of generating synthetic vehicle trajectories with distributional similarities with a given dataset.

% \cite{chen2020trajvae}
% \cite{rao2020lstm}

This paper is organized as follows. Section \ref{sec:method} describes the methodology of this paper. A detailed problem formulation is presented in Section \ref{ProbForm}, and the proposed framework of TrajGAIL is presented in Section \ref{sec:framework}. Section \ref{sec:eval} describes how the performance of the proposed model is evaluated. Section \ref{sec:data} introduces the data used in this study, and Section \ref{sec:baseline} introduces the baseline models for performance comparison. In Section \ref{sec:result}, the evaluation results are presented at both trajectory-level and dataset-level. Finally, Section \ref{sec:conclusion} presents the conclusions and possible future research.

\section{Methodology}\label{sec:method}

The objective of TrajGAIL is to generate location sequences in urban vehicle trajectories that are similar to real vehicle travel paths observed in a road traffic network. Here, the "similarity" between the real vehicle trajectories and the generated vehicle trajectories can be defined from two different perspectives.  First, the \textit{trajectory-level} similarity measures the similarity of an individual trajectory to a set of reference trajectories. For instance, the probability of accurately predicting the next locations---single or multiple consecutive locations as well as the alignment of the locations---are examples of trajectory-level similarity measures. Second, the \textit{dataset-level} similarity measures the statistical or distributional similarity over a trajectory dataset. This type of measure aims to capture how closely the generated trajectory dataset matches the statistical characteristics such as OD and route distributions in the real vehicle trajectory dataset. In this section, we present the modeling framework of TrajGAIL, where the procedure of driving in a road network is formulated as a partially observable Markov decision process to generate realistic synthetic trajectories, taking into account the similarities defined above.

% First, the \textit{element-wise} similarity is defined as the similarity between the $i^{th}$ elements between real and generated sequences. For instance, the probability of accurately predicting the next location based on the previously visited locations is one measure of element-wise similarity. 
% 

\subsection{Problem Formulation}\label{ProbForm}
% We define \textit{vehicle trajectory generation problem} as follows: 
% Generate a sample vehicle trajectory similar to vehicle trajectories in a given dataset based on the given vehicle trajectory dataset. 

% element-wise similarity

% trajectory-wise similarity

% dataset-wise similarity

% Here, the "similarity" can be interpreted as the trajectory element-wise trajectory, the overall similarity of a whole trajectory, and the distributional similarity over a trajectory dataset.

% Here, the similarity can be interpreted as not only sequence element-wise similarity, but also an overall similarity of a whole sequence and a distributional similarity over a trajectory dataset.

Let $Traj = \left\{(x_1,y_1,t_1), \cdots ,(x_N,y_N,t_N)\right\}$ be an urban vehicle trajectory, where $(x_i,y_i,t_i)$ is the $(x,y)$-coordinates and timestamp $t$ for the $i^{th}$ point of the trajectory, and $LocSeq = \left\{(x_1,y_1), \cdots ,(x_N,y_N)\right\}$ be the location sequence of $Traj$. When location points $(x,y)$ are continuous latitude and longitude coordinates, it is necessary to pre-process these coordinates and match them to a predefined set of discrete locations. Previous studies used different ways of defining discrete locations. For instance, \cite{choi2018network}, \cite{choi2019attention}, and \cite{ouyang2018non} used partitioned networks, so-called cells or zones, while \cite{choi2019real} and \cite{ziebart2008maximum} used road links to represent trajectories. In this paper, we represent a trajectory as a sequence of links to model link-to-link route choice behaviors in urban road networks. The location sequence of each vehicle trajectory is, thus, transformed to a sequence of link IDs by link matching function $f_M$: 

\begin{align}
    % & Traj =\big((x_1,y_1,t_1), ... ,(x_N,y_N,t_N)\big)  \equiv \big((x_1,y_1), ... ,(x_N,y_N)\big) = LocSeq \\
    % & LocSeq = \big((x_1,y_1), \cdots ,(x_N,y_N)\big) \equiv \big(l_1,\cdots,l_M \big) = LinkSeq &
    & LinkSeq = \big(l_1,\cdots,l_M \big) = f_M\big(LocSeq= \left\{(x_1,y_1), \cdots ,(x_N,y_N)\right\}\big) &
    % & Traj = \big((x_1,y_1,t_1), ... ,(x_N,y_N,t_N)\big) \equiv \big((l_1,t_1),...,(l_M,t_M) \big) = LinkSeq &
\end{align}

where $l_j$ is the link ID of the $j^{th}$ visited link along the trajectory. The goal of this study is to generate the link sequence of a trajectory by modeling and learning the probability distribution of $LinkSeq$, $P\big(LinkSeq\big) = P\big(L_1=l_1,\cdots,L_M=l_M \big)$ for a discrete random variable $L$ in all possible set of link IDs. Modeling this joint probability distribution is, however, extremely challenging, as also noted in the previous studies \cite{choi2018network,ouyang2018non}. A way to resolve this problem is to use a sequential model based on the Markov property, which decomposes the joint probability to the product of conditional probabilities as follows:

% \begin{equation}
% \begin{split}
% P\big(LinkSeq\big) & = P\big(l_1,\cdots,l_M \big) & \\
%                 & = P\big(l_M | l_{M-1} , \cdots , l_{1}\big) \times \cdots \times P\big(l_2| l_{1}\big) \cdot P\big( l_{1}\big) \\
%                 & = P\big(l_M | l_{M-1}\big) \times \cdots \times P\big(l_2 | l_{1}\big) \times P\big( l_{1}\big) 
% \end{split}
% \end{equation}

\begin{equation}
\begin{split}
P\big(LinkSeq\big) 
& = P\big(L_1 = l_1,\cdots,L_M=l_M \big)\\
& = P\big(L_M=l_M | L_{M-1} = l_{M-1} , \cdots , L_1=l_{1}\big) \times \cdots \times P\big(L_2=l_2| L_1=l_{1}\big) \cdot P\big( L_1=l_{1}\big) \\
& = P\big(L_M=l_M | L_{M-1}=l_{M-1}\big) \times \cdots \times P\big(L_2=l_2 | L_1=l_{1}\big) \times P\big( L_1=l_{1}\big) 
\end{split}
\end{equation}

%so it is necessary to simplify the problem with independent assumption. 
%To use the imitation learning framework, as mentioned in the Introduction, we formulate the problem as a Markov Decision Process (MDP).
%We formulate the vehicle trajectory generation in terms of a Markov Decision Process (MDP). 

The problem of modeling vehicle trajectories using this Markov property can be formulated as a Markov Decision Process (MDP). An MDP is a discrete-time stochastic control process based on the Markov property \cite{howard1960dynamic}. This process provides a mathematical framework for modeling sequential decision making of an agent. An MDP is defined with four variables: $(S,A,T,R)$, where $S$ is a set of states that the agent encounters, $A$ is a set of possible actions, $T(s,a,s')$ is a transition model determining the next state ($s' \in S$) given the current state ($s \in S$) and action ($a \in A$), and $R(s,a)$ is a reward function that gives the agent the reward value (feedback signal) of its action given the current state. If the transition is stochastic, transition model $T(s,a,s')$ can also be denoted as $P(s'|s,a)$. A policy $(\pi_{\theta})$ is defined as a $\theta$-parameterized function that maps states to an action in the deterministic case $(\pi_{\theta} (s) \rightarrow a)$, or a function that calculates the probability distribution over actions $\big(\pi_{\theta} (s) = P(a|s)\big)$ in the stochastic case. The objective of MDP's optimization is to find the optimal policy that maximizes the expected cumulative rewards, which is expressed as:

%The objective of the optimization of MDP is to find a good policy generator ($\pi$), which maximizes the expected cumulative rewards as follows:

\begin{align}
\label{eq:maxreward}
    & \pi_{\theta^*} = \arg \max_\theta \mathbf{E}\Big[\sum_{t=0}^{\infty} \gamma^{t} \cdot R(s,a)\Big] & 
    & a \sim Categorical \Big(\pi_{\theta} (s)=P(a|s) \Big)&
\end{align}
where $ \pi_{\theta^*}$ is the optimal policy with parameter $\theta^*$, $a$ is a sampled action from $\pi_{\theta} (s)= P(a|s)$ (in this study, we use discrete action space, so actions are sampled from categorical distribution), and $\gamma$ is the discount rate of future rewards.

%where $\pi(s)$ is the policy 
%where $a \sim \pi(s)$, and $\gamma$ is the discount rate. \\

How to define the four variables of MDP is critical to the successful training of a policy model. The states $(s)$ should incorporate enough information so that the next action $(a)$ is determined based only on the current state $(s)$, and the transition model $T$ should correctly reflect the transition of states in the environment it models. Finally, the reward function $R$ should give a proper training signal to the agent to learn the optimal policy.

In TrajGAIL, the vehicle movement in a road network is formulated as an MDP. We set road segments or links as states and transitions between links as actions. In this case, the transition model can be defined as a deterministic mapping function that gives the next link $(s')$ given the current link $(s)$ and the link-to-link movement choice $(a)$, i.e.,  $T:(s, a) \mapsto s'$. The policy represents a driver's route choice behavior associated with selecting the next link at each intersection. This road network MDP, thus, produces vehicle trajectories---more specifically, link sequences---as a result of sequential decision making modeled by this policy.
% $T:S\times A \rightarrow S$

As mentioned above, MDP assumes that the action $(a)$ is determined based only on the current state $(s)$. However, it is likely that vehicles' link-to-link movement choice is affected by not only the current location but also the previous locations. Moreover, vehicle movements in a road network is a result of complex interactions between a large number of drivers and road environment such as the generation and distribution of trips and the assignment of the routes and, therefore, the link choice action cannot be determined solely by road segment information alone as a state. The model needs more information such as origin, destination, trip purpose, and the prevailing traffic state. Incorporating all such information in the state definition, however, makes the problem intractable due to an extremely large state space. It is, thus, desirable to relax the assumption such that the action is determined based on the current state as well as some \textit{unobservable} states.

%Give more context about formulating vehicle movement in a road network as an MDP. Readers should understand this road network MDP before being introduced to POMDP. Give a simple example, e.g., road segments as states, transition between segments as actions, and a trajectory as sequential decision making in MDP. What does "policy" mean in this road network MDP
%Partially Observable MDP
%We have to define the states and actions of MDP. The easiest option would be selecting links as states. However, to properly select an adequate action, the model needs more information than just the current link such as origin, the purpose of the trip, and current time. The problem is that these variables may be unnoticeable in some instances. 

\begin{figure}
  \centering
  \includegraphics[width=0.8\textwidth]{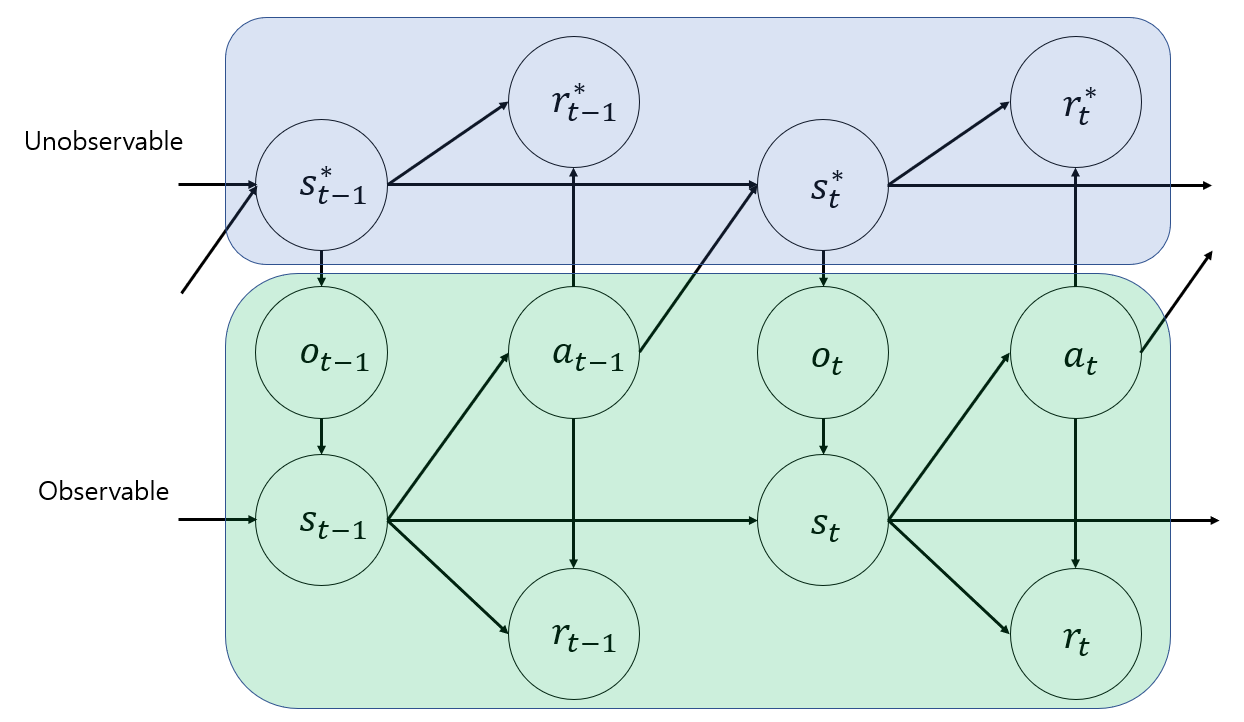}
  \caption{ Partially observable Markov Decision Process }
  \label{fig:pomdp}
\end{figure}

This can be achieved by employing a partially observable MDP (POMDP). A POMDP assumes that an MDP determines the model dynamics, but the agent cannot directly observe the underlying states. Instead of directly using the states as MDP, POMDP uses a surrogate state such as probability distribution over the set of possible states \cite{kaelbling1998planning} and belief state \cite{rao2010decision}. Figure \ref{fig:pomdp} shows the graphical model of POMDP. It is assumed that there exist latent unobservable states $s^* \in S^*$. We can only partially observe $s^*$ through observation $o \in O$. 
Using $o$ or sequence of $o$'s, latent state $s^*$ is estimated and this estimated latent state, or the belief state, is represented as $s \in S$.

% we have to guess the latent state. This guess, or the belief state, is represented in $s \in S$.

As a result, instead of four variables of MDP, POMDP uses five variables $(O, S, A, T, R)$, where $O$ represents the set of possible observations. The belief state at time state $t$, $s_t$, is estimated based on the sequence of $o$ representing all observations up to the current time $ t $, which is assumed to be the estimate of the latent unobservable state, $ \hat{s}^*_t $, as follows: 

% \begin{align}\label{eq:obtos}
%   & s_t=f(o_1,...,o_t) \qquad \qquad  s \in S, o \in O & \\
%   & s_t = \hat{s}^*_t \approx s^*_t
% \end{align}
% \hfill

\begin{align}\label{eq:obtos}
  & s_t=f(o_1,...,o_t)=\hat{s}^*_t \qquad \qquad  s \in S, o \in O
\end{align}

In TrajGAIL, the observation space, $O$, is defined as the ID of links in the road network and two virtual tokens representing the start and the end of a trip $(Start, End)$. Actions $A$ are transitions between links. In \cite{ziebart2008maximum}, the set of actions includes all possible link transitions.
% , which result in a large action space requiring high computational cost. 
However, this can lead to a very large action space even with a moderate-sizes network with hundreds of links, requiring high computational cost. To reduce the computational complexity, we instead define a set of common actions that represent possible movements between two connected links, namely, $[Straight, Left, Right, Terminate]$, where $Straight$, $Left$, and $Right$ represent the movement direction at the end of each link (at intersections) and $Terminate$ represents the termination of a trip (i.e., a vehicle reached its destination). These four actions are sufficient for our current study as we consider a grid-structured network, where all intersections are four-way intersections. However, it is also possible to model general networks with more diverse intersection structures such as five-way or T-shape intersections as we can apply a "mask" that allows flexibility to further define specific actions available for each link, which would be a subset of the network-wide common action set. For instance, one can define six actions for a network with the maximum intersection size of six and specify only a subset of available actions for each link if it has less than six connected roads.

To summarize, we formulate a partially observable Markov Decision Process to develop a generative model for vehicle trajectories, which produces the optimal policy describing optimal actions given a sequence of observations. 

\subsection{Model Framework}\label{sec:framework}

\subsubsection{Preliminaries and Background - Imitation Learning}
% To solve the vehicle trajectory generation problem represented in POMDP formulation, we use the imitation learning framework. 
In this study, the imitation learning framework is used to develop a generative model represented in POMDP formulation. Imitation learning is a learning problem that aims to train a model that can act like a given expert. Usually, demonstrations of decisions of the expert are given as a training dataset. In this study, a real vehicle trajectory dataset serves as expert demonstrations so that the model learns the decision-making process of vehicle movements in a road network observed in the given dataset.
%
%Imitation learning is a learning problem that aims to train a model that can perform a given task based on expert demonstrations. 
%
There are mainly two categories of approaches in imitation learning: \textit{behavior cloning} and \textit{inverse reinforcement learning}. 

Behavior cloning considers the imitation learning problem as a supervised learning problem. In behavior cloning, given the expert demonstrations, the state and action sequence is divided into independent state-action pairs and a model is trained to directly learn the relationship between input (state) and output (action) based on these sample pairs. The biggest advantage of behavior cloning is simplicity. However, because of its simplicity, the model fails to make proper generalization in complex tasks. Simple generative models based on Markov Chain \cite{gambs2010show} and Recurrent Neural Networks \cite{choi2018network,choi2019attention,liu2016predicting} can be classified into this category of imitation learning.

The inverse reinforcement learning (IRL) uses an indirect approach.
The objective of IRL is to find the reward function that the agent is optimizing given the measurements of agents' behavior and sensory inputs to the agents \cite{russell1998learning}.
It is assumed that the experts follow certain rules known as a reward function. The main idea of IRL is to learn this reward function to imitate the experts based on the history of experts' behaviors in certain situations. It is called "inverse" reinforcement learning because it learns the reward function that represents the experts' decisions from their states and actions, whereas the reinforcement learning (RL) learns to generate states and actions from a given reward function. Some of the key papers on IRL problems include \cite{ng2000algorithms, abbeel2004apprenticeship, ziebart2008maximum, wulfmeier2015deep, ho2016generative}, which readers are referred to for more details on IRL.

Given expert policy $\pi_E$, the objective 
% function
of IRL is 
% represented as follows:
to find a reward function ($r$) that maximizes the difference between the expected rewards from the expert and the RL agent ($\mathbf{E}_{\pi_E} [r(s,a)] - RL(r)$) such that the expert performs better than all other policies \cite{ho2016generative}, where an expectation with respect to a policy, $\mathbf{E}_{\pi} [r(s,a)]$, is used to denote an expectation with respect to the trajectory it generates (i.e., $\gamma$-discounted cumulative reward), $\mathbf{E} \Big[\sum_{t=0}^{\infty} \gamma^t r(s_t,a_t) \Big]$. This is achieved by minimizing the expected reward from RL agent ($RL(r)$) and by maximizing the expected reward from the expert ($\mathbf{E}_{\pi_E} [r(s,a)]$), while minimizing the reward regularizer ($\psi(r)$). On the other hand, when a reward function ($r$) is given, the objective of RL is to find a policy ($\pi$) that maximizes the expected reward ($\mathbf{E}_\pi[r(s,a)]$) while maximizing the entropy of the policy ($H(\pi)$).

\begin{align}\label{eq:irl}
    & IRL(\pi_E) = \underset{r \in \mathbf{R}}{\arg\min} \Big( \psi(r) +  RL(r) - \mathbf{E}_{\pi_E} [r(s,a)]  \Big) &
    %\arg\max_{r \in R} \mathbf{E}_{\pi_E} [r(s,a)] - RL(c) 
\end{align}

\noindent
where,

\begin{align}
    & RL(r) =   \max_{\pi \in \Pi} \Big( H(\pi) + \mathbf{E}_\pi[r(s,a)] \Big) &
  \label{eq:rl}
\end{align}

\noindent
where 
% $r$ is a reward function in the state-action dimension $(\mathbf{R}^{S \times A})$, 
$\mathbf{R}$ is the largest possible set of reward functions ($\mathbf{R} = \{ r:S \times A \longrightarrow \mathbb{R} \} $), 
% $\mathbf{E} [r(s,a)]$ denotes the $\gamma$-discounted reward $\mathbf{E} \Big[\sum_{t=0}^{\infty} \gamma^t r(s_t,a_t) \Big]$,
$\psi(r)$ is the convex reward function regularizer, and $H(\pi) = \mathbf{E}_\pi [- \log \pi(a|s)]$  is the causal entropy of the policy $\pi$ \cite{ho2016generative}.

It is interesting to investigate the relationship between $RL$ and $IRL$ in Eq. (\ref{eq:irl}) and Eq. (\ref{eq:rl}). $RL$ tries to find the optimal policy $\pi$ that maximizes the expected rewards, and $IRL$ tries to find the optimal reward function that maximizes the difference between expert policy ($\pi_E$) and $RL$'s policy ($\pi$). In some sense, $RL$ can be interpreted as a generator that creates samples based on the given reward, and $IRL$ can be interpreted as a discriminator that distinguish the expert policy from $RL$'s policy. This relationship is similar to the framework of Generative Adversarial Networks (GAN). GANs use an adversarial \textit{discriminator} ($D$) that distinguishes whether a sample is from real data or from synthetic data generated by the \textit{generator} ($G$). The competition between generator and discriminator is formulated as a minimax game. As a result, when the model is converged, the optimal generator would produce synthetic sample data similar to the original data. Eq. (\ref{eq:gan}) shows the formulation of minimax game between $D$ and $G$ in GANs.

% $RL(r)$ works as a generator and the optimal reward function $IRL(\pi_E)$ works as a discriminator. These are similar to the generative adversarial networks proposed by \cite{goodfellow2014generative}, which train a generative model $G$ by solving minimax game with a discriminative classifier $D$.

\begin{align}\label{eq:gan}
    & \min_G \max_D \Big( \mathbf{E}_{x \sim p_{data}(x)} \big[\log D(x)\big] + \mathbf{E}_{z \sim p_{z}(z)} \big[\log \left(1-D(G(z)) \right) \big] \Big) &
\end{align}

With a proper selection of the regularizer $\psi(r)$ in \textit{IRL} formulation in Eq. (\ref{eq:irl}), \cite{ho2016generative} proposed \textit{generative adversarial imitation learning} (GAIL). The formulation of minimax game between the discirminator ($D$) and the policy ($\pi$) is shown in Eq. (\ref{eq:gail})

% the  can be transformed into the generative adversarial network formulation as shonw in Eq. (\ref{eq:gail}). 

\begin{align}
    & \min_D \max_\pi \Big( \mathbf{E}_{\pi} \big[\log D(s,a)\big] + \mathbf{E}_{\pi_E} \big[\log \left(1-D(s,a) \right) \big] - \lambda H(\pi) \Big) &
  \label{eq:gail}
\end{align}

Eq. (\ref{eq:gail}) can be solved by finding a saddle point $(\pi, D)$. To do so, it is necessary to introduce function approximations for $\pi$ and $D$ since both $\pi$ and $D$ are unknown functions and it is very difficult, if not impossible, to define a exact function form for them. Nowadays, deep neural networks are widely used for function approximation. By computing the gradients of the objective function with respect to the corresponding parameters of $\pi$ and $D$, it is possible to train both generator and discriminator through backpropagation. In the implementation, we usually take gradient steps for $\pi$ and $D$ alternatively until both networks converge.

While GAIL provides a powerful solution framework for synthetic data generation, the original GAIL model \cite{ho2016generative} could not be directly used for our problem of vehicle trajectory generation. From our experiments, we found that the standard GAIL tends to produce very long trajectories with many loops, indicating that vehicles are constantly circulating in the network. This is because the generator in GAIL tries to maximize the expected cumulative rewards and creating a longer trajectory can earn higher expected cumulative rewards as there is no penalty of making a trajectory longer.

There are several ways to address this issue. Possible approaches include giving a negative reward whenever a link is visited to penalize a long trajectory or using positional embedding (the number of visited links) as state. However, a better approach would be to let the model know the vehicle's visit history (a sequence of links visited so far) and learn that it is unrealistic to visit the same link over and over. Our proposed TrajGAIL framework achieves this and addresses the limitation of GAIL in trajectory generation by assuming POMDP and using RNN embedding layer.

\subsubsection{TrajGAIL: Generative Adversarial Imitation Learning Framework for Vehicle Trajectory Generation}\label{sec:trajgail}

\begin{figure}
  \centering
  \includegraphics[width=\textwidth]{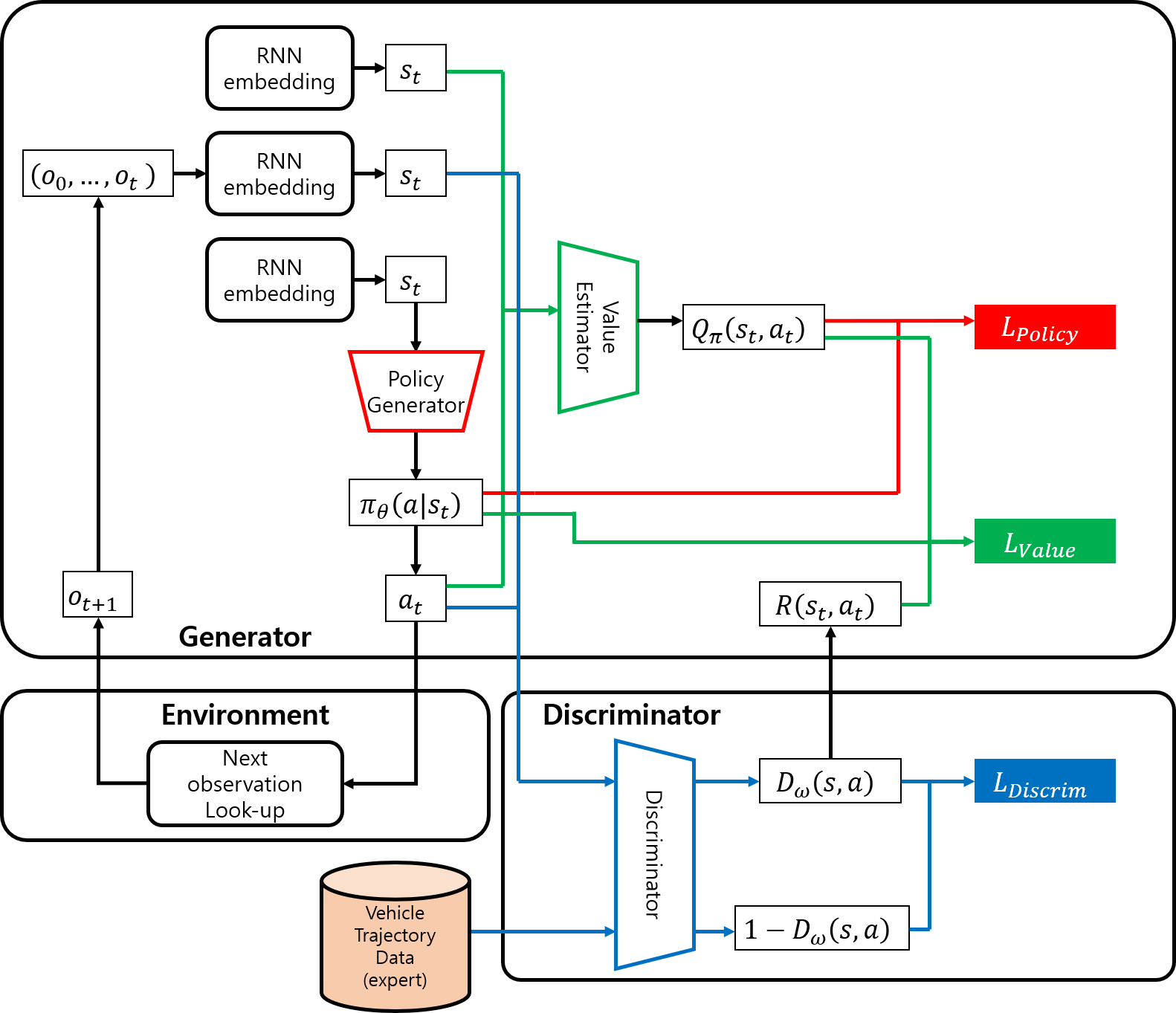}
  %\caption{Generative adversarial imitation learning framework for vehicle trajectory generation.}
  \caption{The model framework of TrajGAIL}
  \label{fig:gailframework}
\end{figure}

TrajGAIL uses POMDP to formulate vehicle trajectory generation as a sequential decision making problem and GAIL to perform imitation learning on this POMDP to learn patterns in observed trajectories to generate synthetic vehicle trajectories similar to real trajectories. 
In vehicle trajectory generation, it is important to take into account not only the previous locations of the trajectories, but also the expected future locations that the trajectory is expected to visit. By using POMDP, TrajGAIL considers how realistic the previously visited locations are. By using GAIL, TrajGAIL can also consider how realistic future locations would be because the imitation learning framework in GAIL uses an objective function to maximize the expected cumulative future rewards when generating new actions, which captures how realistic the remaining locations will be.
It is noted that, in this study, TrajGAIL focuses on generating location sequences (link sequences) of trajectories without considering time components. Throughout the paper, we use the term \textit{trajectory generation} to refer to the generation of link sequences representing trajectory paths for the sake of brevity. Figure \ref{fig:gailframework} shows the model framework of TrajGAIL. As in GAIL, TrajGAIL consists of the discriminator and the generator, where the discriminator gives reward feedback to the vehicle trajectories generated by the generator until both converge. The generator works as a reinforcement learning agent, and the discriminator works as an inverse reinforcement learning agent. Below we provide more details on each of these two modules.

\textbf{The Generator of TrajGAIL}. The primary role of the generator is to make realistic synthetic vehicle trajectories. The generator creates $N$ trajectories by a policy roll-out, or an execution of a policy from initial state ($Start$ of trip) to terminal state ($End$ of trip). A trajectory starts with the virtual token $Start$. By sequentially applying the policy generator until the current observation reaches the other virtual token $ End $, the generator produces a whole vehicle trajectory. As our problem is formulated as a POMDP, we need to map the sequence of observations into the latent states. \cite{rao2010decision} suggested that the belief states ($s_t$), the estimate of probability distribution over latent states, can be computed recursively over time from the previous belief state. The posterior probability of state $i$ at the $t$-th observation, denoted by $s_t(i)$ can be calculated as follows:

\begin{equation}
\begin{split}
    \label{eq:transition}
    s_t(i) & = P(s^*_t=i|o_t,a_{t-1},o_{t-1},\cdots,a_0,o_0) \\
          & \propto  P(o_t|s^*_t=i) \cdot P(s^*_t=i|a_{t-1},o_{t-1},\cdots,a_0,o_0) \\
          & \propto  P(o_t|s^*_t=i) \cdot \sum_j T(j,a_{t-1} , i) s_{t-1}(j) 
    % & s_t(i) \propto P(o_t|s^*_t=i) \sum_j T(j,a_{t-1} , i) s_{t-1}(j) &
\end{split}
\end{equation}

\noindent
where $s^*_t$ is the latent states at the $t$-th observation including both observable and unobservable variables, $P(o|s^*)$ is the probability of observation $ o $ given $s^*$, and $T(s, a, s')$ is the transition model that maps the current state ($s$) and the action ($a$) to the next state ($s'$).

Eq. (\ref{eq:transition}) indicates that the current belief state vector $(s_t)$ is a combination of the information from the current observation $(P(o_t|s^*_t))$ and the feedback from the previous computation of the belief state $(s_{t-1})$. In \cite{rao2010decision}, the author recognized the similarity between the structure of this equation and recurrent neural networks (RNN) and suggested using RNN for belief state estimation. Many previous studies on the next location prediction problem suggest that RNNs show great performance in embedding the sequence of locations into a vector \cite{choi2018network, choi2019attention, feng2018deepmove}. Accordingly, we use an RNN embedding layer to map the sequence of observations (link IDs) to a belief state vector. Since the entire historical sequence is embedded in the current (belief) state via RNN embedding and the actions are still determined based only on the current state, the Markov assumption in MDP is not violated, while sequential information can be effectively captured within the model.
In the implementation of RNN embedding layer, the size of input tensor (observation sequence) is $[B\times L]$ and the size of output tensor (belief state vector) is $[B \times H]$, where $B$ is the batch size, $L$ is the maximum observation sequence length in the batch, and $H$ is the number of hidden neurons.

Based on the belief state vector $(s_t)$, the \textit{policy generator} within the TrajGAIL generator module calculates the probability of the next action $(\pi (a|s_t))$. The policy $\pi (a|s_t)$ has a size of $[B \times A]$, where $A$ is the size of action space. The next action is sampled from a multinomial distribution with the probability $(\pi (a|s_t))$. 
The next observation is determined by the \textit{next observation look-up} table of the road network environment, $T_o(o_t,a_t,o_{t+1})$.
The next observation loop-up table maps the current observation ($o_t$) and the action ($a_t$) to the next observation ($o_{t+1}$). 
This should be defined based on the map geometry data which contains information on connections between links.
This process continues until the current observation reaches the virtual token $End$. 
% add explanation on next observation loop-up (it is map info)

In reinforcement learning, a \textit{value function} is often used to calculate the expected return of the actions at the current state. Here, we use a state-action value function $Q_{\pi} (s,a)$, which is estimated via the \textit{value estimator} in TrajGAIL's generator. The state-action value function, $Q_{\pi} (s,a)$, has size of $[B \times 1]$ since it represents a value scalar of each input observation sequence. The value estimator has a separate RNN embedding layer to process the sequence of observations into the belief state. Based on the processed belief state and a given action, the value estimator calculates the expected return of the action at current belief state. The estimated value, or the expected return, is used as a coefficient when updating the policy generator. If the estimated value of a given action is large, the policy generator model is reinforced to give the similar actions more often. This value estimator is also modeled as a deep neural network, which is trained to minimize the value objective function, $J_{Value}$, defined as follows:\\

\begin{equation}\label{eq:grad_value}
\begin{split}
J_{Value} & = \mathbf{E} \Big[ \big(   Q_{\pi} (s_t,a_t) - G(s_t,a_t)   \big)^2  \Big] \\
            & = \mathbf{E} \Big[ \big(   Q_{\pi} (s_t,a_t) - (R(s_t,a_t) + \gamma \cdot \mathbf{E} [R(s_{t+1},a_{t+1})] +\gamma^2 \cdot \mathbf{E} [R(s_{t+2},a_{t+2})] + ...   \big)^2  \Big] \\
            & = \mathbf{E} \Big[ \big(   Q_{\pi} (s_t,a_t) - (R(s_t,a_t) + \gamma (\mathbf{E} [R(s_{t+1},a_{t+1})] +\gamma \cdot \mathbf{E} [R(s_{t+2},a_{t+2})] + ...)   \big)^2  \Big] \\
            & = \mathbf{E} \Big[ \big(   Q_{\pi} (s_t,a_t) - (R(s_t,a_t) + \gamma \cdot \mathbf{E}[Q_{\pi} (s_{t+1},a_{t+1})]   )   \big)^2  \Big] \\
            & = \mathbf{E} \Big[ \big(   Q_{\pi} (s_t,a_t) - (R(s_t,a_t) + \gamma \cdot \sum \pi(a_{t+1} | s_{t+1})\cdot Q_{\pi} (s_{t+1},a_{t+1})  )   \big)^2  \Big] 
\end{split}            
\end{equation}

where a mean squared error (MSE) loss between the value estimate $(Q_{\pi} (s,a))$ and the actual $\gamma$ discounted return $(G(s,a))$ is used. 

The objective of the policy update is to maximize the expected cumulative reward function as shown in Equation (\ref{eq:maxreward}). We define the policy objective as $J(\theta)$ and we maximize the policy objective to improve the policy generator at every iterations. 
In order to compute the gradient of the policy objective, 
we use the Policy Gradient Theorem \cite{sutton2000policy}.
% to update the policy generator. 
In the Policy Gradient Theorem, 
for any differentiable $\theta$-parameterized policy $\pi_\theta$, the policy gradient of policy objective $\nabla_{\theta} J_{PG}(\theta)$ is given as:

\begin{align}
    & \nabla_{\theta} J_{PG}(\theta) = \mathbf{E} \Big[ \nabla_\theta \log \pi_{\theta} (a|s) \cdot Q_{\pi_\theta} (s,a) \Big] = \nabla_\theta J_{Policy} &
\end{align}

\noindent
where the last equality indicates that the gradient of $J_{PG}$ is equal to the gradient of $J_{Policy}$ given by $J_{Policy} =  \mathbf{E} \Big[ \log \pi_{\theta} (a|s) \cdot Q_{\pi_\theta} (s,a) \Big]$ although $J_{PG}$ and $J_{Policy}$ themselves are different. As their gradients are the same, we use $J_{Policy}$ in the training procedure.
% TODO
% gradient of J_PG = gradient J_Policy

Additionally, we add an entropy maximization objective \cite{ziebart2008maximum, ho2016generative} to the policy objective $J(\theta)$ in order to prevent the policy from converging to a local optimal policy. Often the global optimal policy is difficult to learn because of the sparsity of reward function. In this case, the policy would converge to a local optimal policy that only generates a limited variety of trajectories without considering the underlying distribution of actions. The entropy maximization objective helps counteract this tendency, guiding the policy to learn the underlying distribution of actions and eventually learn the underlying distribution of trajectories. As a result, we use the following equation to update the parameters of the policy generator to maximize the policy objective. 

\begin{align}\label{eq:grad_policy}
    % & \nabla_{\theta} J_{Policy} = \mathbf{E} \Big[ \nabla_\theta \log \pi_{\theta} (a|s) \cdot Q_{\pi_\theta} (s,a) \Big] - \lambda \nabla_{\theta} H(\pi_\theta) &
    %
\begin{split}
    & J(\theta) = J_{PG}(\theta) + \lambda H(\pi_\theta) \\
    & \nabla_{\theta} J(\theta) = \nabla_{\theta} J_{PG}(\theta) + \lambda \nabla_{\theta} H(\pi_\theta)= \nabla_{\theta} J_{Policy} + \lambda \nabla_{\theta} H(\pi_\theta)\\
\end{split}
\end{align}

\textbf{The Discriminator of TrajGAIL}. The discriminator solves the classification problem by distinguishing real vehicle trajectories from generated vehicle trajectories. 
As the generator gets improved to create more realistic vehicle trajectories, the discriminator's ability to classify the generated trajectories from the real trajectories is also improved through iterative parameter updates and fine-turning.
% 
% As the generator improves to make more realistic vehicle trajectories, the discriminator makes an improvement to classify them from real ones. 
% 
This competition of two neural networks is the fundamental concept of the generative adversarial learning framework. In the discriminator update step, the samples from the real vehicle trajectory dataset are labeled as 0, and the samples from the generator are labeled as 1. For both real and generated vehicle trajectories, we put the sequence observation and the action taken at the last observation as an input, process the sequence of observation into a belief state through an RNN embedding layer, and calculate the probability ($D_\omega (s,a)$) that the given sequence of observations and the action are from the generator. The discriminator probability, $D_\omega (s,a)$, has size of $[B \times 1]$ similar to $Q_{\pi} (s,a)$. The parameters of the discriminator are updated to minimize the binary cross-entropy loss. The $\omega$-parameterized discriminator is updated to minimize the discriminator objective, $J_{Discrim}$, with the following gradient term.

\begin{align}\label{eq:discrim}
    & \nabla_{\omega} J_{Discrim} =  \Bigg[\mathbf{E}_{ (s,a) \sim \pi_{\theta} } \Big[ \nabla_{\omega} \log(D_{\omega} (s,a)) \Big] + \mathbf{E}_{(s,a) \sim \pi_{E}} \Big[ \nabla_{\omega} \log(1- D_{\omega} (s,a)) \Big] \Bigg] &
\end{align}

The discriminator gives the training signal to the generator through the reward function ($R(s,a)$) as shown in Figure \ref{fig:gailframework}. The generator is trained to maximize the binary cross-entropy loss of the discriminator. As the second term of Eq. (\ref{eq:discrim}) is irrelevant to the parameters of the generator, the objective of the generator is to maximize the first term of Eq. (\ref{eq:discrim}). As a result, the reward function is defined as follows:

\begin{align}
    & R(s,a) = - \log (D_{\omega}(s,a)) &
\end{align}

\textbf{RNN embedding layer}.
In Figure \ref{fig:gailframework}, it is noted that the policy generator, the value estimator, and the discriminator use their own RNN embedding layers to embed the sequence of observations to the latent states. Using a separate RNN embedding layer for each of these three modules enables each module to interpret the sequence of observations and update RNN parameters in such a way as to maximize its performance in the task given to the module. For example, the discriminator might have a different interpretation on the sequence states from the policy generator, because the discriminator might focus more on the information on the whole sequence such as trajectory length to execute a discriminative task, while the policy generator might focus more on the current position to decide the next link-to-link transition. 

It is possible to implement TrajGAIL with a single shared RNN embedding layer. We implemented and tested this structure, but the results showed that using a shared RNN embedding layer produced a high variance in the three objectives in the early stage of training, which can lead to the failure of training. This is because the three objectives are updating the RNN embedding layer in different directions so that updating one objective is affecting the other objectives in unintended ways. 
% The new structure with one RNN embedding layer can be viewed as an application of Multi-Task Learning (MTL) to our original structure. 

\begin{figure}
  \centering
  \includegraphics[width= \textwidth]{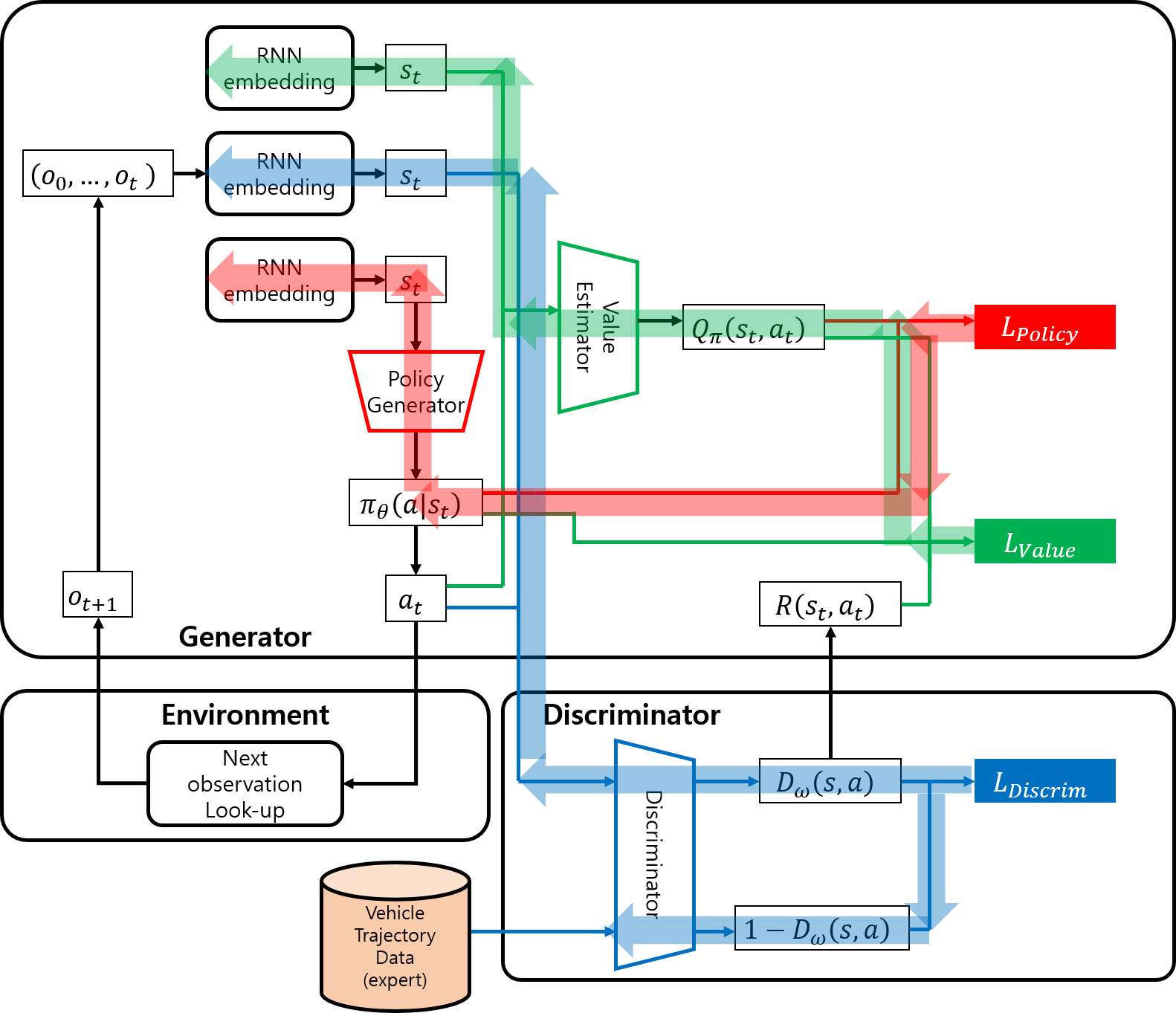}
  \caption{Backpropagation schema of TrajGAIL (red arrows for policy objective, green arrows for value objective, and blue arrows for discriminator objective)}
  \label{fig:gailbackprop}
\end{figure}

\textbf{Backpropagation}. Figure \ref{fig:gailbackprop} shows the schema of backpropagation to update parameters in the whole TrajGAIL framework. There are three different objective functions, $J_{Policy}$, $J_{Value}$, and $J_{Discrim}$, used to update parameters associated with policy generator, value estimator, and discriminator, respectively. Each objective is backpropagated to update only the related parameters in the deep neural networks. The backpropagation route of each objective function is indicated in different colors, i.e., red arrows for policy objective ($J_{Policy}$), green arrows for value objective ($J_{Value}$), and blue arrows for discriminator objective ($J_{Discrim}$). 
The backpropagation routes through the RNN embedding layers in Figure \ref{fig:gailbackprop} represent that each of these three modules updates its own RNN embedding layer. These RNN embedding layers play an integral role in combining the POMDP concept within the GAIL framework, which are the main distinction between TrajGAIL and stardard GAIL models.
% 
% 
% 
% Sharing the same RNN embedding layer can be interpreted as \textit{Hard Parameter Sharing} \cite{caruana1997multitask} in the context of MTL. Also, there is a \textit{Soft Parameter Sharing} MTL \cite{duong2015low, yang2016trace}, which has a regularization term to minimize the difference between the parameters of the shared component. 

\textbf{Training Techniques}. During the implementation of generative adversarial network algorithms, there are several techniques that can be used to avoid training failure and facilitate model convergence. The first technique used in this study is to maintain similar learning levels between the discriminator and the generator while training. If one overpowers the other, a proper competition cannot be formed and, thus, the model cannot learn from the competition mechanism of generative adversarial networks.
% If the discriminator is too strong compared to the generator, the discriminator error is too small and the discriminator loss generates gradients close to 0 and 1
% 
% If the discriminator is too strong compared to the generator, the generator may fail to learn as the discriminator error is too small. If the generator is too strong compared to the discriminator, the learning may also fail as the discriminator is too weak to teach the ability to generalize. 
Options to balance the learning levels include setting different learning rates and/or using different numbers of update-steps for training the generator and the discriminator. In this study, we used the same learning rate for both discriminator and generator and used a different number of update-steps. At each iteration, the generator is updated six times, while the discriminator is updated twice, because the discriminator usually learned faster in this study. The second technique is to use a sufficient number of trajectories generated for training. At each iteration, a certain number of trajectories are created and used to update the parameters of the model. If the number of generated trajectories is small, the model may result in a problem called "mode collapse." The mode collapse is defined as a case where the generator collapses, producing a limited variety of data \cite{dumoulin2016adversarially,lin2018pacgan}. Sometimes the generator oscillates among a few data points without converging to the equilibrium. In our case, the mode collapse leads the generator to produce trajectories for a few specific routes only. Using a sufficient number of sample trajectories at each training iteration solved this problem in our case. The details of the sample size used in this study are provided in the description of the evaluation results below.

\section{Performance Evaluation}\label{sec:eval}

\subsection{Data}\label{sec:data}

The performance of TrajGAIL was evaluated based on two different datasets. The first dataset is a virtual vehicle trajectory dataset generated by a microscopic traffic simulation model, AIMSUN. The second dataset includes the data collected by the digital tachograph (DTG) installed in the taxis operating in Gangnam District in Seoul, South Korea.

\begin{figure}
  \centering
  \includegraphics[width=0.6\textwidth]{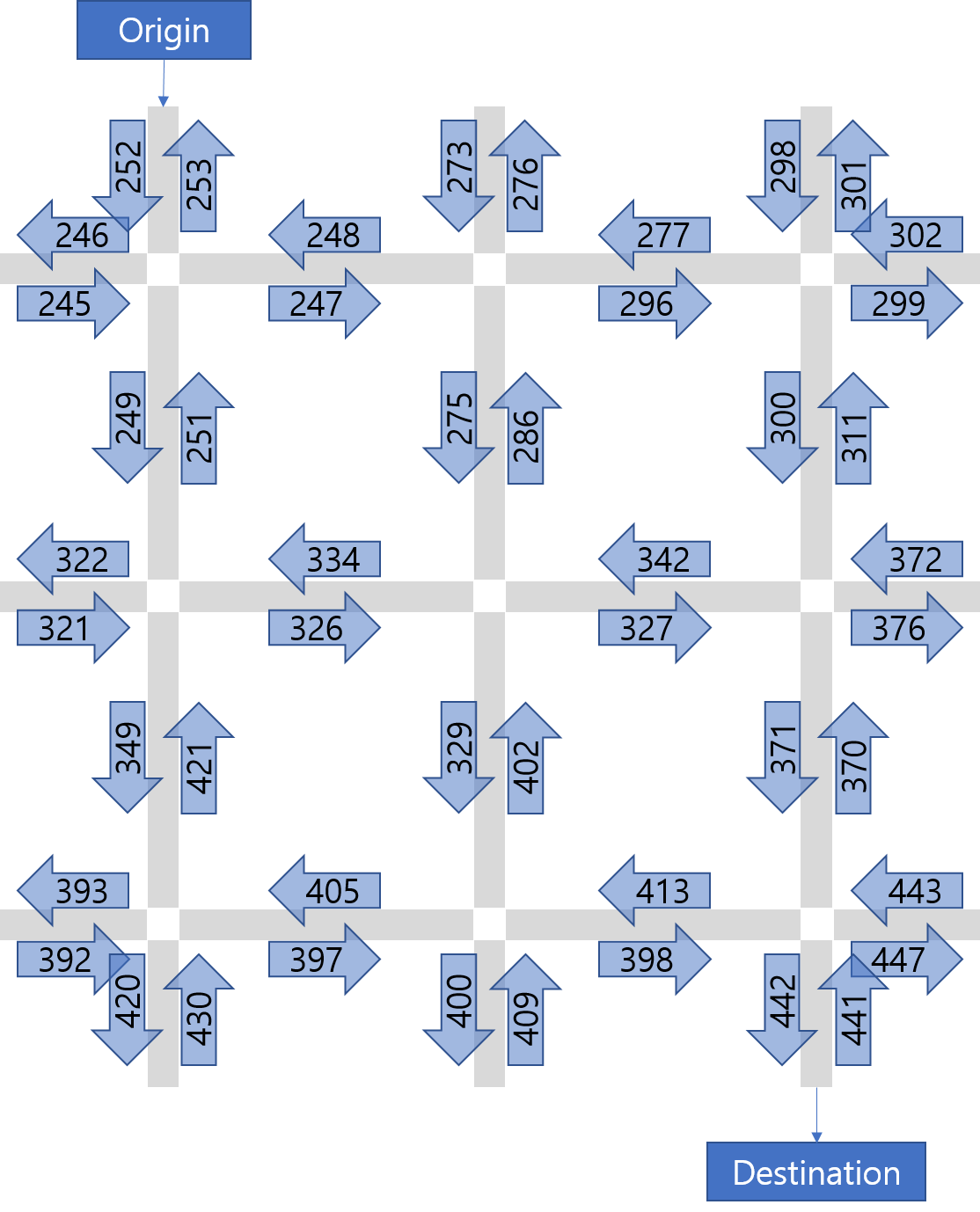}
      \caption{Simulated traffic network in AIMSUN environment. The origin and destination location is for Single-OD pattern}
  \label{fig:aimsun_single}
\end{figure}

AIMSUN uses dynamic traffic assignment \cite{barcelo2005dynamic} to select the appropriate route for each vehicle. We can select five different route choice models: Binomial, C-Logit, Proportional, Multinomial Logit, and Fixed. The first four algorithms use a predefined cost function and sample a route with the corresponding random distribution. The last algorithm only considers the travel time in free-flow condition and makes greedy choices, in which most vehicles use the route with shortest travel time for each OD.

The first dataset consists of data with three different demand patterns. The first demand pattern is called "\textbf{Single-OD}" pattern. The Single-OD pattern has only one origin source and one destination sink as shown in Figure \ref{fig:aimsun_single}. The origin source is connected to the Link 252, and the destination sink is connected to the Link 442. There are six possible shortest path candidates.

\begin{figure}
    \centering
    \begin{subfigure}{.4\textwidth}
    %   \centering
      % include first image
      \includegraphics[width=\textwidth]{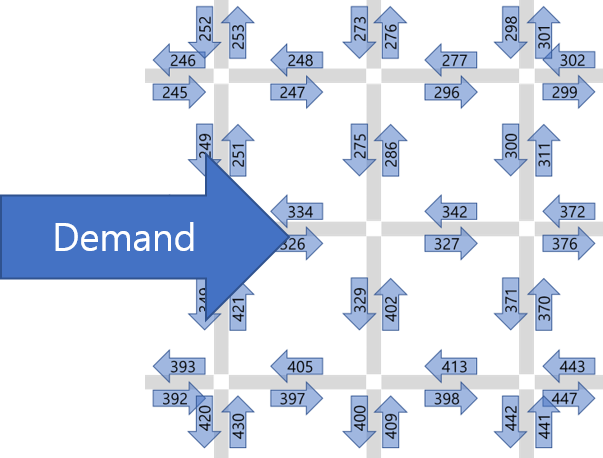}  
      \caption{One-way Multi-OD}
      \label{fig:aimsun_oneway}
    \end{subfigure}
    \hspace{6em}% Space between image A and B
    \begin{subfigure}{.45\textwidth}
      \centering
      % include second image
      \includegraphics[width=\textwidth]{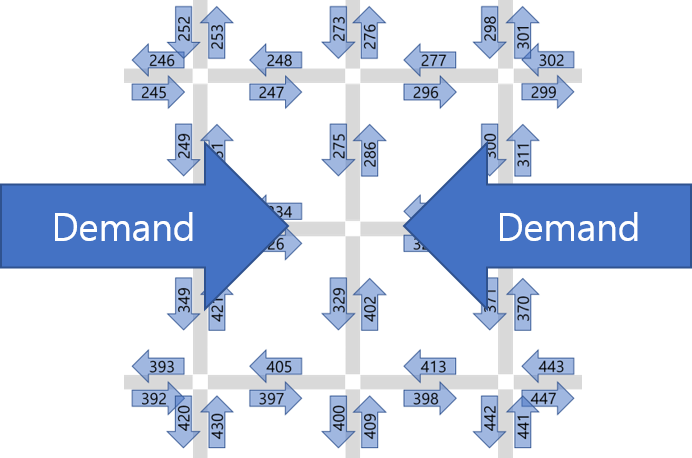}  
      \caption{Two-way Multi-OD}
      \label{fig:aimsun_twoway}
    \end{subfigure}
    \caption{Multi-OD demand patterns. Blue arrows indicates the major demand flows.}
    \label{fig:aimsun_multi}
\end{figure}

The second and third demand patterns use multiple origins and destinations. In these cases, the vehicle sources are connected to all 12 links that are at the boundary of the network, whose directions are towards the inside of the network (Link 252, 273, 298, 302, 372, 443, 441, 409, 430, 392, 321, 245). The remaining 12 boundary links, whose directions are towards the outside of the network, (Link 253, 276, 301, 299, 376, 447, 442, 400, 420, 393, 322, 246) are connected to the vehicle sinks. There can be 132 origin-destination pairs, excluding direct U-turns from the origin such as Link 245 to Link 246. The second demand pattern is called "\textbf{One-way Multi-OD}" pattern, where we assume there are major demand flows from the origin links on the left  (Link 245, 321, 392) to the destination links on the right (Link 299, 376, 447) as shown in the Figure \ref{fig:aimsun_oneway}. The major flows include all combinations of the origins (left) and destinations (right). The third demand pattern is called "\textbf{Two-way Multi-OD}" pattern, shown in Figure \ref{fig:aimsun_twoway}, where major demand flows from the origin links on the right (Link 302, 372, 443) to the destination inks on the left (Link 246, 322, 393) are added to the One-way Multi-OD pattern.

\begin{figure}
  \centering
  \includegraphics[width=0.8\textwidth]{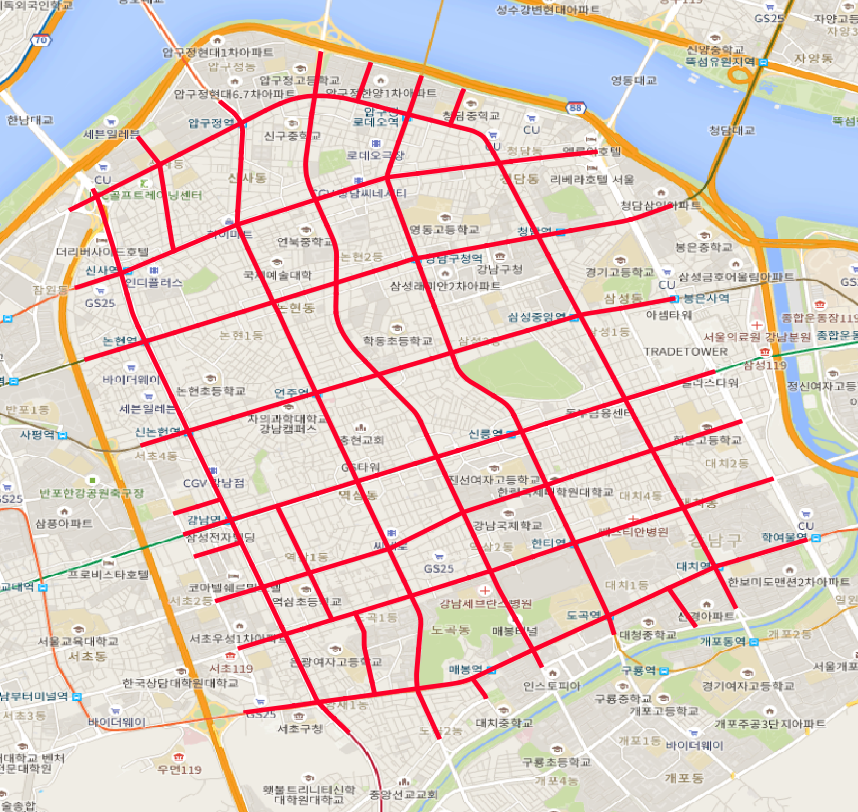}
      \caption{Major road links in Gangnam district (Map data ©2020 SK Telecom)}
  \label{fig:gangnam}
\end{figure}

The second dataset is collected by the DTG installed in taxis. The taxis operating in Seoul city are installed with digital tachographs (DTG) and collect the driving records such as position (longitude and latitude), speed, and passenger occupancy. By linking the data points with the same taxi ID in chronological order, the raw data points are transformed into a taxi trajectory dataset. Among the Seoul taxi trajectories, the trajectories that passed Gangnam district are selected for this study. Gangnam district has major road links in a grid structure as shown in Figure \ref{fig:gangnam}, so there are multiple choices in routes with similar travel distance for a given OD within the district.
% 
% 
% Then, we searched for taxis that passed Gangnam district to make a taxi trajectory dataset of Gangnam district. 
Since each taxi trajectory includes multiple trips associated different passengers, we extract a sub-trajectory covering the trip of each passenger and these passenger-level sub-trajectories are referred to as "\textbf{Gangnam}" dataset. The origin and destination of the vehicle trajectory in the Gangnam dataset represent the passenger OD demand in Gangnam district, and the routing patterns represent the route choice behaviors of the taxi drivers. As mentioned in Section \ref{ProbForm}, trajectory data need to be converted into link sequences. We use a map matching algorithm in \cite{brakatsoulas2005map} to map taxi trajectory data points to the underlying road links. After the data preparation, we obtained a total of 59,553 vehicle trajectories collected in Gangnam district on January 1st, 2018.

% We only used the main links (shown in Figure \ref{fig:gangnam}) in Gangnam district for simplicity.

\subsection{Baseline Models}\label{sec:baseline}

We tested TrajGAIL against three baseline models:

\begin{itemize}

% \subsubsection{Mobility Markov Chain}
\item Mobility Markov Chain

Mobility Markov Chain (MMC) \cite{gambs2010show} is one of the earliest models for the next location prediction problem based on the Markov model. MMC models each vehicle's trajectory as a discrete stochastic process, where the probability of moving to a next location depends on the immediate previous link observation. 

\begin{align}
    & P(o_{next}=i|o_{prev}=j) = \frac{N(o_{next} = i |o_{prev}=j)}{\sum_{\forall k \in O } N(o_{next} = k|o_{prev}=j)} &
\end{align}

\noindent
where $N(o_{next}=i|o_{prev}=j)$ is the number of occurrences in the real vehicle trajectory dataset that a vehicle moves from link $j$ to link $i$

\item{ Recurrent Neural Network model for Next Location Prediction}
% \subsubsection{ Recurrent Neural Network model for Next Location Prediction}

Several previous studies such as \cite{liu2016predicting, choi2018network, choi2019attention} suggest that recurrent neural networks show good performance in predicting the next location by learning spatio-temporal features of trajectory data. When a vehicle trajectory is given, RNN cells repeatedly process and calculate the hidden state. The RNN cells decide which information to keep and which to forget. The RNN cell then calculates the probability of the next location. The cross-entropy loss is used to calculate the estimation error. When the RNN model is used as a generative model, an input vector, starting with the virtual token $Start$, is passed through RNN to compute the predictive probability over the possible next locations. One location is sampled with multinomial distribution, and the sampled next location is used as the next input vector. The procedure continues until the current location reaches the virtual token $End$ representing the end of the trip. In this study, we use Long Short Term Memory (LSTM) \cite{hochreiter1997long} for RNN cells.

% \begin{align}
%   \begin{flalign}
%     P(o_{t+1}|o_t,o_{t-1},...,o_1) = RNN(\hat(o)_{t+1}|o_t,o_{t-1},...,o_1)
%   \end{flalign}
% \end{align}
% \hfill
% %

% \noindent
% where $RNN(o_{t+1}|o_t,o_{t-1},...,o_1)$ is the probability calculated based on RNN 

\item{Maximum Entropy Inverse Reinforcement Learning} 
% \subsubsection{Maximum Entropy Inverse Reinforcement Learning} 

Maximum Entropy IRL (MaxEnt) \cite{ziebart2008maximum} is one of the most widely used IRL models. MaxEnt uses a probabilistic approach based on the principle of maximum entropy to resolve the ambiguity in choosing distributions over decisions. MaxEnt uses a linear reward function for simplicity and uses a training strategy of matching feature expectations between the observed expert policy and the learner's behavior. In \cite{ziebart2008maximum}, the feature expectation is expressed in terms of expected \textit{state visitation frequency}, meaning that the MaxEnt model calculates the expected number of visitation at each state (link in a road network in this study), and matches it with the actual number of visitation in the expert dataset. In this study, we extend the idea of matching state visitation to matching state-action visitations. We call the original MaxEnt model using state visitation frequency \textit{MaxEnt(SVF)} and the new one using state-action visitation frequency \textit{MaxEnt(SAVF)}.

\end{itemize}

%The feature expectation is expressed in terms of expected state visitation frequencies. The MaxEnt model is trained to minimize the average error on expected state visitation frequency. 
%The model creates $N$ trajectories similar to the policy roll-out of GAIL. 

\subsection{Results}\label{sec:result}

This section shows the training and testing results of the TrajGAIL and the baseline models. 
TrajGAIL and the baseline models are trained for AIMSUN and Gangnam datasets. And they  are tested in various aspects with different performance measures. Two different evaluation levels are defined to measure trajectory-level similarity and dataset-level similarity.
Section \ref{sec:training} shows the training result using convergence curves. Section \ref{sec:tra_eval} and Section \ref{sec:data_eval} show the testing result based on the trajectory-level similarity and dataset-level similarity.

\subsubsection{Training Procedure} \label{sec:training}
\textbf{Hyperparameters.}
For the AIMSUN-based datasets, we first generate 20,000 vehicle trajectories for each demand scenario. Then, we split the total dataset into \textit{training} and \textit{testing} datasets in 0.7:0.3 ratio. As a result, we obtain 14,000 vehicle trajectories for training and 6,000 vehicle trajectories for testing. For the Gangnam dataset, we used the same ratio of 0.7:0.3, and this makes 41,687 vehicle trajectories for training and 9,866 vehicle trajectories for testing. The training dataset is only used to train each model, and all the results in Section \ref{sec:tra_eval} and Section \ref{sec:data_eval} are based on the testing dataset.
As mentioned in \textbf{Training Techniques} in Section \ref{sec:trajgail}, it is important to use sufficient number of sample trajectories at each training iterations. We tested different numbers of sample trajectories for each dataset and concluded that 2,000 sample trajectories are enough for the Single-OD dataset and 20,000 sample trajectories are enough for the Multi-OD datasets and Gangnam datasets. For a proper comparison of the model's performance, all models generate 20,000 sample trajectories for both trajectory-level evaluation and dataset-level evaluation. 
The more details on the hyperparameters used for training TrajGAIL is shown in Table \ref{tab:hypparam}.

\begin{table}
\caption{Hyperparameters used for TrajGAIL}
\label{tab:hypparam}
\begin{center}
% \resizebox{\textwidth}{!}{
\begin{tabular}{l l}
\hline
Hyperparameter & Value  \\ \hline
Number of iterations & 20,000      \\
Number of samples & 20,000      \\
Number of discriminator updates & 2        \\
Number of generator updates & 6        \\
Number of hidden neurons in each layer & 64 \\
Number of layers in RNN embedding & 3 \\
Learning rate & 0.00005 \\
Discount rate of reward ($\gamma$ in Eq. (\ref{eq:grad_value})) & 0.95 \\
Entropy coefficient ($\lambda$ in Eq. (\ref{eq:grad_policy}))   & 0.01 \\
\hline
\end{tabular}
\end{center}
\end{table}

\begin{figure}[hb!]
  \centering
  \includegraphics[width=0.8\textwidth]{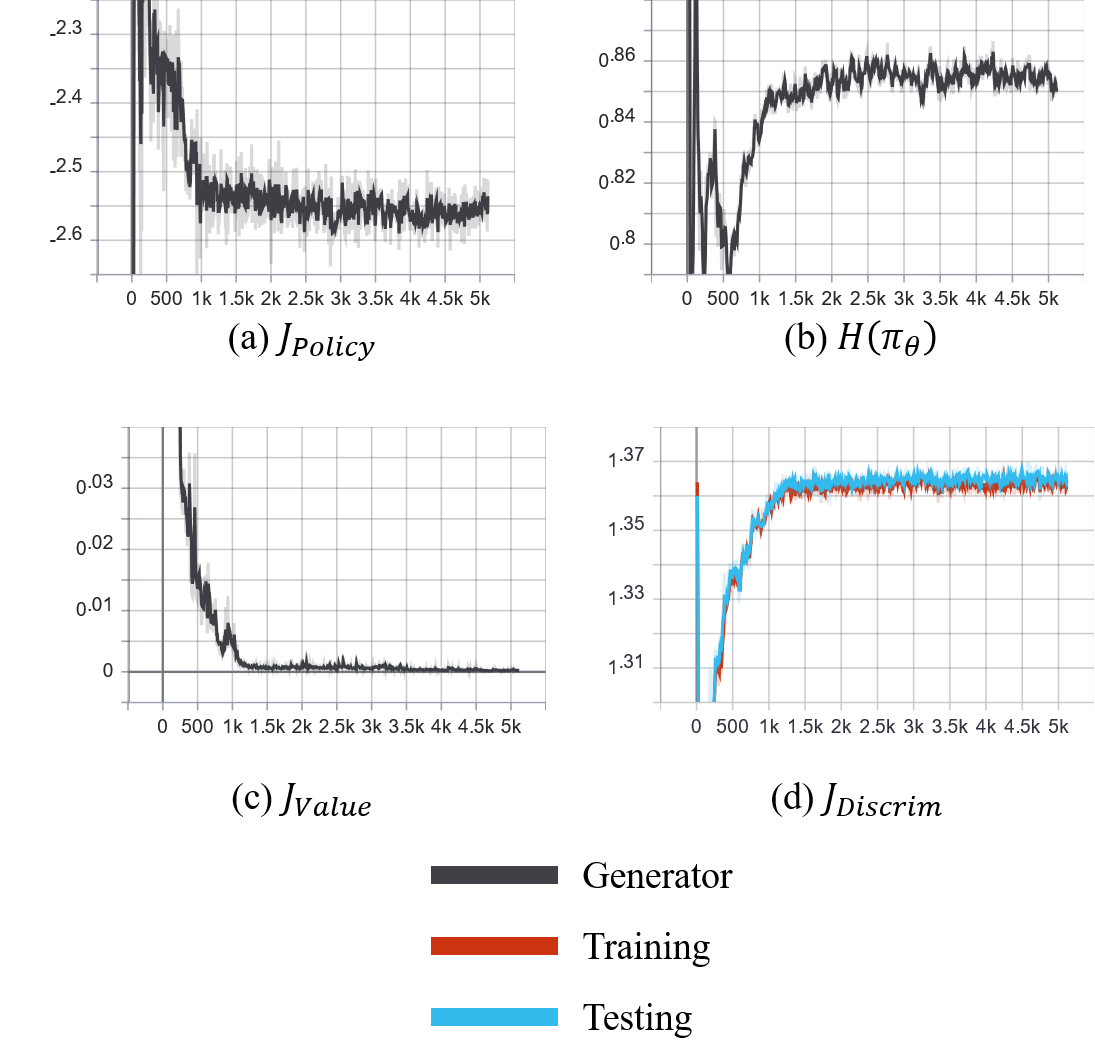}
  \caption{ Convergence curve of objective functions based on a sample case of "One-way Multi OD Binomial" using the AIMSUN dataset.}
  \label{fig:traintest}
\end{figure}

\textbf{Convergence Curve.}
Figure \ref{fig:traintest} shows the convergence curve of the objective functions ($J_{Policy}, J_{Value}, J_{Discrim}$) and the causal entropy ($H(\pi_\theta )$) for the "One-way MultiOD Binomial" case based on the AIMSUN dataset. At the very beginning of the iterations (0 to 100 iterations), $J_{Policy}$ increases and $J_{Discrim}$ decreases as the policy generator is designed to maximize $J_{Policy}$ and the discriminator is designed to minimize $J_{Discrim}$. As the policy generator starts to produce more realistic trajectories, it gets difficult to discriminate from the perspective of the discriminator. As the discriminator starts to distinguish the real trajectories from the generated trajectories, it gets difficult for the generator to generate more realistic trajectories. As a result, $J_{Policy}$ tends to decrease and $J_{Discrim}$ tends to increase in the middle of the iterations (100 to 1000 iterations). Afterwards, $J_{Value}$ is almost converged to zero, and the generator and the discriminator makes small changes to win the minimax game. The entropy $H(\pi_\theta )$ is converged to maximize the causal entropy at this point. It is noticeable that $J_{Discrim}$ is converged to a value close to 1.38. According to \cite{goodfellow2014generative}, the discriminator objective ($J_{Discrim}$) converges to $log4 = 1.38629\cdots$ as the generator produces realistic outputs. This is because the discriminator cannot distinguish the real trajectories from the generated trajectories and, thus, the best strategy becomes a random guess, which gives 50:50 chance of getting it right.

% The first one is trajectory-level evaluation and the second one is dataset-level evaluation.

\begin{figure}[hb!]
  \centering
  \includegraphics[width=0.8\textwidth]{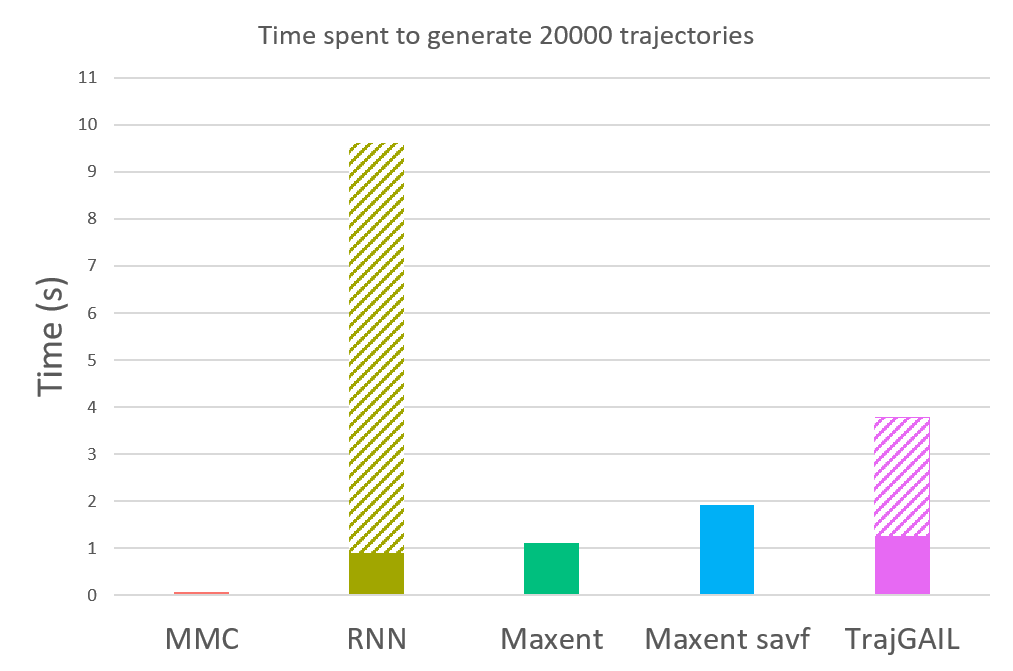}
  \caption{Computation time taken to generate 20,000 vehicle trajectories.}
  \label{fig:comptime}
\end{figure}

\textbf{Computation Time.} 
Figure \ref{fig:comptime} shows the computation time taken to generate 20,000 vehicle trajectories using five different models. This computation time is measured on a workstation with i9-10900KF CPU, 64GB RAM, and Nvidia Geforce RTX 3080. MMC and MaxEnt models only used CPU when computation time is measured. The computation times of RNN and TrajGAIL are measured not only by using CPU, but also by using GPU. In Figure \ref{fig:comptime}, the shaded parts of the bar chart are the computation time gain by using parallel processing of GPU when using RNN and TrajGAIL. Overall, all five models have the capability to generate 20,000 vehicle trajectories in less than 2 seconds.

% 
% value
% 
% 

\begin{figure}[htb!]
  \centering
  \includegraphics[width=0.8\textwidth]{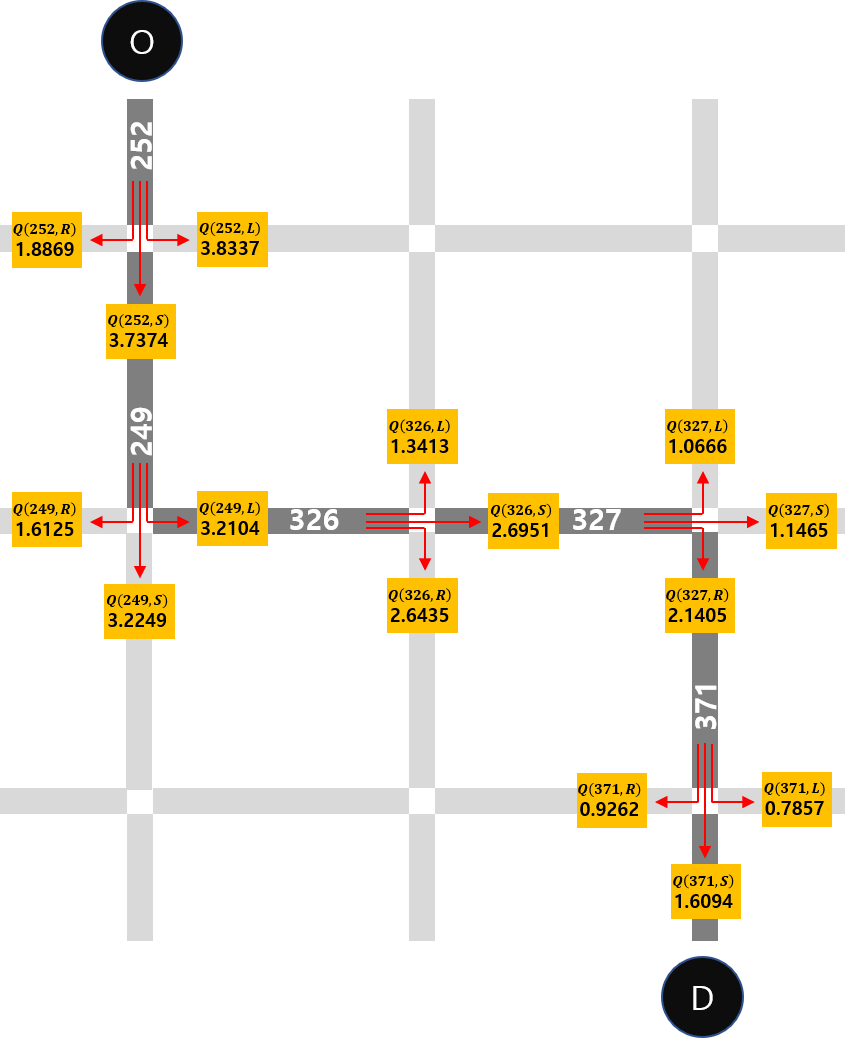}
  \caption{Estimated values of $Q_{\pi} (s,a)$ along a selected route in Single-OD Binomial AIMSUN dataset}
  \label{fig:values}
\end{figure}

\textbf{Visualizing Estimated Values.}
% need more explanation
To provide an intuition on how trajectories are generated in TrajGAIL, we visualize the estimated values of the state-action value function, $Q_{\pi} (s,a)$, in Figure \ref{fig:values}. Value function $Q_{\pi} (s,a)$ calculates the expected return of the actions at each state and these estimated values are used to update policy $\pi(a|s)$. During the training, the policy is shaped in such a way to select actions that have high estimated values.
Figure \ref{fig:values} shows the estimated values along one selected route in the \textbf{Single-OD Binomial} demand pattern in the AIMSUN dataset. All trajectories in the expert dataset have the same origin and same destination. There are six feasible routes that connect the given origin and destination, and Figure \ref{fig:values} is based on one of the feasible routes. Along the route, there are five decision points (the intersections to choose the next link) and, at each intersection, there are three estimated values associated with the three possible actions: \textit{Left} (\textit{L}), \textit{Straight} (\textit{S}), and \textit{Right} (\textit{R}). Note that state \textit{s} in $Q_{\pi} (s,a)$ represents link ID. Since $Q_{\pi} (s,a)$ is defined as the \textit{expected discounted cumulative reward}, which is the sum of reward over the decision points between the current state and the end of a trajectory, the value of $Q_{\pi} (s,a)$ tends to decrease as an intersection is closer to the destination because there are fewer decision points over which the reward is summed. As such, the comparison of the estimated $Q_{\pi} (s,a)$ values is meaningful only within each intersection, not across intersections.

% The average value at each intersection shows a decreasing tendency. This is because we use the \textit{expected discounted cumulative reward} as a value so that as the subject vehicle gets closer to the destination, the expected value decreases. As a result, the values should be compared among the branches in the same intersection.
% 
In Figure \ref{fig:values}, at the first intersection, the values of taking \textit{Straight} and \textit{Left} are higher than the value of taking \textit{Right}. This is because taking \textit{Right} here is unrealistic as there is no way that this next link is connected to the destination. On the other hand, taking \textit{Straight} and taking \textit{Left} have similarly high values because both choices are realistic (although \textit{Straight} is the chosen action in this particular route, \textit{Left} is also chosen in other feasible routes). At the second intersection, taking \textit{Straight} and taking \textit{Left} show higher values than taking \textit{Right} and, at the third intersection, \textit{Straight} and \textit{Right} show higher values than \textit{Left} (taking \textit{Left} at the third intersection makes some kind of detour which is unrealistic). Similarly, the values are higher for \textit{Right} at the fourth intersection and for \textit{Straight} at the fifth intersection, which are the realistic action at each respective intersection. Overall, this demonstrates that TrajGAIL can capture realistic actions at each intersection and translate this information to the value function and policy to generate realistic trajectories.

% However, when the computation time of RNN and TrajGAIL are measured, we measured them not only by only using CPU, but also by using CPU and GPU together.

\subsubsection{Trajectory-level Evaluation}\label{sec:tra_eval}

In the trajectory-level evaluation, we measure how similar each generated vehicle trajectory is to a real trajectory. Two widely used evaluation metrics in sequence modeling are used to evaluate this trajectory-level similarity: BLEU score \cite{papineni_bleu:_2002}  and METEOR score \cite{banerjee_meteor:_2005}.

% experiment setup
% BLEU & METEOR is tested against all trajectory in exp data
% relative comparison between models

In the next location prediction problem, it is common to use the probability of correctly predicting the next location to measure the model's performance.  For example, in \cite{choi2018network}, a complementary cumulative distribution function of the prediction probability is used to measure how accurately the model predicts the next one, two, or three consecutive locations. While this measure is intuitive and easy to interpret, it has  a drawback that it only considers element-wise prediction accuracy and does not take the whole sequence into account. The element-wise performance measures can be sensitive to small local mispredictions and tend to underestimate the model's performance.  As such, this study employs a BLEU score and METEOR score that consider the whole sequence. They are more robust and accurate as a performance measure for sequence modeling. 

BLEU is one of the most widely used metrics in natural language processing and sequence-to-sequence modeling. When reference sequences are given, BLEU scans through the sequence and checks if the generated sequence contains identical chunks, or a contiguous sequence of $n$ elements found in the reference sequences. Here, BLEU uses a modified form of precision to compare a reference sequence and a candidate generated sequence by clipping. For the generated sequence, the number of each chunk is clipped to a maximum count ($m_{max}$) to avoid generating the same chunks to get a higher score.

\begin{align}
\label{eq:bleu_precision}
& P_n=\frac{ \sum_{i \in C}   \min{(m_i,m_{i,max})}   }{w_t} &
\end{align}

\noindent
where $n$ is the number of elements considered as a chunk; $C$ is a set of unique chunks in the generated sequence; $m_i$ is the number of occurrences of chunk $i$ in the generated sequence; $m_{i,max}$ is the maximum number of occurrences of chunk $i$ in one reference sequence; and $w_t$ is the total number of chunks in the generated sequence.

${BLEU}_{n}$ score is defined as a multiplication of the geometric mean of $P_n$ and a brevity penalty. A brevity penalty is used to prevent very short candidates from receiving too high scores. 

\begin{align}
\label{eq:bleu}
& {BLEU}_{n} = min\Big(1, \frac{L_{gen}}{L_{ref,close}}  \Big) \cdot \Big(  \prod_{i=1}^{n} P_i  \Big)^{\frac{1}{n}} & 
\end{align}

\noindent
where $L_{gen}$ represents the length of the generated sequence, and  $L_{ref,close}$ represents the length of a reference sequence that has the closest length to the generated sequence.

% \subsubsection{METEOR score}
METEOR \cite{banerjee_meteor:_2005} is originally designed as an evaluation metric for machine translation. It can measure similarities in terms of both the occurrences of trajectory elements and the alignment of the elements in a trajectory. METEOR first creates an alignment matching between the generated sequence and the reference sequence. The alignment matching is a set of mappings between the most similar sequence elements. Since it is often used for natural language processing, the most similar sequence element refers to the exact match, synonyms, and the stems of words. In this study, it is difficult to define the "similar" observation and state, so we only use the exact match in the alignment matching. In alignment matching, every element in the candidate sequence should be mapped to zero or one element in the reference sequence. METEOR chooses an alignment with the most mappings and the fewest crosses (fewer intersections between mappings). Based on the chosen alignment, a penalty term is calculated as follows:

\begin{align}
\label{eq:meteor_penalty}
    & p=0.5 \Big( \frac{c}{w_{map}} \Big)^3 &
\end{align}

\noindent
where $c$ is the number of chunks of elements with no crossings, and $w_{map}$ is the number of elements that have been mapped. Then, we calculate the weighted harmonic mean between precision $P$ and recall $R$ with a ratio of the weights, 1:9.

\begin{align}
\label{eq:meteor_mean}
    & F_{mean}  = \frac{10}{\frac{1}{P} + \frac{9}{R}} = \frac{10PR}{R+9P} &
\end{align}

\noindent
where $P=\frac{m}{w_{gen}}$ and $R=\frac{m}{w_{ref}}$; $m$ is the number of elements in the generated sequence that is also found in the reference sequence; and $w_{gen}$ and $w_{ref}$ are the number of elements in the generated and reference sequence, respectively.

Finally, the METEOR score, $M$, is defined as follows:

\begin{align}
\label{eq:meteor}
    & M=F_{mean}(1-p) &
\end{align}

\begin{algorithm}
	\caption{Pseudo-code for score evaluation in trajectory-level evaluation}
\hspace*{\algorithmicindent} \textbf{Input:} $Tr$ (input trajectory)\\ 
\hspace*{\algorithmicindent} \hspace{0.45in} $D_{ref}$ (reference trajectory dataset)\\
\hspace*{\algorithmicindent} \hspace{0.45in} $f$ (score calculation function) \\	    
\hspace*{\algorithmicindent} \textbf{Output:} $Score$ \\
	\begin{algorithmic}[1]
	\State $Score$ $\gets$ 0
		\For {$Tr_{ref} \in D_{ref}$}
		\State $Score_{new}$ = $f(Tr, Tr_{ref})$ \Comment{calculate score between $Tr$ and $Tr_{ref}$}
		\If{$Score \leq Score_{new}$} \Comment{compare scores}
		\State $Score \gets Score_{new}$ \Comment{update score}
		\EndIf
		\EndFor
	\State\Return $Score$ \Comment{return score}
	\end{algorithmic} 
	\label{algorithm:trajgail_trajlevel}
\end{algorithm}

For both BLEU and METEOR, the higher the score, the better the model performance. For BLEU score, we use $n=4$ for Eq. (\ref{eq:bleu}).
% For BLEU, we used $n=4$.

For each model, 20,000 synthetic trajectories are generated for score evaluation. Unlike a supervised learning model, a generative model as in our case does not have a single ground-truth trajectory that can be matched with each generated trajectory. As such, we consider any observed expert trajectory in the training dataset a possible reference trajectory and evaluate if the generated trajectory matches any of the available reference trajectories.
More specifically, for each generated trajectory, scores are calculated against each reference trajectory in the training dataset. Then, we select the maximum value among the scores as a representative score for the given generated trajectory. As a result, a generated trajectory can be assigned a high score if it can be matched to any reference trajectory. Algorithm \ref{algorithm:trajgail_trajlevel} shows the pseudo-code for score evaluation.

\begin{figure}[!ht]
    \centering
    \begin{subfigure}{0.6\textheight}
      \centering
      % include first image
        \includegraphics[width=\textwidth]{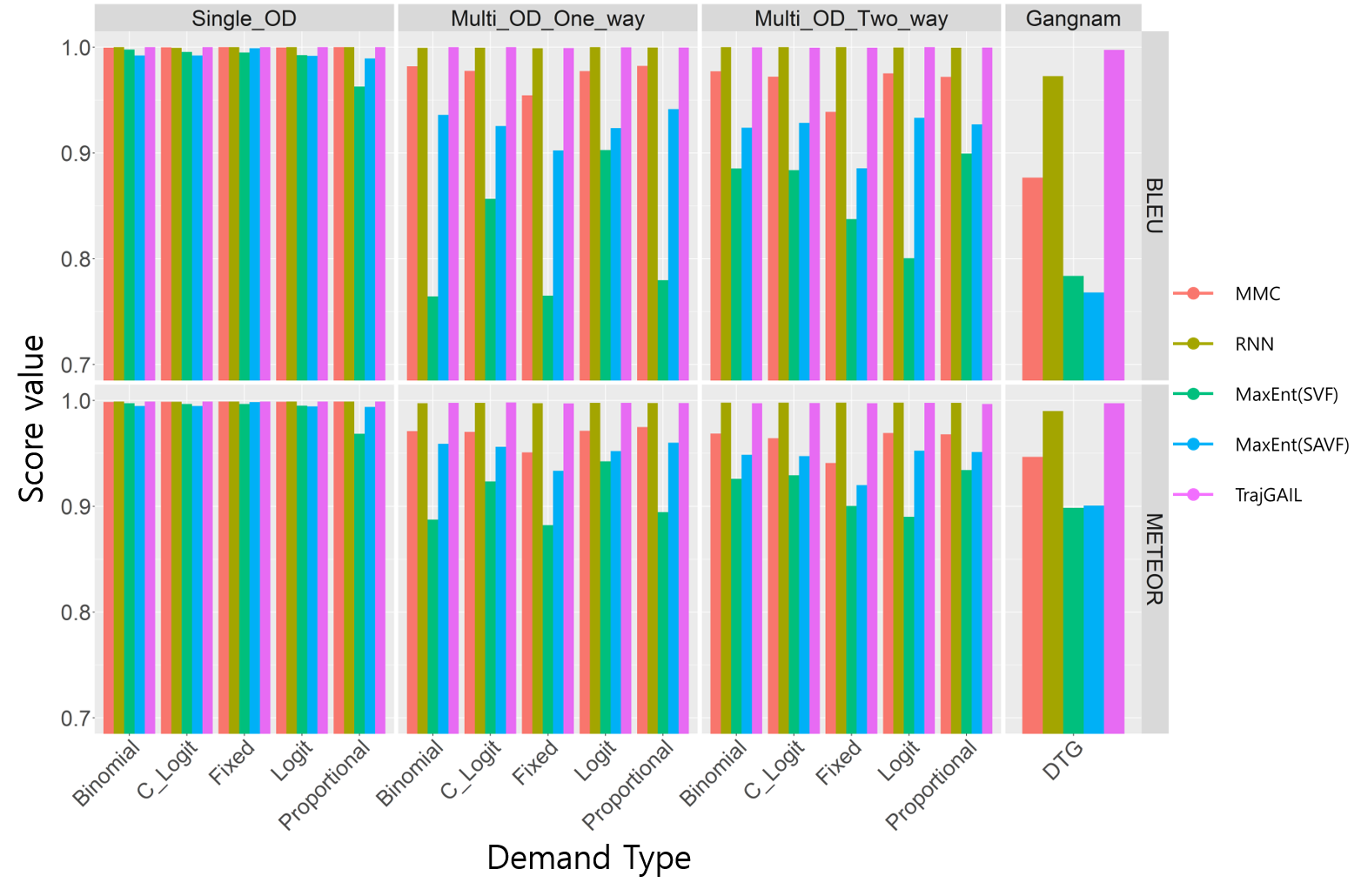}
        \caption{Average}
      \label{fig:avgscore}
    \end{subfigure}
    \par\bigskip % maximise vertical space here instead
    \begin{subfigure}{0.6\textheight}
      \centering
      % include second image
      \includegraphics[width=\textwidth]{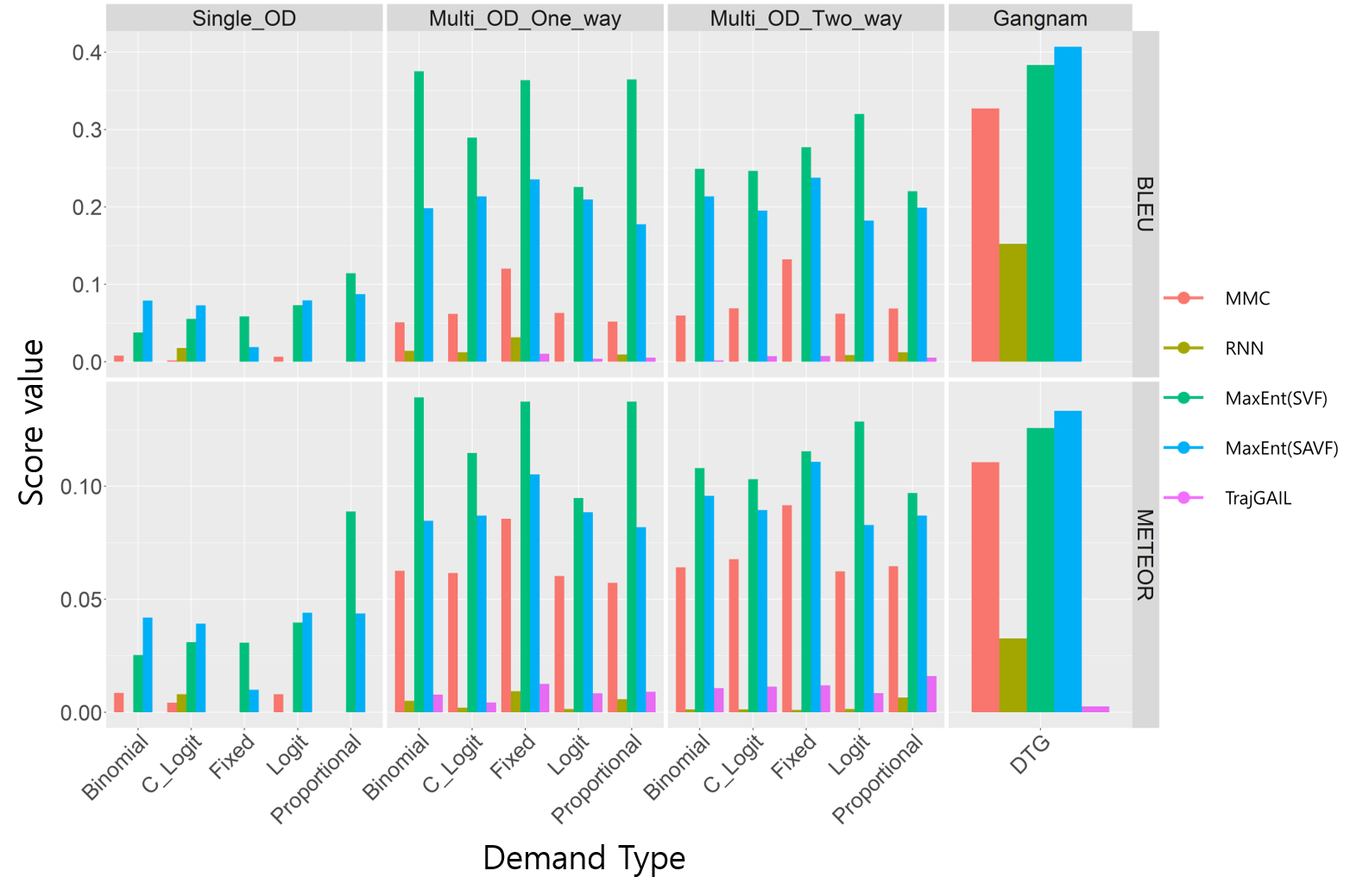}
    \caption{Standard deviation}
    \label{fig:sdscore}
    \end{subfigure}
    \caption{Result of BLEU and METEOR score of the generated vehicle trajectories of each model.}
    \label{fig:seqscore}
\end{figure}

% In Figure \ref{fig:seqscore}, the score results are shown in the box plots. The upper and lower bounds of each box represent the median and mean score results.
Figure \ref{fig:seqscore} shows the result of each model in different datasets and demand types. Figure \ref{fig:avgscore} shows the average score and Figure \ref{fig:sdscore} shows the standard deviation of the score result.
When the models are tested with the Single-OD datasets, the result shows that all five models show good performance in most cases. MMC, RNN, MaxEnt(SAVF), and TrajGAIL scored more than 0.99 in both BLEU and METEOR with all five different demand types. MaxEnt(SVF) showed the lowest score (0.9627) in "Proportional" demand.

% MaxEnt(SVF) shows some low scores compared to the other models, but all models have scores greater than 0.99 in different SingleOD demand patterns.

% As explored in Section \ref{sec:data_eval}, a
As the complexity of the dataset increases, the model's performance decreases. In Figure \ref{fig:seqscore}, MMC, MaxEnt(SVF), and MaxEnt(SAVF) show decreases in both scores when tested with the One-way Multi-OD and Two-way Multi-OD datasets. However, the scores of RNN and TrajGAIL (in the second and fifth bars within each test group) only slightly decrease. In fact, both models received perfect scores (i.e., BLEU = 1.0, METEOR = 1.0) except for few cases. In addition, when the models are tested with the Gangnam dataset, the average score of RNN and TrajGAIL is 0.9726 and 0.9974 for BLEU, respectively, and 0.9899 and 0.9974 for METEOR, respectively. Overall, the standard deviations of RNN and TrajGAIL are significantly lower than that of the other models, suggesting a higher level of robustness and lower fluctuations in model performance across different trajectory cases. The main reason for the good performance of RNN and TrajGAIL is that both models are capable of capturing sequential information in trajectories considering the history of multiple previously visited locations, as opposed to determining the next locations based only on the current location as in the other three models. However, it is observed that RNN sometimes generates trajectories traversing unknown routes (link sequences not found in the given dataset) and the relatively lower scores of RNN compared to those of TrajGAIL are attributed to instances of such unrealistic trajectories. 

\newpage
\newpage

\subsubsection{Dataset-level Evaluation}\label{sec:data_eval}
In the dataset-level evaluation, the statistical similarity between a generated trajectory dataset and a real trajectory dataset is assessed. There are many aspects of a dataset that can be considered for statistical similarity, such as the distributions of trajectory length, origin, destination, origin-destination pair, and route. Among these variables, route distribution is the most difficult to match since producing the similar route distribution requires matching all other variables, including the lengths, origins, and destinations of vehicle trajectories in a real dataset. As such, we use a measure of route distribution similarity to evaluate dataset-level model performance.

As with the trajectory-level evaluation, each model generates 20,000 vehicle trajectories to make a synthetic trajectory dataset. We first identified all the unique routes observed in the real dataset and counted their occurrences in the synthetic dataset. 
The synthetic trajectories that travel unknown routes (i.e., the routes that did not occur in the real dataset) were marked as "unknown" trajectories. The route frequencies are calculated by dividing the route counts by the total number of trajectories in a dataset. The route frequencies, or the routes' empirical probability distribution of the synthetic dataset is compared with that of the real dataset. In this study, we use Jensen-Shannon distance $(d_{JS})$ to measure the similarity of two route probability distributions.

Jensen-Shannon distance is a widely used distance metric for two probability distributions. Given two discrete probability distributions $p$ and $q$, the Jensen-Shannon distance $(d_{JS})$ is defined as follows:
\begin{align}
 \label{eq:js}
    & d_{JS} (p,q) = \sqrt{D_{JS} (p,q)} = \sqrt{\frac{D_{KL} \Big(p||\frac{p+q}{2}\Big) + D_{KL} \Big(q||\frac{p+q}{2}\Big)}  {2}} &
 \end{align}

\noindent
where $D_{JS}$ is the Jensen-Shannon divergence, and $D_{KL}$ is the Kullback-Leibler divergence. The Kullback-Leibler divergence from $q$ to $p$, $D_{KL} (p||q)$, is defined as:

\begin{align}
  \label{eq:kl}
    & D_{KL} (p||q) = E \big[ \log (p_i) - \log (q_i) \big]  = \sum_i p_i \log \frac{p_i}{q_i} &
\end{align}

$D_{KL} (p||q)$ is also known as the relative entropy of $p$ with respect to $q$. Since $D_{KL}$ is an asymmetric similarity measure, it cannot be used as a distance metric. As a result, $d_{JS}$, a symmetrized version of $D_{KL}$, is often used to measure the distance between two probability distributions. The value of $d_{JS}$ ranges from 0 to 1, where $d_{JS} = 0$ happens when the two probability distributions are identical  and $d_{JS} = 1$ happens when the two distributions are completely different.

% Jensen Shannon Divergence

\begin{table}
\caption{Jensen-Shannon distance ($d_{JS}$) of Route Distribution}
\label{tab:route}
\begin{center}
\resizebox{\textwidth}{!}{
\begin{tabular}{c c c c c c c}
\hline
Dataset & Demand Type & MMC  & RNN  & MaxEnt(SVF) & MaxEnt(SAVF) & TrajGAIL    \\ \hline
SingleOD& Binomial      & 0.0866 & \textbf{0.0606}  & 0.0903 & 0.0748 & 0.0916 \\
SingleOD& C-Logit       & 0.0381          & 0.0527   & 0.1145 & 0.0650 & \textbf{0.0275} \\
SingleOD& Proportional  & \textbf{0.0192} & 0.0599   & 0.2364 & 0.0568 & 0.0274     \\
SingleOD& Logit         & 0.0448          & 0.0526   & 0.1011 & 0.0683 & \textbf{0.0284}   \\
SingleOD& Fixed         & \textbf{0.0038} & 0.0153   & 0.0594 & 0.0490 & 0.0311 \\ 
\hline
One-way MultiOD & Binomial     & 0.2822  & 0.2446    & 0.5234 & 0.3813 & \textbf{0.2125} \\
One-way MultiOD & C-Logit      & 0.3032  & 0.2501    & 0.4666 & 0.3874 & \textbf{0.1987} \\
One-way MultiOD & Proportional & 0.2988  & 0.2604    & 0.5044 & 0.3825 & \textbf{0.2059} \\
One-way MultiOD & Logit        & 0.3375  & 0.2799    & 0.4531 & 0.3893 & \textbf{0.2163} \\
One-way MultiOD & Fixed        & 0.3763  & \textbf{0.1747} & 0.5529 & 0.4629 & 0.1791 \\ 
\hline
Two-way MultiOD & Binomial     & 0.3018  & 0.3005  & 0.5011  & 0.4042  & \textbf{0.2062}\\
Two-way MultiOD & C-Logit      & 0.3328  & 0.2587  & 0.4986  & 0.4409  & \textbf{0.2072}\\
Two-way MultiOD & Proportional & 0.3430  & 0.2739  & 0.4801  & 0.4388  & \textbf{0.2090}\\
Two-way MultiOD & Logit        & 0.3375  & 0.2833  & 0.5815  & 0.4337  & \textbf{0.2021}\\
Two-way MultiOD & Fixed        & 0.3763  & 0.1783  & 0.5694  & 0.4981  & \textbf{0.1694} \\ 
\hline
Gangnam & DTG                 & 0.4701  & 0.4823  & 0.7098 & 0.5558  & \textbf{0.4230} \\ 
\hline
\end{tabular}
}

\end{center}
\end{table}

Table \ref{tab:route} shows the result of Jensen-Shannon distance ($d_{JS}$) of route distribution tested with different models. The lower the distance value, the better the model performance. The best model (with the lowest value) in each test case (row) is marked in bold. With the Single-OD datasets, all models except MaxEnt(SVF) show good results with $d_{JS}$ less than $0.1$. The $d_{JS}$ of MaxEnt(SVF) is considerably larger than that of the other models. 
It is worth noting that MMC shows good performance, especially under the Proportional and Fixed demand patterns, which is surprising considering the simplicity of the MMC model. 
Although there are some differences in the distance measures, all models except MaxEnt(SVF) were able to reproduce a synthetic trajectory dataset with high degrees of statistical similarity to the real dataset, primarily because the travel patterns in the Single-OD dataset are very simple.

The Multi-OD datasets (One-way Multi-OD and Two-way Multi-OD) have more complex trajectory patterns than the Single-OD datasets. As a result, all models show an increase in $d_{JS}$. The increase rate is significantly large in the MaxEnt models. The main reason for this is because MaxEnt models use a simple linear function to describe reward functions and the linear reward function lacks the ability to model the complex non-linear patterns of real vehicle trajectories. Both RNN and TrajGAIL have recurrent neural networks in common to use sequential information in generating synthetic trajectory. Especially, the performance of TrajGAIL is noticeably better than the other models because it not only uses sequential embedding of visited locations but also uses the reward function from the discriminator, which offers an additional guidance for a model to produce trajectories that match the real observations.

% \begin{figure}[!ht]
% \centering
% \includegraphics[width=0.7\textwidth]{img/unknown.png}
% \caption{The number of unknown trajectories generated by each model in Gangnam dataset.}
% \label{fig:unknown}
% \end{figure}

% \begin{table}
% \caption{The number of unknown trajectories among 20,000 generated vehicle trajectories for Gangnam dataset}
% \label{tab:unknown}
% \begin{center}
% \begin{tabular}{c c c c c}
% \hline
% MMC  & RNN  & MaxEnt(SVF) & MaxEnt(SAVF) & TrajGAIL    \\ \hline
% 4892 & 2323 & 12300 & 11217 & 181\\ 
% \hline
% \end{tabular}
% \end{center}
% \end{table}

The model performance overall decreases when tested on the Gangnam dataset, i.e., the $d_{JS}$ values of all five models are above 0.4. TrajGAIL, however, still shows the best performance among the five models. The main reason for relatively large $d_{JS}$ values is that the real-world trajectory patterns in the Gangnam dataset are more complex and sparse than the simulated trajectory patterns in the other datasets. For instance, there are many rare routes with counts less than 25 among 59,553 trajectories in the Gangnam dataset. Such rare routes are difficult for a model to learn due to the limited sample sizes and, consequently, the models end up generating "unknown" trajectories when attempting to reproduce these rare routes. Among 20,000 generated trajectories, the number of unknown trajectories generated by TrajGAIL is 181, while the other four models generate more than 2,000 unknown trajectories.

\begin{figure}
\centering
\includegraphics[width=\textwidth]{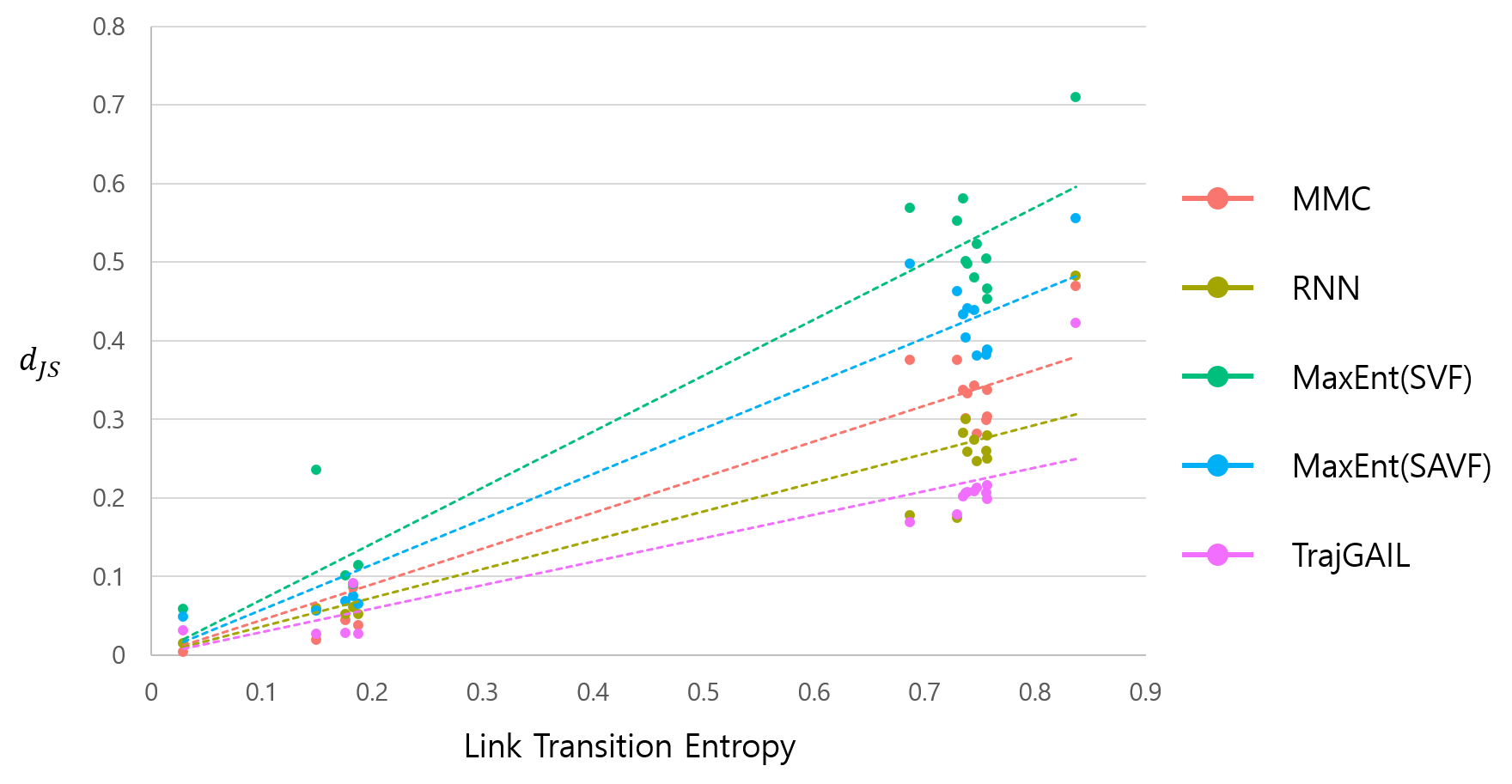}
\caption{Relationship between the link transition entropy and the Jensen-Shannon distance.}
\label{fig:entropy_jsd}
\end{figure}

In Table \ref{tab:route}, there is a tendency that $d_{JS}$ increases as the complexity of the dataset increases. To further investigate how the dataset complexity affects a model performance, we use information entropy to quantify the complexity of a dataset. More specifically, we define the \textit{link transition entropy} of a given vehicle trajectory dataset as follows:

\begin{align}
  \label{eq:entropy}
     & H(D) = \frac{1}{|L|} \sum_{l_i \in L} \Big(  \sum_{l_j \in L} -P(l_j|l_i) \log P(l_j|l_i) \Big) &
\end{align}

\noindent
where $H(D)$ is the link transition entropy of dataset $D$, $L$ is a set of possible links in $D$, and $P(l_j|l_i)$ is an empirical transition probability that a vehicle moves from link $l_i$ to link $l_j$. The intuition of this measure is to represent the dataset complexity in term of how complex or unpredictable a vehicle movement at each intersection is. For instance, a dataset would have a low $H(D)$ if all vehicles move in one direction at every intersection (e.g. all vehicles go straight at one intersection, all vehicles turn right at another intersection, and so forth), and a dataset would have a high $H(D)$ if all vehicles randomly choose the directions at all intersections.

Figure \ref{fig:entropy_jsd} shows the relationship between the link transition entropy and the Jensen-Shannon distance of route distribution from each model. In the figure, the entropy-distance results of each model are fitted into a linear model. The slope of the linear model is defined as the \textit{complexity sensitivity} of each model. A model with a high complexity sensitivity has a more significant drop in model performance as a dataset becomes more complex. On the other hand, a model with a low complexity sensitivity can learn the trajectory patterns regardless of the complexity of the dataset. The results show that TrajGAIL shows the lowest complexity sensitivity, suggesting its robustness and ability to scale to a larger and more complex datasets. RNN and MMC show the second and third lowest complexity sensitivity, while the two MaxEnt models have the highest complexity sensitivity.

\subsubsection{Discussion}

Overall, MaxEnt(SVF) showed poor performance in trajectory generation. One reason for this might be because it focuses on matching the state visitation frequency, which only cares about the element-wise similarity rather than the route-level similarity. When MaxEnt is trained to match the state-action visitation frequency, i.e., MaxEnt(SAVF), the performance improves to the level similar to the other models, which implies the importance of considering sequential information reflecting vehicles' link-to-link transition actions in generating realistic trajectories. The similar mechanism is also used in TrajGAIL, where the discriminator of TrajGAIL calculates the immediate reward based on the current state and the sampled action and this discriminator reward works in a similar way to matching the state-action visitation frequency. 

% Combining the result in Section \ref{sec:tra_eval} and the result in Section \ref{sec:data_eval}, it is to note that 
From the evaluation results, TrajGAIL and RNN are found to be the most suitable models for synthetic trajectory generation. TrajGAIL, however, outperforms RNN in most of the test cases, especially in matching route distributions to the real vehicle trajectory datasets. It is, thus, worth discussing in more detail the difference between RNN and TrajGAIL and how TrajGAIL addresses the limitations of RNN. 
Overall, the generator of TrajGAIL is similar to the RNN model. However, TrajGAIL's generator has better performance than the RNN model. 
% The difference comes in how the training procedure is designed for each model. 
The difference comes from the modeling architecture, i.e., how the model is designed to give proper training signals to the generator.
RNN is trained to minimize the cross-entropy loss between the probability of the predicted next location and the real next location as a label. This error is backpropagated through the input sequence to update model parameters to predict the next location based on the previously visited locations. While RNN can incorporate the previous locations into the next location prediction, it does not consider the rest of the trip. 
In contrast, TrajGAIL uses the reward function from the discriminator combined with the value estimator to consider the rest of the trip. The reward function from the discriminator captures how realistic the current state and action are. The value estimator then calculates the $\gamma$-discounted cumulative rewards, which capture how realistic the remaining states and actions will be. Using these two functions, the generator of TrajGAIL can learn a more comprehensive distribution of a given trajectory dataset and generate trajectories reflecting the sequential patterns along the whole trajectories in real data.

\section{Summary}\label{sec:conclusion}
% \subsection{Summary}
This study proposes TrajGAIL, a generative adversarial imitation learning framework for urban vehicle trajectory generation. In TrajGAIL, drivers' movement decisions in an urban road network are modeled as a partially observable Markov decision process (POMDP). The generative adversarial imitation learning is then used to learn the underlying decision process demonstrated in the given trajectory dataset. This allows TrajGAIL to generate new synthetic trajectory data that are similar to real-world trajectory observations. 

% In terms of contributions, this study have made several theoretical and methodological advancements in the field of trajectory analysis and synthetic trajectory generation. TrajGAIL adopted the combination of POMDP and RNN embedding to encode the historical sequence in a current state without violating the Markov assumption required for the MDP-based imitation learning framework. This overcomes the disadvantage of standard GAIL and traditional IRL (e.g., MaxEnt) that they cannot take into account the previous locations in selecting the next location in a trajectory. TrajGAIL demonstrated the superiority of GAIL over GAN in generating location sequences from a generative adversarial framework. The imitation learning approach in GAIL allows the consideration of not only the previous locations but also future locations to make the whole trajectories more realistic, while GAN does not take into account future locations and is prone to generate unrealistic trajectories with excessive lengths and several loops. In terms of technical contributions, the use of deep learning approach (i.e., GAIL) over traditional IRL in the context of imitation learning overcomes the IRL's drawback of high computation cost when dealing with a large state-action space and, thus, offers better scalability for the applications using large-scale traffic networks.

% the given trajectory dataset's underlying reward function. 

The model's performance is evaluated on different datasets with various traffic demand patterns against three baseline models from previous studies. The evaluation is divided into two levels: trajectory-level evaluation and dataset-level evaluation.
In the trajectory-level evaluation, the generated vehicle trajectories are evaluated in terms of BLEU and METEOR, the two most widely used scores in sequence modeling. In the dataset-level evaluation, the statistical similarity between the generated vehicle trajectory dataset and the real vehicle trajectory dataset is measured using the Jensen-Shannon distance of route distribution. The results show that TrajGAIL can successfully generate realistic trajectories that capture trajectory-level sequence patterns as well as match route distributions in the underlying datasets, evidenced by significantly higher performance measures of TrajGAIL compared to the other models. A model's performance sensitivity with respect to the complexity of a trajectory dataset was further investigated by measuring the link transition entropy of the dataset and analysing its relationship with model performance. The results show that TrajGAIL is least sensitive to dataset complexity among the tested models, suggesting its robustness in learning complex patterns of real vehicle movements and ability to scale to a larger and more complex dataset.

\chapter{Concluding Remark}\label{chapter:conclusion}

\section{Conclusion of Dissertation}
The overall objective of this dissertation is to leverage massive amounts of urban vehicle trajectory data, which become increasingly available nowadays, to better understand city mobility dynamics and enhance the design and operations of transportation systems.
% summary of 
The overall contribution of this dissertation is that we applied state-of-the-art deep learning models to the research topics in urban vehicle trajectory analytics to discover system-wide mobility patterns in urban traffic networks and to better understand spatiotemporal patterns in urban vehicle trajectory data.
In this dissertation, two main research topics are selected based on the necessity, importance, and applicability of deep learning methods. The selected research topics are \textit{next location prediction} and \textit{synthetic trajectory generation}. For each research topics, the current research challenges are identified from the previous studies, and appropriate research approaches are proposed to resolve the issues in the current research challenges. In addition, in the main chapters, based on the proposed research approaches, a novel deep-learning-based solutions to each research topic is proposed and the performance of the model is evaluated based on different measures against different baseline models.
In Section \ref{conclusion:remarks}, a detailed discussion on the contributions of each main chapters is presented, as well as the limitation and future research directions.

\section{Concluding Remark on Three Research Topics}\label{conclusion:remarks}
\subsection{Contribution}

% \subsubsection{Map Matching}

% \colorbox{yellow}{conclusion on map matching will be added later}

\subsubsection{Next Location Prediction}
The overall goal of \textit{Next location prediction} in Chapter \ref{chapter:nlp_rnn} and Chapter \ref{chapter:nlp_arnn} is the ability to predict individual vehicles’ movements — at least in terms of a sequence of aggregated spatial locations — and hence anticipate the flow of vehicles at a given location and time more accurately. Chapter \ref{chapter:nlp_rnn} showed a promising direction toward achieving this ability by applying deep learning with Recurrent Neural Networks (RNN) on urban vehicle trajectory data. As a way to represent complex urban vehicle trajectories as simpler location sequences, Chapter \ref{chapter:nlp_rnn} proposes a method to partition the network into cells (clustering-based Voronoi tessellation) so that entire vehicle movements can be expressed in terms of combinations of a finite set of cells. Mapping trajectories onto cells not only reduces computational complexity but also allows working with multi-source and multi-resolution trajectories. It not only smoothens local noises caused by the sensors and data collection system, but also improve interpretability of spatiotemporal features in urban vehicle trajectory data \cite{garcia2012survey}. Different cell sizes are tested for performance evaluation and discussions are provided on the impacts of cell size on trajectory prediction performance. 
In Chapter \ref{chapter:nlp_rnn}, RNN model is trained based on the large amounts of urban vehicle trajectory data from Bluetooth detectors installed in Brisbane, Australia, and the model is tested to predict the next locations of individual vehicles.
% Using large amounts of Bluetooth detector based urban vehicle trajectory data collected in Brisbane, Australia, Chapter \ref{chapter:nlp_rnn} trains a RNN model to predict cell sequences. 
The model performance is evaluated by computing the probability of correctly predicting the next $k$ consecutive cells. Compared to the baseline model (TRN) that relies on a simple transition matrix, the proposed RNN model shows substantially better prediction results. The network-level aggregate measures such as total cell visit count and inter-cell flow are also used in the performance evaluation and it is observed that the RNN model can replicate real-world traffic patterns.

Based on the RNN model in Chapter \ref{chapter:nlp_rnn}, in Chapter \ref{chapter:nlp_arnn}, a novel approach is proposed to incorporate network traffic state data into next location prediction for urban vehicle trajectory. Attention mechanism is used as an interface to connect the network traffic state input data to the vehicle trajectory predictor. ARNN model, which is Attention-based RNN model for cell sequence prediction, is compared with RNN model, which is RNN model for cell sequence prediction, in terms of conventional scoring methods in sequence prediction. The result shows that ARNN model outperformed RNN model. The result shows that it is effective to use attention mechanism to structurally connect the network traffic state input to RNN model to predict the vehicle’s future locations. Especially, it is promising that the ARNN model showed significant performance improvement in terms of METEOR which considers not only the cells to be visited but also the alignment of the cells in sequence. The performance improvement rates tend to decrease and converge to 1 as the original number of cell sequence increase. For the further improvement of the ARNN model, this problem should be studied to maintain the performance improvement rate at steady level.

\subsubsection{Synthetic Trajectory Generation}
Chapter \ref{chapter:trajgen} proposes TrajGAIL, a generative adversarial imitation learning framework for urban vehicle trajectory generation. In TrajGAIL, drivers' movement decisions in an urban road network are modeled as a partially observable Markov decision process (POMDP). The generative adversarial imitation learning is then used to learn the underlying decision process demonstrated in the given trajectory dataset. This allows TrajGAIL to generate new synthetic trajectory data that are similar to real-world trajectory observations. 

The model's performance is evaluated on different datasets with various traffic demand patterns against three baseline models from previous studies. The evaluation is divided into two levels: trajectory-level evaluation and dataset-level evaluation.
In the trajectory-level evaluation, the generated vehicle trajectories are evaluated in terms of BLEU and METEOR, the two most widely used scores in sequence modeling. In the dataset-level evaluation, the statistical similarity between the generated vehicle trajectory dataset and the real vehicle trajectory dataset is measured using the Jensen-Shannon distance of route distribution. The results show that TrajGAIL can successfully generate realistic trajectories that capture trajectory-level sequence patterns as well as match route distributions in the underlying datasets, evidenced by significantly higher performance measures of TrajGAIL compared to the other models. A model's performance sensitivity with respect to the complexity of a trajectory dataset was further investigated by measuring the link transition entropy of the dataset and analysing its relationship with model performance. The results show that TrajGAIL is least sensitive to dataset complexity among the tested models, suggesting its robustness in learning complex patterns of real vehicle movements and ability to scale to a larger and more complex dataset. 

Lastly, it is worth mentioning the connection between our TrajGAIL model and traditional route choice models in transportation such as discrete choice model and dynamic traffic assignment (DTA) because there are similarities between them in that they both aim to model an individual driver's choice behaviors in selecting routes in traffic networks. A main difference lies in how "route" and "choice" are defined. In traditional route choice models, a route is "selected" from a pre-defined set of alternative routes for a given OD and, thus, a choice is made at the OD and route level. In our generative modeling approach, on the other hand, a route is "constructed" as a result of sequential decisions (link-to-link transitions) along the journey and, thus, a choice is made at the intersection and link level. Although a decision is made locally at each intersection, the \textit{learned reward function} captures the network-wide route choice patterns and, thus, can give correct signals at each local decision point to enable the generated trajectories to exhibit realistic route- and network-level patterns globally. Another major difference lies in modeling approach. Traditional route choice models are "model-based" (or theory-based) in that they rely on behavioral assumptions and theories (e.g., utility maximization, user equilibrium) and aim to explain "why" a specific driver chooses a certain route. Our generative approach is "data-driven" and does not rely on any behavioral assumptions. It does not try to explain "why" but instead focuses "what" patterns exist in the actual realized data and "how" to reproduce them.

In sum, TrajGAIL offers a data-driven alternative to traditional route choice models for describing the underlying route distribution of a traffic network in synthetic data generation problems, without requiring the identification of ODs (Origin-Destination paris) and the computationally expensive enumeration of route sets. There are limitations in TrajGAIL, however, that currently it does not take into account the effects of traffic conditions or interactions with other vehicles and, thus, is unable to serve as a route choice model for DTA or traffic simulation models. It is an important and interesting future research topic to consider how data-driven deep learning models could complement or replace existing theory-based modeling components to improve DTA and traffic simulation. The extensions to TrajGAIL (e.g., attention, cGAIL, and infoGAIL) can allow the consideration of additional information and, thus, could be potentially used in that direction.

Chapter \ref{chapter:trajgen} have made several theoretical and methodological advancements in the field of trajectory analysis and synthetic trajectory generation. TrajGAIL adopted the combination of POMDP and RNN embedding to encode the historical sequence in a current state without violating the Markov assumption required for the MDP-based imitation learning framework. This overcomes the disadvantage of standard GAIL and traditional IRL (e.g., MaxEnt) that they cannot take into account the previous locations in selecting the next location in a trajectory. TrajGAIL demonstrated the superiority of GAIL over GAN in generating location sequences from a generative adversarial framework. The imitation learning approach in GAIL allows the consideration of not only the previous locations but also future locations to make the whole trajectories more realistic, while GAN does not take into account future locations and is prone to generate unrealistic trajectories with excessive lengths and several loops. In terms of technical contributions, the use of deep learning approach (i.e., GAIL) over traditional IRL in the context of imitation learning overcomes the IRL's drawback of high computation cost when dealing with a large state-action space and, thus, offers better scalability for the applications using large-scale traffic networks.

% \subsection{Contribution}
% \subsubsection{Map Matching}
% \subsubsection{Next Location Prediction}
% % Contribution for chapter 5

% % Contribution for chapter 6

% \subsubsection{Synthetic Trajectory Generation}

\newpage
\subsection{Limitation and Future Researches}
% \subsubsection{Map Matching}

% \colorbox{yellow}{limitation and future research directions on map matching will be added later}

\subsubsection{Next Location Prediction}
% Limitation for chapter 5
% There are some limitations in Chapter \ref{chapter:nlp_rnn}, so further works should cover such topics.
There are some limitations in Chapter \ref{chapter:nlp_rnn} and Chapter  \ref{chapter:nlp_arnn}, so future researches should cover such topics.
First of all, throughout Chapter \ref{chapter:nlp_rnn} and Chapter \ref{chapter:nlp_arnn}, we focus more on the behaviors in urban vehicle trajectories as a group than on the behaviors as an individual. The predictions that we produce usually focus on the \textit{average} behaviors of the users sharing the same traffic networks. This approach is used to understand the general behaviors and general patterns of urban traffic networks, but one can argue that user-specific predictions would bring more applicability of next location predictions. Using user-specific predictions can be applied to many applications, especially to Location-based Services (LBS). The models in Chapter \ref{chapter:nlp_rnn} and Chapter \ref{chapter:nlp_arnn} would be good starting points for user-specific predictions because they give a baseline of average behaviors. A user-specific prediction model can be developed by combining user-specific semantics with the baseline of average behaviors. However, it requires a massive amount of trajectory data for each user to develop and train a user-specific prediction model, which makes it difficult to do it at current state.

The second and third limitations are about the network-wide traffic state used in Chapter \ref{chapter:nlp_arnn}. Chapter \ref{chapter:nlp_arnn} proposed the attention-based recurrent neural network model (ARNN) which incorporates network-wide traffic state into next location prediction. Chapter \ref{chapter:nlp_arnn} showed a promising result by using traffic state information to enhance the performance of the next location prediction model based on recurrent neural network (RNN). In ARNN model, the network traffic state data is normalized by using the historical maximum value of each cell. This makes easy to represent the network traffic state, but this may lead to some problems that normalized data of the cells with very low traffic may be too sensitive to small number of vehicles and count it as heavy congestion. This makes the model overreact to these cells and makes the cell sequence prediction confused. For the further improvement of this study, different types of normalization methods should be tested. 

In Chapter \ref{chapter:nlp_arnn}, we only provide the network-wide traffic state at the beginning time of the trip. As we discussed with Figure \ref{figure5} and Figure \ref{figure6} in Section \ref{arnn:scoreresult}, the performance improvement rate decreases as the original sequence length increases.
This observation has an important implication for the influence of pre-trip information in route choice behaviors. The fact that the ARNN improves the prediction at the early stages of a journey implies that the pre-trip information indeed influences travellers’ route choice decisions and differentiates route choice patterns between different pre-trip traffic conditions. 
The fact that the effect of pre-trip information fades away at the later stages of the journey may be the indication of drivers’ reliance on en-route trip information instead of pre-trip information and thus indicates a need for incorporating such en-route information into the model to further improve the model performance.
To further improve the model, it is required to periodically update the network-wide traffic state that next location prediction model uses to consider the effect of en-route trip information as well as the effect of pre-trip information. 
Updating the traffic state information periodically requires an additional structure. This structure should have realistic assumptions. Since this is beyond the scope of this dissertation, so it will be considered as a future research direction.

There is a trade-off between choosing link-level analytics and cell-level analytics. The link-level analytics consider road links and POIs (Point-of-Interests) as analytical units. In contrast, cell-level analytics consider small traffic networks (or regions, cells, etc.) as analytical units. The link-level analytics deals with the split ratio at intersections and link-level route choice behavior. Cell-level analytics deals with regional traffic flow and macroscopic demand and routing patterns. It is possible to analyze microscopic behaviors with link-level analytics, but cell-level analytics ignore these behaviors and focus more on macroscopic behaviors. 

One significant shortcoming of cell-level analytics is that it assumes a "homogeneity" inside one cell. A homogeneity inside one cell means that vehicles are evenly distributed in a given cell, and the congestion levels of each link in the cell should be the same or similar. In many practical applications of cell-level analytics, this assumption works as a significant constraint. In many cases, it is not easy to guarantee homogeneity inside a cell. 
As a result, many research directions could be made due to this limitation. First, it is necessary to study proper network-partitioning methodology to guarantee homogeneity inside a cell. In this research, data points in all the trajectories are combined and clustered in space based on the desired radius. Through clustering, some level of homogeneity can be achieved, but it still needs a deep study on this. 
Second, it is essential to perform an error analysis to prove that there is no systematical error. Identifying major error sources and analyzing geographical locations of these error sources can be used to plan future research directions.
% 
% Lastly, it is important to connect cell-level analytics and link-level analytics. In this case, it is possible to combine link-level analytics using high-resolution data and cell-level analytics. This data fusion method 

% Also, it is possible to combine link-level analytics using high-resolution data and cell-level analytics using low-resolution data. There are many different vehicle trajectory data pr

% In the error analysis of \cite{choi2019real}, major errors of next location prediction occur at the boundaries of the study network. 

\subsubsection{Synthetic Trajectory Generation}
There are several directions in which the current study in Chapter \ref{chapter:trajgen} can be extended to further improve the trajectory generation performance. Currently, we only encode the sequence of links into the belief states in POMDP as a way to incorporate unobserved variables in modeling and predicting next locations. There are, however, other variables that can help sequence prediction in addition to the visited link sequence information. For instance, traffic conditions in a road network, vehicles' origin and destination information, and temporal information such as time-of-day and day-of-week can all provide additional information to further improve trajectory prediction and generation. While the current study did not consider this as we focus on introducing the theoretical aspects of TrajGAIL and analyzing the effects of the structural differences between TrajGAIL and other models, we will consider incorporating other variables in our future study.

Additional information can be incorporated into TrajGAIL in various ways. \cite{choi2019attention} used an attention-based RNN model to incorporate network traffic states into the next location prediction problem. A similar attention mechanism can be employed to give the network traffic state information to the generator of TrajGAIL. As human drivers use traffic state information from navigation services to make better decision in route choices, this attention mechanism can guide the generator to pay attention to the traffic states of certain locations in the road network to select actions more accurately, in a way that human drivers do.

Another way is to use a conditional version of generative adversarial modeling framework such as conditional-GAN (cGAN) \cite{mirza2014conditional} and conditional-GAIL (cGAIL) \cite{zhang2019unveiling}, which allow models to generate synthetic data conditioned on some extra information. 
For example, we can feed trajectories' "origin" locations as additional input to TrajGAIL during training such that the generator and discriminator are trained to learn vehicle trajectory patterns given origin locations. This would enable the model to further capture and distinguish different route choice behaviors specific to different origin regions.

Instead of feeding extra information to the model in a supervised way as in cGAN and cGAIL, it is also possible to achieve this in an unsupervised way as in infoGAN \cite{chen2016infogan} and infoGAIL \cite{li2017infogail}. Instead of explicit condition inputs, InfoGAN and infoGAIL introduce latent variables, which are used by a model to automatically distinguish certain behaviors in data in a meaningful and interpretable way. For instance, it is possible to build a trajectory generation model that can automatically cluster and distinguish patterns in trajectory data by different latent features such as origins, destinations, and time-varying traffic demands and use this information to guide the generating process.

% We have proposed a novel FD MIMO-CCRN framework providing a reasonable performance improvement compared with the conventional MIMO-CCRN framework...
% \cite{banerjee_meteor:_2005}
% \cite{choi2018network}
% 

\bibliographystyle{apalike}
\bibliography{main}

\acknowledgment[4]
박사 학위라는 긴 여정의 마침표를 찍기까지 많은 분들의 도움이 있었음을 다시금 생각하게 됩니다. 길다면 길고 짧다면 짧다고 할 수 있는 6년의 시간동안 많은 분들을 만났고, 한 분 한 분에게서 다양한 영감을 얻고 배울 수 있었습니다. 이렇게 감사의 글로 그 동안 저에게 도움을 주신 많은 분들께 마음을 전합니다.

먼저, 지도교수님이신 여화수 교수님께 깊은 감사와 존경의 인사를 전하고 싶습니다. 처음 연구실에 들어와서 연구를 시작한 순간부터 항상 따뜻한 조언과 지도를 아끼지 않고 해주신 덕분에 많이 성장할 수 있었습니다. 연구가 어렵거나, 갈 길을 잃었을 때 교수님께서 방향을 잡아주시고 이끌어 주셨기에 위기 때마다 주저앉지 않고 다음단계로 도약할 수 있었던 것 같습니다. 학자로서, 같은 연구 분야에 있는 선배님으로서 항상 존경할 수 있는 스승님을 둘 수 있다는 것은 크나큰 영광인 것 같습니다. 앞으로도 긴 학문의 길에 스승님의 은혜와 가르침을 잊지 않고 감사하는 마음을 가지고 연구에 매진하도록 노력하겠습니다.

다음으로는 바다건너 호주에서 가르침을 이어 주신 김지원교수님께 감사의 인사를 드리고 싶습니다. 김지원 교수님과의 연구가 시작된 이후, 연구를 보는 시야와 생각이 넓어질 수 있었습니다. 교수님과 함께 진행했던 여러가지 프로젝트, 실험들을 통해 교수님의 연구방법을 체득할 수 있었고, 힘든 박사과정을 즐거운 연구로 채워나갈 수 있었던 것 같습니다. 김지원 교수님 같은 멘토를 만나게 된 것은 저의 연구 인생에 크나큰 행운이라고 생각합니다. 진심으로 감사드립니다.
그리고 저의 박사학위 논문 심사를 맡아주신 장기태교수님, 김아영교수님 그리고 김영철교수님께 깊은 감사인사 올립니다. 교수님들의 깊은 식견을 통해 세심하게 조언해 주신 덕분에 부족했던 부분을 채울 수 있었고, 나아가 앞으로의 연구 방향에 대해서도 깊이 생각해 볼 수 있었던 소중한 시간이었습니다. 

6년간의 대학원 생활을 동고동락하며 함께 머리 맞대며 연구해준 연구실 사람들에게도 이 자리를 통해 고마움을 전합니다. 석사과정 때부터 많은 조언을 주신 세현이형과, 제가 힘들 때 마다 말동무가 되어 주었던 동훈이형, 그리고 본받고싶은 성훈이형과 같이 좋은 선배들을 둔 것은 저의 큰 행운이었습니다. 학위과정동안 함께 많은 토론을 한 교원이형, 진원이형, 예은이, 화평이, 그리고 특히나 이번 박사학위논문 심사를 같이 준비하며 서로 힘이 되어준 동호형과 정윤누나에게도 감사의 인사를 전합니다. 그리고 이제 우리 연구실의 미래를 책임질 지후, 주희, 지영이, 지웅이, 혜영이, 유진이, 등봉(Tengfeng)이와 수제에게도 감사한다는 말과 잘 부탁한다는 말 하고 싶습니다. 이외에도 현정누나, 병준씨, 성준이형, 시몬이형, 종해형, 민주형, 용준이형, 수민누나, 은혜누나에게도 감사드립니다.

이 자리에 있기까지 뒤에서 묵묵히 저를 믿어주시는 부모님께 깊이 감사드립니다. 대학생이 되어 집을 떠나와 자주 찾아뵙지도 못하는 아들을 항상 격려해주시고 도와주신 부모님이 계셨기에 긴 공부의 과정을 지치지않고 이어나갈 수 있었습니다. 또한 저를 항상 응원해주시는 누나와 매형, 장인어른과 장모님께도 감사의 인사 드립니다.

마지막으로, 파리에서 만난 그 순간부터 저의 한쪽이었던 나의 아내 이수영과 내 딸 최리아에게 감사의 뜻을 전하며 글을 마칩니다.

% 2021. 06. 22.
% 최성진 올림

%%
%% 약력 시작
%% Curriculum Vitae
%%
% @command curriculumvitae 이력서
% @options [1 | 2 | 3 |4 ]
% - 1 : 본문과 약력이 둘 다 한글일 때  | 2 : 본문은 한글인데 약력이 영어일 때 | 3 :  본문과 약력이 둘 다 영어일 때  | 4 : 본문은 영어인데 약력이 한글일 때 
%% It is optional and you can change form of this in the class file if you want.
\curriculumvitae[3]
\noindent {\Large \textbf{Seongjin Choi} (최성진)}
\vspace{2mm}

\noindent\begin{tabular*}{\textwidth}{@{\extracolsep{\fill}} l r}
\hline
\hline
  Email: & \href{mailto:benchoi93@kaist.ac.kr}{\color{black}{benchoi93@kaist.ac.kr}} \\
  Github: & \href{https://github.com/benchoi93}{\color{black}{https://github.com/benchoi93}} \\
  Google Scholar: & \href{https://scholar.google.com/citations?user=tyLWFk4AAAAJ}{\color{black}{https://scholar.google.com/citations?user=tyLWFk4AAAAJ}} \\
  Phone: & +82-10-5177-4130 \\  
\hline
\hline
\end{tabular*}

\vspace{2mm}
\noindent {\large \textbf{Education \& Training}}
\vspace{2mm}

\noindent\begin{tabular*}{\textwidth}{@{\extracolsep{\fill}} r r r}
\hline
2017.02 - 2021.08 & Ph.D. & \makecell[l]{Korea Advanced Institute of Science and Technology\\
Department of Civil and Environmental Engineering \\
\\
\textbf{Deep Learning Based Urban Vehicle Trajectory Analytics} \\
Superviser : Prof. Hwasoo Yeo
} \\ \hline
2015.09 - 2017.02 & M.S. & \makecell[l]{Korea Advanced Institute of Science and Technology\\ 
Department of Civil and Environmental Engineering \\
\\
\makecell[l]{\textbf{Development of Simulation-based Lane Change Control} \\ \textbf{for Autonomous Vehicles}}\\
Superviser : Prof. Hwasoo Yeo
} \\ \hline
2011.02 - 2015.08 & B.S. & \makecell[l]{Korea Advanced Institute of Science and Technology\\ 
Department of Civil and Environmental Engineering
} \\ \hline\hline
\end{tabular*}

\vspace{5mm}
\noindent {\large \textbf{Academic Achievement}}
% \vspace{2mm}

\noindent\begin{longtable}{@{\extracolsep{\fill}} l m{13cm} }
\hline
\noindent { \makecell[l]{\textbf{International} \\ \textbf{Journals}}} 
& \begin{itemize}
    \item \textbf{Choi, Seongjin}, Jiwon Kim, and Hwasoo Yeo. “TrajGAIL: Generating Urban Trajectories using Generative Adversarial Imitation Learning.” arXiv preprint arXiv:2007.14189 (2020). \textbf{[Accepted in Transportation Research Part C]}
    \item Lee, Donghoun, Sehyun Tak, \textbf{Seongjin Choi}, and Hwasoo Yeo. “Development of risk predictive collision avoidance system and its impact on traffic and vehicular safety.” Transportation research record 2673.7 (2019): 454-465.
\end{itemize}\\
\hline
\noindent { \makecell[l]{\textbf{International} \\ \textbf{Journals}}} 
& \begin{itemize}
    \item Kim, Yeeun, \textbf{Seongjin Choi}, and Hwasoo Yeo. “Extended Urban Cell Transmission Model Using Agent-based Modeling.” Procedia Computer Science 170 (2020): 354-361.
    \item Kim, Yeeun, \textbf{Seongjin Choi}, Jihyuk Park, and Hwasoo Yeo. “Agent-based Mesoscopic Urban Traffic Simulation based on Multi-lane Cell Transmission Model.” Procedia Computer Science 151 (2019): 240-247.
    \item \textbf{Choi, Seongjin}, Jiwon Kim, and Hwasoo Yeo. “Attention-based Recurrent Neural Network for Urban Vehicle Trajectory Prediction.” arXiv preprint arXiv:1812.07151 (2018).
    \item \textbf{Choi, Seongjin}, Hwasoo Yeo, and Jiwon Kim. “Network-wide Vehicle Trajectory Prediction in Urban Traffic Networks Using Deep Learning” Transportation Research Record (2018).
    \item \textbf{Choi, Seongjin}, Jonghae Suh, and Hwasoo Yeo. “Microscopic Analysis of Climbing Lane Performance at Freeway Uphill Section.” Transportation Research Procedia 21 (2017): 98-109.
\end{itemize}\\
\hline
\noindent { \makecell[l]{\textbf{International} \\ \textbf{Conferences}}} 
& \begin{itemize}
    \item \textbf{Choi, Seongjin}, Hwasoo Yeo, and Jiwon Kim. Network-wide Vehicle Trajectory Prediction in Urban Traffic Networks Using Deep Learning. The 99th Transportation Annual Meeting (January 2020).
    \item \textbf{Choi, Seongjin}, Jiwon Kim, Hwapyeong Yu, and Hwasoo Yeo. “Real-time Prediction of Arterial Vehicle Trajectories: An Application to Predictive Route Guidance for an * Emergency Vehicle.” 2019 IEEE Intelligent Transportation Systems Conference (ITSC). IEEE, (January 2019).
    \item Kim, Yeeun, \textbf{Seongjin Choi}, Jihyuk Park, and Hwasoo Yeo. Agent-based Mesoscopic Urban Traffic Simulation based on Multi-lane Cell Transmission Model. The 10th International Conference on Ambient Systems, Networks and Technologies. Acadia University, (May 2019).
\end{itemize}\\
\hline
\noindent { \makecell[l]{\textbf{International} \\ \textbf{Conferences}}} 
& \begin{itemize}
    \item \textbf{Choi, Seongjin}, Jiwon Kim, and Hwasoo Yeo. Attention-based Recurrent Neural Network for Urban Vehicle Trajectory Prediction. The 10th International Conference on Ambient Systems, Networks and Technologies. Acadia University, (May 2019).
    \item Lee, Donghoun, Sehyun Tak, \textbf{Seongjin Choi}, and Hwasoo Yeo. Development of risk predictive collision avoidance system and its impact on traffic and vehicular safety. The 98th Transportation Annual Meeting (January 2019).
    \item Kim, Yeeun, \textbf{Seongjin Choi}, and Hwasoo Yeo. Incorporation of Driver Distraction in Car-following model based on Driver’s Eye Glance Behavior. 2018 21st International Conference on Intelligent Transportation Systems (ITSC). IEEE, (October 2018).
    \item \textbf{Choi, Seongjin}, Hwasoo Yeo, and Jiwon Kim. Network-wide Vehicle Trajectory Prediction in Urban Traffic Networks Using Deep Learning. The 97th Transportation Annual Meeting (January 2018).
    \item \textbf{Choi, Seongjin}, Sehyun Tak, Jihu Kim, and Hwasoo Yeo. Traffic Event Classification using Convolutional Neural Network. The 30th KKHTCNN Symposium on Civil Engineering (November 2017).
    \item Tak, Sehyun, Hwasoo Yeo, Yeeun Kim and \textbf{Seongjin Choi}. A Study on the Dynamics of Driver Vision Transitions and its Impacts on Vehicle Safety. 10th SHRP 2 Safety Data-Symposium: From Analysis to Results (October 2017)
    \item Tak, Sehyun, Donghoun Lee, \textbf{Seongjin Choi}, and Hwasoo Yeo. Collision Avoidance System with Uni-directional Communication for Mitigating the Adverse Effects on Following Vehicles. Urban Transport 2017 (September 2017)
    \item \textbf{Choi, Seongjin}, and Hwasoo Yeo. Framework for simulation-based lane change control for autonomous vehicles. Intelligent Vehicles Symposium (IV), 2017 IEEE (June 2017).
\end{itemize}\\
\hline
\hline
\noindent { \makecell[l]{\textbf{International} \\ \textbf{Conferences}}} 
& \begin{itemize}
    \item Tak, Sehyun, \textbf{Seongjin Choi}, Donghoun Lee, and Hwasoo Yeo. A Comparison Analysis of Track-Based Train Operation System and Communication-Based Train Operation System for Train Safety. The 96th Transportation Research Board Annual Meeting (January 2017).
    \item Tak, Sehyun, \textbf{Seongjin Choi}, and Hwasoo Yeo. The Effect of Communication and GPS Uncertainty on Safety Performance of Communication-based Train Control. The 1st Asian Conference on Railway Infrastructure and Transportation, 359 (October 2016).
    \item \textbf{Choi, Seongjin}, Jonghae Suh, and Hwasoo Yeo. Microscopic Analysis of Climbing Lane Performance at Freeway Uphill Section. 2016 International Symposium of Transport Simulation (June 2016)
\end{itemize} \\
\hline
\noindent { \makecell[l]{\textbf{Domestic} \\ \textbf{Journals}}} 
& \begin{itemize}
    \item \textbf{Choi, Seongjin}, Jiwon Kim, Hwapyeong Yu, Dongho Ka, and Hwasoo Yeo. “Deep-learning based urban vehicle trajectory prediction.” Journal of Korean Society of Transportation 37.5 (2019): 422-429.
    \item Kim, Yeeun, \textbf{Seongjin Choi}, and Hwasoo Yeo. “A study on development of a car-following model for accident simulation caused by driver distraction.” Journal of Korean Society of Transportation 37.1 (2019): 39-50.
\end{itemize} \\ 
\hline\hline
\end{longtable}

\vspace{2mm}
\noindent {\large \textbf{Patents and Copyrights}}
% \vspace{2mm}

\noindent\begin{longtable}{@{\extracolsep{\fill}} l m{14cm} }
\hline\hline
\noindent { \textbf{Patents}} 
& \begin{itemize}
    \item 10-2018-0143229: Server and Method for Managing Shared Autonomous Vehicles
    \item 10-2018-0141857: Traffic Simulator for Verification of ITS System
\end{itemize} \\ 
\hline
\noindent { \textbf{Softwares}} 
& \begin{itemize}
    \item C-2016-011948: Advanced Oversaturated Freeway Flow Algorithm for Uphill Segment
    \item C-2017-003416: Simulation-based Lane Change Control System for Autonomous Vehicles
    \item C-2017-012423: Signal Control For Smart City (SCSC)
    \item C-2020-025123: Smart Traffic Simulator (STSIM)
\end{itemize} \\ 
\hline\hline
\end{longtable}

%     % @environment personaldata 개인정보
%     % @command     name         이름
%     %              dateofbirth  생년월일
%     %              birthplace   출생지
%     %              domicile     본적지
%     %              address      주소지
%     %              email        E-mail 주소
%     % - 위 6개의 기본 필드 중에 이력서에 적고 싶은 정보를 입력
%     % input data only you want
%     \begin{personaldata}
%         \name       {Seongjin Choi}
%         \dateofbirth{1993}{02}{25}
%         \email{benchoi93@kaist.ac.kr}
%     \end{personaldata}

%     % @environment education 학력
%     % @options [default: (none)] - 수학기간을 입력
%     \begin{education}
%         % \item[2008. 3.\ --\ 2011. 2.] 창동 고등학교 (졸업)
%         \item[2011. 2.\ --\ 2015. 8.] 한국과학기술원 건설및환경공학과 (학사)
%         \item[2015. 9.\ --\ 2017. 2.] 한국과학기술원 건설및환경공학과 (석사)
%         \item[2017. 3.\ --\ 2021. 8.] 한국과학기술원 건설및환경공학과 (박사)
%     \end{education}

%     % @environment career 경력
%     % @options [default: (none)] - 해당기간을 입력
%     % \begin{career}
%     %     \item[2015. 3.\ --\ 2016. 2.] 한국과학기술원 전기및전자공학부 조교
%     % \end{career}

%     % @environment activity 학회활동
%     % @options [default: (none)] - 활동내용을 입력
% %%    \begin{activity}
% %%        \item J. Choi, \textbf{Yong-Hyun Kim}, K.J. Chang, and D. Tomanek,
% %%             \textit{Occurrence of itinerant ferromagnetism in C/BN superlattice
% %%             nanotubes}, 5th Asian Workshop on First-Principles Electronic
% %%             Structure Calculations, Seoul (Korea), October., 2002.
% %%    \end{activity}

%     % @environment publication 연구업적
%     % @options [default: (none)] - 출판내용을 입력
%     \textbf{International Journals}

  \label{paperlastpagelabel}     % <-- 추가 부분: 마지막 페이지 위치 지정	
%% 본문 끝
\end{document}